# On Action Theory Change

**Ivan José Varzinczak**                                    IVAN.VARZINCZAK@MERAKA.ORG.ZA
*Meraka Institute, CSIR*
*Pretoria, South Africa*

## Abstract

As historically acknowledged in the Reasoning about Actions and Change community, intuitiveness of a logical domain description cannot be fully automated. Moreover, like any other logical theory, action theories may also evolve, and thus knowledge engineers need revision methods to help in accommodating new incoming information about the behavior of actions in an adequate manner. The present work is about changing action domain descriptions in multimodal logic. Its contribution is threefold: first we revisit the semantics of action theory contraction proposed in previous work, giving more robust operators that express minimal change based on a notion of distance between Kripke-models. Second we give algorithms for syntactical action theory contraction and establish their correctness with respect to our semantics for those action theories that satisfy a principle of modularity investigated in previous work. Since modularity can be ensured for every action theory and, as we show here, needs to be computed at most once during the evolution of a domain description, it does not represent a limitation at all to the method here studied. Finally we state AGM-like postulates for action theory contraction and assess the behavior of our operators with respect to them. Moreover, we also address the revision counterpart of action theory change, showing that it benefits from our semantics for contraction.

## 1. Introduction

Consider an intelligent agent designed to perform rationally in a dynamic world, and suppose that she should reason about the dynamics of an automatic coffee machine (Figure 1).

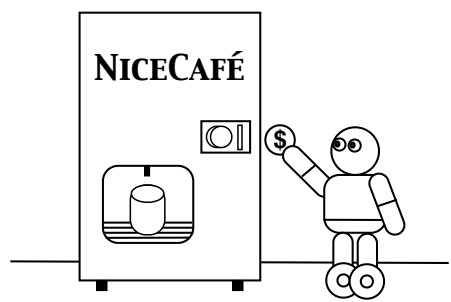

Figure 1: The coffee deliverer agent.

Suppose, for example, that the agent believes that coffee is always a hot beverage. Suppose now that some day she gets a coffee at the machine and observes that it is cold. In such a case, the agent must change her beliefs about the relationship between the two propositions "I hold a coffee" and "I hold a hot drink". This example is an instance of the problem of changing propositional belief bases and has been largely addressed in the literature about belief revision (Alchourrón, Gärdenfors, & Makinson, 1985; Gärdenfors, 1988; Hansson, 1999) and belief update (Katsuno & Mendelzon, 1992).

Next, let our agent believe that whenever she buys a coffee from the machine, she gets a hot drink. This means that in every state of the world that follows the execution of buying a coffee, the agent ends up with a hot drink. Now, in a situation where the machine is running out of cups, after buying, the coffee runs through the shelf and the agent, contrary to what she was expecting, does *not* hold a hot drink in her hands.

Imagine now that the agent never considered any relation between buying a coffee on the machine and its service availability, in the sense that she always believed (quite reasonably) that buying does not prevent other users from using the machine. Nevertheless, someday our agent is queuing to buy a coffee and observes that just after the agent before her has bought, the machine went out of order (maybe due to a lack of coffee powder).

Completing our agent's struggle in discovering the intricacies of operating a coffee machine, let us suppose now that she always believed that if she has a token, then it is possible to buy coffee, provided that some other preconditions like being close enough to the button, having a free hand, etc, are satisfied. Eventually, due to a blackout, the agent realizes that she does not manage to buy her coffee, even with a token.

The last three examples illustrate cases in which changing the beliefs about the *behavior* of the action of buying coffee is mandatory. In the first one, buying coffee, once believed to have a deterministic outcome, namely always a hot drink, has now to be seen as nondeterministic or, alternatively, to have a different effect in a more specific context (e.g. if there is no cup in the machine). In the second example, buying a coffee is now known to have side-effects (ramifications) which one was not aware of. Finally, in the last example, the feasibility of the action under concern is questioned in the light of new information showing a context that was not known to preclude its execution.

Such cases of theory change are very important when one deals with logical descriptions of dynamic domains: it may always happen that one discovers that an action actually has a behavior that is different from that one has always believed it had.

Up to now, theory change has been studied mainly for knowledge bases in classical logics, both in terms of revision and update. Since the work by Fuhrmann (1989), only in a few recent studies has it been considered in the realm of modal logics, viz. in epistemic logic (Hansson, 1999) and in dynamic logics (Herzig, Perrussel, & Varzinczak, 2006). Recently some studies have investigated revision of beliefs about *facts* of the world (Shapiro, Pagnucco, Lespérance, & Levesque, 2000; Jin & Thielscher, 2005) or the agent's goals (Shapiro, Lespérance, & Levesque, 2005). In our scenario, this would concern for instance the truth of *token* in a given state: the agent believes that she has a token, but is actually wrong about that. Then she might subsequently be forced to revise her beliefs about the current state of affairs or change her goals according to what she can perform in that state. Such belief revision operations do not modify the agent's beliefs about the *action laws*. On the other hand, here we are interested exactly in such modifications. Starting with Baral and Lobo's work (1997), some recent studies have been done on that issue (Eiter, Erdem, Fink, & Senko, 2005) for domain descriptions in action languages (Gelfond & Lifschitz, 1993).

We here take a step further in this direction and propose a method which is more robust by integrating a notion of *minimal change* and complying with *postulates* of theory change.

The present text is structured as follows: in Section 2 we establish the formal background that will be used throughout this article. Sections 3–6 are the core of the work: in Section 3 we present the central definitions for a semantics of action theory change, providing justifications for the design choices here made (Section 4). Section 5 is devoted to the syntactical counterpart of our operators while Section 6 to the proof of its correspondence with the semantics under certain acceptable conditions. In Section 7 we discuss some postulates for contraction/erasure and then present a semantics for action theory revision (Section 8). After a discussion on and comparison with existing work in the field (Section 9), we conclude with an overview and future directions of research.

## 2. Logical Preliminaries

Following the tradition in the Reasoning about Actions and Change (RAC) community, we consider action theories to be finite collections of statements that have the particular form (Shanahan, 1997):

- if *context*, then *effect* after *every execution* of *action* (effect laws);

- if *precondition*, then *action executable* (executability laws).

Statements mentioning no action at all represent laws about the underlying structure of the world, i.e., its possible states (static laws).

Several logical frameworks have been proposed to formalize such statements (Shanahan, 1997). Among the most prominent ones are the first-order based Situation Calculus (McCarthy & Hayes, 1969; Reiter, 2001), the family of Action Languages (Gelfond & Lifschitz, 1993; Giunchiglia, Kartha, & Lifschitz, 1997), the Fluent Calculus (Thielscher, 1997), and Propositional Dynamic Logic (PDL) (Harel, Tiuryn, & Kozen, 2000) with different specific extensions thereof (De Giacomo & Lenzerini, 1995; Castilho, Gasquet, & Herzig, 1999; Zhang & Foo, 2001; Castilho, Herzig, & Varzinczak, 2002).

Here we opt to formalize action theories using the multimodal logic $K_n$ (Popkorn, 1994). Among the main reasons for such a choice are:

- We benefit from the well defined semantics for multimodal logics which, as we are going to see in the sequel, provides simple and intuitive foundations on which to build the meaning of changing action domain descriptions.

- $K_n$ syntax allows us to express all the afore mentioned types of laws without requiring the full expressiveness of PDL or the machinery of a first-order language.

- Since $K_n$ is the core of all above mentioned PDL-based action formalisms, all we shall say in the sequel should smoothly transfer to them.

- Contrary to first-order based approaches, $K_n$ is decidable and has several implemented theorem provers for it available in the literature.

## 2.1 Action Theories in Multimodal Logic

Let $\mathfrak{Act} = \{a_1, a_2, \ldots, a_n\}$ be the set of all atomic *action constants* of a given dynamic domain. An example of atomic action is *buy*. To each atomic action $a$ there is associated a modal operator $[a]$. We here suppose that our multimodal logic is *independently axiomatized* (Kracht & Wolter, 1991), i.e., the logic is a fusion and there is no interaction between the different modal operators.[1]

$\mathfrak{Prop} = \{p_1, p_2, \ldots, p_n\}$ denotes a finite set of *propositional constants*, also called *fluents* or *elementary atoms*. Examples of those are *token* ("the agent has a token") and *coffee* ("the agent holds a coffee"). $\mathfrak{Lit} = \{p, \neg p \; : \; p \in \mathfrak{Prop}\}$ is the set of *literals*. We use $\ell$ to denote a literal. If $\ell = \neg p$, then we identify $\neg \ell$ with $p$. By $|\ell|$ we denote the atom in $\ell$.

We use small Greek letters $\varphi, \psi, \ldots$ to denote *Boolean (propositional) formulas*. They are recursively defined in the usual way:

$$\varphi \; ::= p \mid \top \mid \bot \mid \neg \varphi \mid \varphi \wedge \varphi \mid \varphi \vee \varphi \mid \varphi \rightarrow \varphi \mid \varphi \leftrightarrow \varphi$$

($\varphi \oplus \psi$ denotes $(\varphi \vee \psi) \wedge \neg(\varphi \wedge \psi)$.) $\mathfrak{Fml}$ is the set of all Boolean formulas. An example of a Boolean formula is *coffee* $\rightarrow$ *hot*. A propositional valuation $v$ is a *maximal consistent* set of literals. We denote by $v \Vdash \varphi$ the fact that $v$ satisfies a propositional formula $\varphi$. By $val(\varphi)$ we denote the set of all valuations satisfying $\varphi$. By $\mathsf{CPL}$ we denote Classical Propositional Logic and $\models_{\overline{\mathsf{CPL}}}$ is its respective consequence relation. $Cn(\varphi)$ denotes all logical consequences of $\varphi$ in $\mathsf{CPL}$, i.e., $Cn(\varphi) = \{\psi \; : \; \varphi \models_{\overline{\mathsf{CPL}}} \psi\}$.

If $\varphi$ is a propositional formula, $atm(\varphi)$ denotes the set of elementary atoms *actually* occurring in $\varphi$. For example, $atm(\neg p_1 \wedge (\neg p_1 \vee p_2)) = \{p_1, p_2\}$.

For $\varphi$ a Boolean formula, $IP(\varphi)$ denotes the set of its *prime implicants* (Quine, 1952), i.e., the weakest terms (conjunctions of literals) that imply $\varphi$. As an example, $IP(p_1 \oplus p_2) = \{p_1 \wedge \neg p_2, \neg p_1 \wedge p_2\}$. For more on prime implicants, their properties and how to compute them, see the chapter by Marquis (2000). With $\pi$ we denote a prime implicant, and given $\ell$ and $\pi$, $\ell \in \pi$ abbreviates '$\ell$ is a literal of $\pi$'. For a given set $X$, $\bar{X}$ denotes its complement. Hence $\overline{atm(\pi)}$ denotes $\mathfrak{Prop} \setminus atm(\pi)$.

We denote complex formulas (possibly with modal operators) by $\Phi, \Psi, \ldots$ They are recursively defined in the following way:

$$\Phi \; ::= \varphi \mid [a]\Phi \mid \neg \Phi \mid \Phi \wedge \Phi \mid \Phi \vee \Phi \mid \Phi \rightarrow \Phi \mid \Phi \leftrightarrow \Phi$$

$\langle a \rangle$ is the dual operator of $[a]$, defined by $\langle a \rangle \Phi =_{\text{def}} \neg [a] \neg \Phi$. An instance of a complex formula in our scenario example is $\neg coffee \rightarrow [buy] coffee$.

Given a complex formula $\Phi$, with $act(\Phi)$ we denote the action names occurring in $\Phi$, i.e., the modalities of $\Phi$. For example, $act([a_2]p_1 \wedge ([a_1]p_2 \rightarrow [a_2]p_3)) = \{a_1, a_2\}$.

The semantics here is the standard semantics of multimodal logic $\mathsf{K}_n$ (Popkorn, 1994).

**Definition 2.1 ($\mathsf{K}_n$-Model)** *A $\mathsf{K}_n$-model is a tuple $\mathscr{M} = \langle W, R \rangle$ where $W$ is a set of valuations (also called possible worlds), and $R$ maps action constants $a$ to accessibility relations $R_a \subseteq W \times W$.*

---

1. Later on we will see that this is a requirement to ensure that an action theory is modular.

As an example, for $\mathfrak{Act} = \{a_1, a_2\}$ and $\mathfrak{Prop} = \{p_1, p_2\}$, we have the $\mathsf{K}_n$-model $\mathscr{M} = \langle W, R \rangle$, where

$$W = \{\{p_1, p_2\}, \{p_1, \neg p_2\}, \{\neg p_1, p_2\}\},$$

$$R(a_1) = \left\{ \begin{array}{c} (\{p_1, p_2\}, \{p_1, \neg p_2\}), (\{p_1, p_2\}, \{\neg p_1, p_2\}), \\ (\{p_1, \neg p_2\}, \{p_1, \neg p_2\}), (\{p_1, \neg p_2\}, \{\neg p_1, p_2\}) \end{array} \right\}$$

$$R(a_2) = \{(\{p_1, p_2\}, \{\neg p_1, p_2\}), (\{\neg p_1, p_2\}, \{\neg p_1, p_2\})\}$$

Figure 2 gives a graphical representation of the model $\mathscr{M}$.

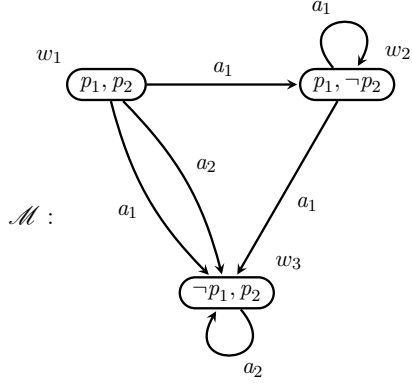

Figure 2: Example of a $\mathsf{K}_n$-model for $\mathfrak{Act} = \{a_1, a_2\}$, and $\mathfrak{Prop} = \{p_1, p_2\}$.

Notice that our definition of $\mathsf{K}_n$-model does not follow the traditional notion from modal logics: here no two worlds satisfy the same valuation. This is a pragmatic choice, as we will see in Section 5. Nevertheless, all we shall say in the sequel can be straightforwardly formulated for standard $\mathsf{K}_n$ models as well.

**Definition 2.2 (Truth Conditions)** *Given a $\mathsf{K}_n$-model $\mathscr{M} = \langle W, R \rangle$,*

- $\models^{\mathscr{M}}_{w} p$ *($p$ is true at world $w$ of $\mathscr{M}$) iff $w \Vdash p$ (valuation $w$ satisfies $p$, i.e., $p \in w$);*

- $\models^{\mathscr{M}}_{w} [a]\Phi$ *iff $\models^{\mathscr{M}}_{w'} \Phi$ for every $w'$ such that $(w, w') \in R_a$;*

- $\models^{\mathscr{M}}_{w} \Phi \wedge \Psi$ *iff $\models^{\mathscr{M}}_{w} \Phi$ and $\models^{\mathscr{M}}_{w} \Psi$;*

- $\models^{\mathscr{M}}_{w} \neg \Phi$ *iff $\not\models^{\mathscr{M}}_{w} \Phi$, i.e., not $\models^{\mathscr{M}}_{w} \Phi$;*

- *truth conditions for the other connectives are as usual.*

By $\mathcal{M}$ we will denote a (possibly empty) set of $\mathsf{K}_n$-models.

A $\mathsf{K}_n$-model $\mathscr{M}$ is a model of $\Phi$ (denoted $\models^{\mathscr{M}} \Phi$) if and only if for all $w \in W$, $\models^{\mathscr{M}}_{w} \Phi$. In the model depicted in Figure 2, we have $\models^{\mathscr{M}} p_1 \rightarrow [a_2]p_2$ and $\models^{\mathscr{M}} p_1 \vee p_2$. $\mathscr{M}$ is a model of a set of formulas $\Sigma$ (noted $\models^{\mathscr{M}} \Sigma$) if and only if $\models^{\mathscr{M}} \Phi$ for every $\Phi \in \Sigma$. If $\Sigma$ is the set of formulas we start off with (our non-logical theory), then each $\Phi \in \Sigma$ is called a *global axiom*.

**Definition 2.3 (Global Consequence)** *A formula $\Phi$ is a global consequence of a set of global axioms $\Sigma$ in the class of all $\mathsf{K}_n$-models (noted $\Sigma \models_{\overline{\mathsf{K}_n}} \Phi$) if and only if for every $\mathsf{K}_n$-model $\mathscr{M}$, if $\models^{\mathscr{M}} \Sigma$, then $\models^{\mathscr{M}} \Phi$.*

With $\mathsf{K}_n$ we can state laws describing the behavior of actions. One way of doing this is by stating some formulas as global axioms.[2] As usually done in the RAC community (Shanahan, 1997), we here distinguish three types of laws. The first kind of statements are *static laws*, which are constraints on the allowed states of a dynamic domain.

**Definition 2.4 (Static Law)** *A static law is a global axiom $\varphi \in \mathfrak{Fml}$.*

An example of a static law is *coffee* → *hot*, saying that if the agent holds a coffee, then she holds a hot drink. In the Situation Calculus formalism (Reiter, 2001) one would write the first-order formula $\forall s.[coffee(s) \rightarrow hot(s)]$. The set of all static laws of a scenario is denoted by $\mathcal{S} \subseteq \mathfrak{Fml}$. In our example we will have $\mathcal{S} = \{coffee \rightarrow hot\}$.

The second kind of action law we consider is given by the *effect laws*. These are formulas relating an action to its effects, which can be conditional.

**Definition 2.5 (Effect Law)** *Let $\varphi, \psi \in \mathfrak{Fml}$. An effect law for action $a$ is a global axiom of the form $\varphi \rightarrow [a]\psi$.*

The consequent $\psi$ is the effect which always obtains in accessible states (which need not exist in general) when action $a$ is executed in a state where the antecedent $\varphi$ holds. In our Kripke semantics, this means that in every possible world where $\varphi$ holds, every transition by an $a$-labeled arrow (if any) leads to a possible world where $\psi$ holds. If $a$ is a nondeterministic action, then the consequent $\psi$ is typically a disjunction. An example of an effect law is ¬*coffee* → [*buy*]*coffee*, saying that in a situation where the agent has no coffee, after buying, the agent has a coffee. If $\psi$ is inconsistent, then we have a special kind of effect law that we call an *inexecutability law*. For example, we could also have ¬*token* → [*buy*]⊥, expressing that *buy* cannot be executed if the agent has no token. In the Situation Calculus our examples of effect and inexecutability laws would be expressed respectively as $\forall s.[\neg coffee(s) \rightarrow coffee(do(buy,s))]$ and $\forall s.[\neg token(s) \rightarrow \neg Poss(buy,s)]$.

The set of effect laws of a given scenario is denoted by $\mathcal{E}$. In our coffee machine scenario, we could have for example:

$$\mathcal{E} = \left\{ \begin{array}{l} \neg coffee \rightarrow [buy]coffee, \\ token \rightarrow [buy]\neg token, \\ \neg token \rightarrow [buy]\bot \end{array} \right\}$$

Finally, we also define *executability laws*, which stipulate the context where an action is guaranteed to be executable. In $\mathsf{K}_n$, the operator $\langle a \rangle$ is used to express executability. $\langle a \rangle \top$ thus reads "the execution of $a$ is possible". Formally, $\langle a \rangle \top$ being true in a world $w$ means that there is at least one world $w'$ accessible from $w$ via $R_a$ (cf. Definition 2.2).

---

2. An alternative to that is given by Castilho et al. (1999, 2002), with laws being stated with the aid of an extra universal modality and local consequence being thus considered.

**Definition 2.6 (Executability Law)** *Let $\varphi \in \mathfrak{Fml}$. An* executability law *for action $a$ is a global axiom of the form $\varphi \to \langle a \rangle \top$.*

For instance, $token \to \langle buy \rangle \top$ says that buying can be executed whenever the agent has a token. The set of all executability laws of a given domain is denoted by $\mathcal{X}$. In our scenario example we will have $\mathcal{X} = \{token \to \langle buy \rangle \top\}$.

Note that in principle one needs to know nothing about the accessible world $w'$. However, a common (albeit tacit) assumption in the RAC community is that we state executability laws only for actions of which we know the effects, in other words $act(\mathcal{X}) \subseteq act(\mathcal{E})$.

In the Situation Calculus our example would be stated as $\forall s.[token(s) \to Poss(buy, s)]$. However, we point out that, traditionally, in Reiter basic action theories (Reiter, 2001) executability laws and inexecutability laws are mixed together in the form of bi-conditionals like $\forall s.[token(s) \leftrightarrow Poss(buy, s)]$, called *precondition axioms*. For a critique of such a practice and its implications in formalizing dynamic domains, see the work by Herzig and Varzinczak (2007).

With our three basic types of laws, we are able to define action theories:

**Definition 2.7 (Action Theory)** *Given any (possibly empty) sets of laws $\mathcal{S}$, $\mathcal{E}$, and $\mathcal{X}$, $\mathcal{T} = \mathcal{S} \cup \mathcal{E} \cup \mathcal{X}$ is an* action theory.

Given an action theory $\mathcal{T}$ and an action $a$, $\mathcal{E}_a$ (resp. $\mathcal{X}_a$) will denote the set of only those effect (resp. executability) laws about $a$ in $\mathcal{E}$ (resp. $\mathcal{X}$). $\mathcal{T}_a = \mathcal{S} \cup \mathcal{E}_a \cup \mathcal{X}_a$ is then the action theory for $a$.

It is worth noting that for $a_1, a_2 \in \mathfrak{Act}$, $a_1 \neq a_2$, the intuition is indeed that $\mathcal{T}_{a_1}$ and $\mathcal{T}_{a_2}$ overlap only on $\mathcal{S}$, i.e., the only laws that are common to both $\mathcal{T}_{a_1}$ and $\mathcal{T}_{a_2}$ are the laws about the structure of the world. This requirement is somehow related with the underlying modal logic being independently axiomatized (see note above).

## 2.2 The Frame, Ramification and Qualification Problems

During the last 40 years, most of the effort in the reasoning about actions community has been devoted to searching for satisfactory solutions to the frame problem, the ramification problem and the qualification problem.

Roughly speaking, the frame problem (McCarthy & Hayes, 1969) relates to the need for inferring the persistence of some facts of the world after the execution of an action known not to affect them, without having to state that explicitly in the form of frame axioms. (Frame axioms are a special type of effect law, having the form $\ell \to [a]\ell$, for $\ell \in \mathfrak{Lit}$.) In our example, buying a coffee in a context where the agent has already got one does not make it lose the coffee: $coffee \to [buy]coffee$ should be a consequence of our theory. The ramification problem (Finger, 1987) comes from the observation that an action may have several possibly interdependent effects and stating all of them explicitly is a huge task. In our scenario, we want to be able to infer $[buy]hot$ without saying it in the theory, and in such a way some intrinsic causal connection between *coffee* and *hot* is taken into account. Finally, the qualification problem (McCarthy, 1977) amounts to addressing the issue of ensuring that an action is executable in a given context. Specifying all the sufficient conditions for an action to be executable is an incredibly hard task. In our example, one may state $token \rightarrow \langle buy \rangle \top$, but it may well be the case that buying fails due to some condition unforeseen at design time, like the agent's arm being rusty and stuck.

For more on these core problems of the RAC community, the reader is referred to the book by Shanahan (1997).

For the sake of clarity, here we abstract from the frame and ramification problems, and suppose that the agent's theory already entails all the relevant frame axioms. We point out however that all we shall say could have been defined within a formalism with a solution to the frame and ramification problems. For instance, we could have used any suitable solution to the frame problem, like e.g. the dependence relation (Castilho et al., 1999), which is used in the work of Herzig et al. (2006), or a kind of successor state axioms in a slightly modified setting (Demolombe, Herzig, & Varzinczak, 2003). To make the presentation more clear to the reader, here we do not bother with a particular solution to the frame problem and just assume that all frame axioms can be inferred from the action theory. Actually we can suppose that all intended frame axioms are automatically recovered and stated in the theory, more specifically, in the set of effect laws.

Given the largely acknowledged difficulty of the qualification problem in the literature (Shanahan, 1997), we do not assume here any a priori solution to it. Instead, as tacitly assumed in many approaches to reasoning about actions (Castilho et al., 1999; Zhang & Foo, 2001; Reiter, 2001), we suppose that the knowledge engineer may want to state some (not necessarily fully specified) executability laws for some actions. These may be incorrect at the starting point (and in all probability they will be), but revising wrong executability laws is an approach towards its solution and one of the aims of this work. With further information the knowledge engineer will have the chance to change them so that eventually they will correspond to the intuition (cf. Sections 3 and 8).

Having agreed on these points, the action theory of our example will be:

$$\mathcal{T} = \left\{ \begin{array}{c} coffee \rightarrow hot, token \rightarrow \langle buy \rangle \top, \\ \neg coffee \rightarrow [buy]coffee, \\ token \rightarrow [buy]\neg token, \neg token \rightarrow [buy]\bot, \\ coffee \rightarrow [buy]coffee, hot \rightarrow [buy]hot \end{array} \right\}$$

(We have not stated the frame axiom $\neg token \rightarrow [buy]\neg token$ because it can be trivially deduced from the inexecutability law $\neg token \rightarrow [buy]\bot$.)

Figure 3 below shows a $\mathsf{K}_n$-model for the action theory $\mathcal{T}$ above.

We are going to see in the sequel that the finite base $\mathcal{T}$ formalizing the action theory plays a role in the contraction of laws. In particular, the base representing the static laws turns out to be quite important. So given an action theory $\mathcal{T}$, it will be useful to consider models of $\mathcal{T}$ whose possible worlds are *all* the possible valuations allowed by $\mathcal{S}$:

**Definition 2.8 (Canonical Frame)** *Let $\mathcal{T} = \mathcal{S} \cup \mathcal{E} \cup \mathcal{X}$ be an action theory. Then the tuple $\mathscr{M}_{can} = \langle W_{can}, R_{can} \rangle$ is the* canonical frame *of $\mathcal{T}$ if and only if:*

- $W_{can} = val(\mathcal{S})$; *and*

- $R_{can} = \bigcup_{a \in \mathfrak{Act}} R_a$ *s.t.* $R_a = \{(w, w') : for all \varphi \rightarrow [a]\psi \in \mathcal{E}_a, if \models_w^{\mathscr{M}} \varphi, then \models_{w'}^{\mathscr{M}} \psi\}$.

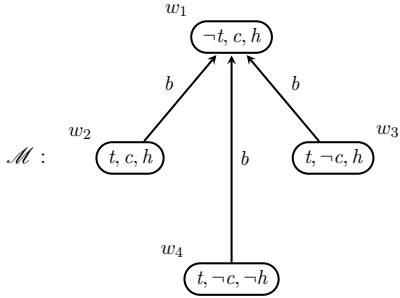

Figure 3: A model for our coffee machine scenario: $b$, $t$, $c$, and $h$ stand for, respectively, *buy*, *token*, *coffee*, and *hot*.

The canonical frame of an action theory need not be one of its models. To witness why, let $\mathfrak{Prop} = \{p\}$, $\mathfrak{Act} = \{a\}$, and consider the simple action theory $\{p \to [a]\bot, p \to \langle a\rangle\top\}$. Then in the associated canonical frame we have $W_{can} = \{\{p\}, \{\neg p\}\}$. Clearly the world $\{p\}$ does not satisfy this theory.

**Definition 2.9 (Canonical Model)** *$\mathscr{M}$ is a* canonical model *of $\mathcal{T}$ if and only if $\mathscr{M}$ is a canonical frame of $\mathcal{T}$ and $\models^{\mathscr{M}} \mathcal{T}$.*

Figure 4 below shows the canonical model of our action theory example $\mathcal{T}$.

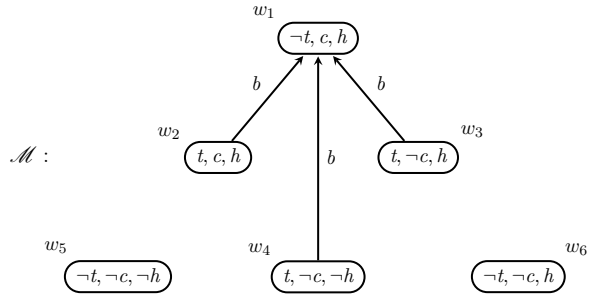

Figure 4: The canonical model for the coffee machine scenario.

## 2.3  Prime Valuations

We say that an atom $p$ is *essential* to a formula $\varphi$ if and only if $p \in atm(\varphi')$ for every $\varphi'$ such that $\models_{\mathsf{CPL}} \varphi \leftrightarrow \varphi'$. For instance, $p_1$ is essential to $\neg p_1 \land (\neg p_1 \lor p_2)$. Given $\varphi$, $atm!(\varphi)$ denotes the set of essential atoms of $\varphi$. (If $\varphi$ is not contingent, i.e., $\varphi$ is a tautology or a contradiction, then $atm!(\varphi) = \emptyset$.)

Given $\varphi$ a Boolean formula, $\varphi*$ is the set of all formulas $\varphi'$ such that $\varphi \models_{\mathsf{CPL}} \varphi'$ and $atm(\varphi') \subseteq atm!(\varphi)$. For instance, $p_1 \lor p_2 \notin p_1*$, as $p_1 \models_{\mathsf{CPL}} p_1 \lor p_2$ but $atm(p_1 \lor p_2) \not\subseteq atm!(p_1)$. Clearly, $atm(\bigwedge \varphi*) = atm!(\bigwedge \varphi*)$, moreover whenever $\models_{\mathsf{CPL}} \varphi \leftrightarrow \varphi'$ is the case, then $atm!(\varphi) = atm!(\varphi')$ and also $\varphi* = \varphi'*$.

**Theorem 2.1 (Least Atom-Set Theorem, Parikh, 1999)** *Let $\varphi$ be a propositional formula. Then $\models_{\overline{CPL}} \varphi \leftrightarrow \bigwedge \varphi*$, and for every $\varphi'$ such that $\models_{\overline{CPL}} \varphi \leftrightarrow \varphi'$, $atm(\varphi*) \subseteq atm(\varphi')$.*

A proof of this theorem is given by Makinson (2007) and we do not state it here. Essentially, the theorem establishes that for every Boolean formula $\varphi$, there is a unique least set of elementary atoms such that $\varphi$ may equivalently be expressed using only atoms from that set. Hence, $Cn(\varphi) = Cn(\varphi*)$.

Given a valuation $v$, $v' \subseteq v$ is a *subvaluation*. Given a set of valuations $W$, a subvaluation $v'$ *satisfies* a propositional formula $\varphi$ modulo $W$ (noted $v' \Vdash_W \varphi$) if and only if $v \Vdash \varphi$ for all $v \in W$ such that $v' \subseteq v$. We say that a subvaluation $v$ *essentially satisfies* $\varphi$ (modulo $W$), noted $v \Vdash_W^! \varphi$, if and only if $v \Vdash_W \varphi$ and $\{|\ell| : \ell \in v\} \subseteq atm!(\varphi)$. If $v \Vdash_W^! \varphi$, we call $v$ an *essential subvaluation* of $\varphi$ (modulo $W$).

**Definition 2.10 (Prime Subvaluation)** *Let $\varphi$ be a Boolean formula and $W$ a set of valuations. A subvaluation $v$ is a* prime subvaluation *of $\varphi$ (modulo $W$) if and only if $v \Vdash_W^! \varphi$ and there is no $v' \subset v$ such that $v' \Vdash_W^! \varphi$.*

A prime subvaluation of a formula $\varphi$ is thus one of the weakest states of truth in which $\varphi$ is true. Hence, prime subvaluations are just another way of seeing prime implicants (Quine, 1952) of $\varphi$. By $base(\varphi, W)$ we will denote the set of all prime subvaluations of $\varphi$ modulo $W$.

**Proposition 2.1** *Let $\varphi \in \mathfrak{Fml}$ and $W$ be a set of valuations. Then for all $w \in W$, $w \Vdash \varphi$ if and only if $w \Vdash \bigvee_{v \in base(\varphi, W)} \bigwedge_{\ell \in v} \ell$.*

**Proof:** Right to left direction is straightforward. For the left to right direction, if $w \Vdash \varphi$, then $w \Vdash \varphi*$. Let $w' \subseteq w$ be the least subset of $w$ still satisfying $\varphi*$. Clearly, $w'$ is a prime subvaluation of $\varphi$ modulo $W$, and then because $w \Vdash \bigwedge_{\ell \in w'} \ell$, the result follows. □

## 2.4 Closeness between Models

When contracting a formula from a model, we will perform a change in its structure. Because there can be several different ways of modifying a model (not all of them minimal), we need a notion of distance between models to identify those that are closest to the original one.

As we are going to see in more depth in the next section, changing a model amounts to modifying its possible worlds or its accessibility relation. Hence, the distance between two $\mathsf{K}_n$-models will depend upon the distance between their sets of worlds and accessibility relations. These here will be based on the *symmetric difference* between sets, defined as $X \dot{-} Y = (X \setminus Y) \cup (Y \setminus X)$.

**Definition 2.11 (Closeness between $\mathsf{K}_n$-Models)** *Let $\mathscr{M} = \langle W, R \rangle$ be a model. Then $\mathscr{M}' = \langle W', R' \rangle$ is at least as close to $\mathscr{M}$ as $\mathscr{M}'' = \langle W'', R'' \rangle$, noted $\mathscr{M}' \preceq_\mathscr{M} \mathscr{M}''$, if and only if*

- *either $W \dot{-} W' \subseteq W \dot{-} W''$;*

- *or $W \dot{-} W' = W \dot{-} W''$ and $R \dot{-} R' \subseteq R \dot{-} R''$.*

This is an extension of Burger and Heidema's relation (Burger & Heidema, 2002) to our modal case. It defines a lexicographic order on the set of all $\mathsf{K}_n$-models. Although simple, this notion of closeness turns out to be sufficient for our purposes here, as we shall see in the sequel. Notice that other notions of distance between models could have been considered as well, namely the *cardinality* of symmetric differences or Hamming distance. (See Section 4 for a discussion on this.)

## 3. Semantics of Action Theory Change

When admitting the possibility of a law $\Phi$ failing, one must ensure that $\Phi$ becomes invalid, i.e., not true in at least one model of the dynamic domain that is formalized. Because there can be lots of such models, we may have a *set* $\mathcal{M}$ of models in which $\Phi$ is (potentially) valid. Thus contracting $\Phi$ amounts to making it no longer valid in this set of models. What are the operations that must be carried out to achieve that? Throwing models out of $\mathcal{M}$ does not work, since $\Phi$ will keep on being valid in all models of the remaining set. Thus one should *add* new models to $\mathcal{M}$. Which models? Well, models in which $\Phi$ is not true. But not any of such models: taking models falsifying $\Phi$ that are too different from our original models will certainly violate the principle of minimal change.

Hence, we shall take some model $\mathcal{M} \in \mathcal{M}$ as basis and manipulate it to get a new model $\mathcal{M}'$ in which $\Phi$ is not true. In our modal semantics, the removal of a law $\Phi$ from a model $\mathcal{M} = \langle W, R \rangle$ means modifying the possible worlds or the accessibility relation in $\mathcal{M}$ so that $\Phi$ becomes false. Such an operation gives as result a *set* $\mathcal{M}_{\Phi}^{-}$ of models each of which is no longer a model of $\Phi$. But if there are several candidates, which ones should we choose? We shall take those models that are *minimal* modifications of the original $\mathcal{M}$, i.e., those which are minimal with respect to our distance $\preceq_{\mathcal{M}}$ between models. Of course, there can be more than one such an $\mathcal{M}'$ that is minimal with respect to $\preceq_{\mathcal{M}}$. In that case, because adding just one of these new models is enough to invalidate $\Phi$, we take all possible combinations $\mathcal{M} \cup \{\mathcal{M}'\}$ of expanding our original set of models $\mathcal{M}$ by one of these minimal models. (Observe that this approach relates to orderly maxichoice contraction Hansson, 1999.) The result will be a *set of sets of models*. In each set of models there will be precisely one model $\mathcal{M}'$ falsifying $\Phi$.

It might be claimed that, as such, our contraction method described above does not respect the so-called principle of categorical matching: the input and output are different sorts of objects, namely a set of models and a set of sets of models. It is easy to see, however, that the reasoning above can be stated in such a way that each output set of models corresponds precisely to the result of *one* contraction operator, satisfying then the referred principle. The choice for defining the result of an operation as a set of possible outputs will become more clear in Section 5, where we are going to present algorithms that correspond exactly to our semantic constructions.

### 3.1 Model Contraction of Executability Laws

To contract an executability law $\varphi \to \langle a \rangle \top$ from *one* model, intuitively we should *remove transitions* leaving $\varphi$-worlds. In order to succeed in the operation, we have to guarantee that in the resulting model there will be at least one $\varphi$-world with no departing $a$-arrow.

**Definition 3.1** *Let $\mathcal{M} = \langle W, R \rangle$. $\mathcal{M}' = \langle W', R' \rangle \in \mathscr{M}^-_{\varphi \to \langle a \rangle \top}$ if and only if*

- $W' = W$;

- $R' \subseteq R$;

- *if $(w, w') \in R \setminus R'$, then $\models^{\mathscr{M}}_w \varphi$; and*

- *there is $w \in W'$ such that $\not\models^{\mathscr{M}'}_w \varphi \to \langle a \rangle \top$.*

Observe that $\mathscr{M}^-_{\varphi \to \langle a \rangle \top} \neq \emptyset$ if and only if $\varphi$ is satisfiable in $W$. Moreover, $\mathscr{M} \in \mathscr{M}^-_{\varphi \to \langle a \rangle \top}$ if and only if $\not\models^{\mathscr{M}} \varphi \to \langle a \rangle \top$.

Just to provide the reader with an insight on how this operation would be carried out in the Situation Calculus, there one should look at a given situation $s$ in which $\varphi$ holds and then modify the interpretation of the predicate $Poss(a)$ so that it becomes false in $s$. Like in our case, there may be many of such situations and then all of them must be taken into account. An essential difference here is that our Kripke structures are always finite, whereas the space of situations is possibly infinite (Reiter, 2001).

To get minimal change, we want such an operation of removing transitions to be minimal with respect to the original model: one should remove a minimum set of transitions which is sufficient to get the desired result.

**Definition 3.2** $contract(\mathcal{M}, \varphi \to \langle a \rangle \top) = \bigcup \min \{ \mathscr{M}^-_{\varphi \to \langle a \rangle \top}, \preceq_{\mathscr{M}} \}$

And now we define the sets of possible models resulting from the contraction of an executability law in a set of models:

**Definition 3.3** *Let $\mathcal{M}$ be a set of models, and $\varphi \to \langle a \rangle \top$ an executability law. Then*

$$\mathcal{M}^-_{\varphi \to \langle a \rangle \top} = \{ \mathcal{M}' : \mathcal{M}' = \mathcal{M} \cup \{\mathscr{M}'\}, \mathscr{M}' \in contract(\mathscr{M}, \varphi \to \langle a \rangle \top), \mathscr{M} \in \mathcal{M} \}$$

In our running example, consider $\mathcal{M} = \{\mathscr{M}\}$, where $\mathscr{M}$ is the model in Figure 4. When the agent discovers that even with a token she does not manage to buy a coffee any more, she has to change her models in order to admit (new) models with states where *token* is the case but from which there is no *buy*-transition at all. Because having just one such a world in each new model is enough, taking those resulting models whose accessibility relations are maximal guarantees minimal change. Hence we will have $\mathcal{M}^-_{token \to \langle buy \rangle \top} = \{\mathcal{M} \cup \{\mathscr{M}'_1\}, \mathcal{M} \cup \{\mathscr{M}'_2\}, \mathcal{M} \cup \{\mathscr{M}'_3\}\}$, where each $\mathscr{M}'_i$ is depicted in Figure 5.

Clearly, if $\varphi$ is not satisfied in $\mathcal{M}$, i.e., $\models^{\mathscr{M}} \neg \varphi$ for all $\mathscr{M} \in \mathcal{M}$, then the contraction of $\varphi \to \langle a \rangle \top$ does *not* succeed. This is in line with the expectations and it relates to the Success Postulate (cf. Section 7.2).

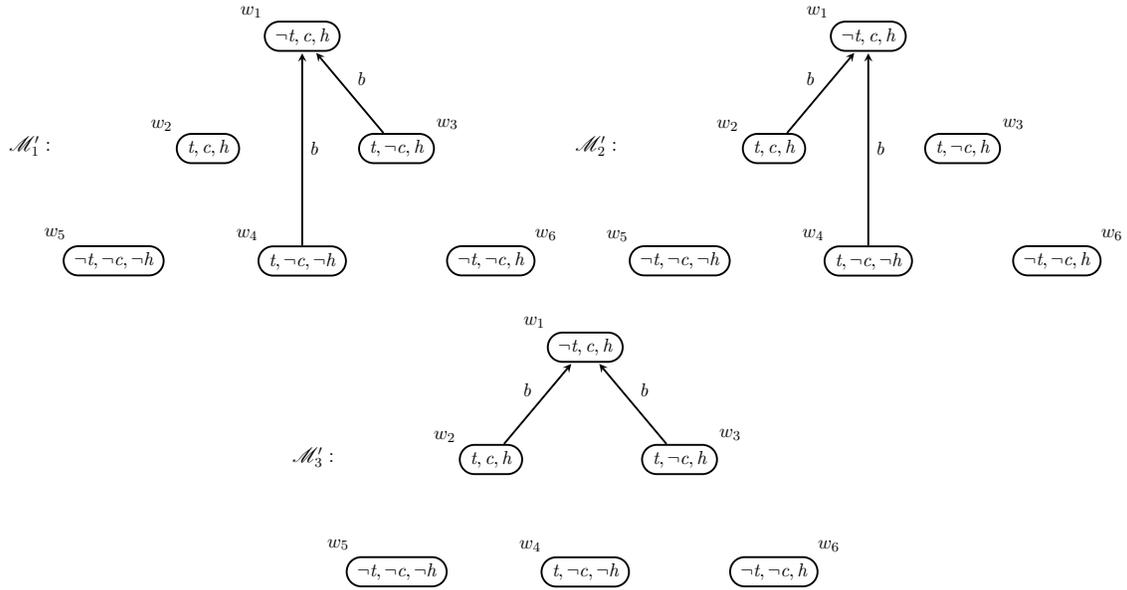

Figure 5: Models resulting from contracting $token \to \langle buy \rangle \top$ in the model $\mathscr{M}$ of Figure 4.

## 3.2 Model Contraction of Effect Laws

When our agent discovers that there may be some cases where after buying she gets no hot drink, she must e.g. give up the belief in the effect law $token \to [buy]hot$ in her set of models. This means that $token \land \langle buy \rangle \neg hot$ shall now be admitted in at least one world of some of the new models of her set of beliefs. Therefore, to contract an effect law $\varphi \to [a]\psi$ from a given model, intuitively we have to *add new transitions* from $\varphi$-worlds to worlds not satisfying $\psi$. As we shall see, the great challenge in such an operation is precisely how to guarantee minimal change.

In our example, when contracting $token \to [buy]hot$ from the model of Figure 4 we shall add transitions from $token$-worlds to $\neg hot$-worlds. Because $coffee \to hot$ is a static law and so is $\neg hot \to \neg coffee$, this should also give us $\langle buy \rangle \neg coffee$ in some $token$-world ($\neg coffee$ is causally relevant to $\neg hot$, i.e., to have $\neg hot$ we must also have $\neg coffee$). This means that if we allow for $\langle buy \rangle \neg hot$ in some $token$-world, we also have to allow for $\langle buy \rangle \neg coffee$ in that same world. The same argument does not necessarily hold for $token$: allowing for $\langle buy \rangle \neg hot$ does not necessarily oblige us to allow for $\langle buy \rangle token$ in the respective world. This is because $token$ is not relevant to $\neg hot$ (as $\neg coffee$ is). This means that we have the freedom either to allow for it or not.

Hence, in our running example we can add transitions from $token$-worlds to $\neg hot \land \neg coffee \land token$-worlds, as well as to $\neg hot \land \neg coffee \land \neg token$. This situation is depicted in Figure 6. For instance, we can add a new $buy$-arrow from the world $\{token, \neg coffee, \neg hot\}$ to one of these candidates (Figure 7).

In the Situation Calculus, such a modification would be slightly different, but with the same intuition behind: one should look at a given situation $s$ in which $\varphi$ holds and then modify the interpretation of the fluents (atoms) in $do(a, s)$, the situation resulting from performing $a$ in $s$. Alternatively, new $\varphi$-situations should lead to at least one $\neg \psi$-situation.

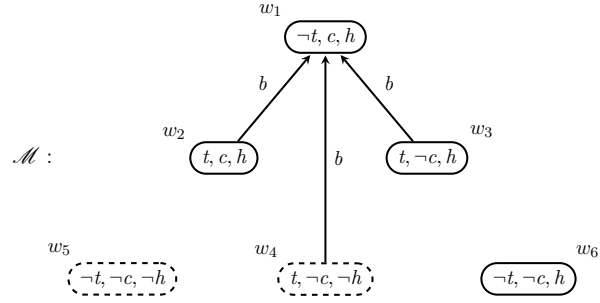

Figure 6: Candidate worlds to receive transitions coming from *token*-worlds.

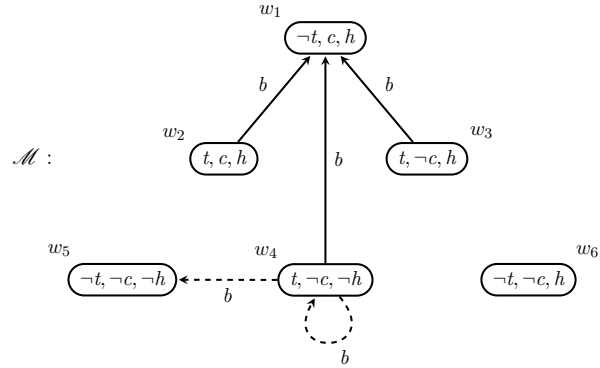

Figure 7: Two candidate new *buy*-arrows to falsify *token* → [*buy*]*hot* in $\mathcal{M}$.

Notice however that this would require the addition of new whole branches to the tree-like first-order model induced by Reiter basic action theories (Reiter, 2001).

Back to our example, observe that adding the new transition to {*token*, ¬*coffee*, ¬*hot*} itself would make us lose the effect ¬*token*, true after every execution of *buy* in the original model ($\models^{\mathcal{M}} token → [buy]¬token$). How do we preserve this law while allowing for the new transition to a ¬*hot*-world? That is, how do we get rid of the effect *hot* without losing effects that are not relevant for that? We here develop an approach for this issue.

When adding a new transition leaving a world $w$ we intuitively want to preserve as many effects as we had before doing so. To achieve this, it is enough to preserve old effects only in $w$ (because the remaining structure of the model remains unchanged after adding the new transition). Of course, we cannot preserve effects that are inconsistent with ¬$\psi$ (those will all be lost). Hence it suffices to preserve only the effects that are consistent with ¬$\psi$. To achieve that we must observe what is true in $w$ and in the target world $w'$:

- The proper effects of the action in world $w'$, i.e., what changes from $w$ to $w'$ ($w' \setminus w$) through the *new* execution of $a$ must be what is obliged to be so: either because those literals that now change from $w$ to $w'$ are necessary to having ¬$\psi$ in $w'$ (like ¬*coffee* in our example) or because they are necessary to have another effect (independent of ¬$\psi$, like ¬*token*) in world $w'$.

- The non-effects of action $a$ in world $w'$, i.e., what does not change from $w$ to $w'$ ($w \cap w'$) through $a$'s *new* execution should be only what is allowed to be so: certain literals are never preserved (like *token* in our example), then when pointing the new transition towards a world where it does not change with respect to the leaving world ($\neg hot \wedge \neg coffee \wedge token$ in our example), we may lose effects that held in $w$ before adding the transition.

This means that the only things allowed to change in the candidate target world must be those that are forced to change, either by some non-related law or because of having $\neg \psi$ modulo a set of states $W$. In other words, we want the literals that (now) change from $w$ to $w'$ to be *at most* those that are sufficient to get $\neg \psi$ modulo $W$, while preserving the maximum of other effects. Every change beyond that is not an intended one. Similarly, we want the literals from $w$ that are (now) preserved in the target world $w'$ to be *at most* those that are usually preserved in a given set of models. Every preservation beyond those may make us lose some law. This looks like prime implicants, and that is where prime subvaluations play their role: the worlds to which the new transition will point are those whose difference with respect to the departing world are literals that are relevant and whose similarity with respect to it are literals that we know do not change.

**Definition 3.4 (Relevant Target Worlds)** *Let $\mathscr{M} = \langle W, R \rangle$ be a model, $w, w' \in W$, $\mathcal{M}$ a set of models such that $\mathscr{M} \in \mathcal{M}$, and $\varphi \to [a]\psi$ an effect law. Then $w'$ is a relevant target world of $w$ with respect to $\varphi \to [a]\psi$ for $\mathscr{M}$ in $\mathcal{M}$ if and only if*

- $\models^{\mathscr{M}}_{w} \varphi$ *and* $\not\models^{\mathscr{M}}_{w'} \psi$;

- *for all* $\ell \in w' \setminus w$:

  - *either there is* $v \in base(\neg \psi, W)$ *such that* $v \subseteq w'$ *and* $\ell \in v$;
  - *or there is* $\psi' \in \mathfrak{Fml}$ *such that there is* $v' \in base(\psi', W)$ *such that* $v' \subseteq w'$, $\ell \in v'$, *and for every* $\mathscr{M}_i \in \mathcal{M}$, $\models^{\mathscr{M}_i}_{w} [a]\psi'$

- *for all* $\ell \in w \cap w'$:

  - *either there is* $v \in base(\neg \psi, W)$ *such that* $v \subseteq w'$ *and* $\ell \in v$;
  - *or there is* $\mathscr{M}_i \in \mathcal{M}$ *such that* $\not\models^{\mathscr{M}_i}_{w} [a]\neg \ell$;

*By $RelTarget(w, \varphi \to [a]\psi, \mathscr{M}, \mathcal{M})$ we denote the set of all relevant target worlds of $w$ with respect to $\varphi \to [a]\psi$ for $\mathscr{M}$ in $\mathcal{M}$.*

Note that we need the set of models $\mathcal{M}$ (and here we can suppose it contains all models of the theory we want to change) because preserving effects depends on what other effects hold in the other models that interest us. We need to take them into account in the local operation of changing one model. (The reason we do not need $\mathcal{M}$ in the definition of the local, one model contraction of executability laws $\mathscr{M}^{-}_{\varphi \to \langle a \rangle \top}$ is that when removing transitions there is no way of losing effects, as every effect law that held in the world from which a transition has been removed remains true in the same world in the resulting model.)

**Definition 3.5** *Let $\mathscr{M} = \langle W, R \rangle$, and $\mathcal{M}$ be such that $\mathscr{M} \in \mathcal{M}$. Then $\mathscr{M}' = \langle W', R' \rangle \in \mathscr{M}_{\varphi \to [a]\psi}^{-}$ if and only if*

- $W' = W$;

- $R \subseteq R'$;

- *If $(w, w') \in R' \setminus R$, then $w' \in RelTarget(w, \varphi \to [a]\psi, \mathscr{M}, \mathcal{M})$; and*

- *there is $w \in W'$ such that $\not\models_w^{\mathscr{M}'} \varphi \to [a]\psi$.*

Observe that $\mathscr{M}_{\varphi \to [a]\psi}^{-} \neq \emptyset$ if and only if $\varphi$ and $\neg\psi$ are both satisfiable in $W$. Moreover, $\mathscr{M} \in \mathscr{M}_{\varphi \to [a]\psi}^{-}$ if and only if $\not\models^{\mathscr{M}} \varphi \to [a]\psi$.

Because having just one world where the law is no longer true in each model is enough, taking those resulting models whose accessibility relations are minimal with respect to the original one guarantees minimal change.

**Definition 3.6** $contract(\mathscr{M}, \varphi \to [a]\psi) = \bigcup \min\{\mathscr{M}_{\varphi \to [a]\psi}^{-}, \preceq_{\mathscr{M}}\}$

Now we can define the possible sets of models resulting from contracting an effect law from a set of models:

**Definition 3.7** *Let $\mathcal{M}$ be a set of models, and $\varphi \to [a]\psi$ an effect law. Then*

$$\mathcal{M}_{\varphi \to [a]\psi}^{-} = \{\mathcal{M}' : \mathcal{M}' = \mathcal{M} \cup \{\mathscr{M}'\}, \mathscr{M}' \in contract(\mathscr{M}, \varphi \to [a]\psi), \mathscr{M} \in \mathcal{M}\}$$

Taking again $\mathcal{M} = \{\mathscr{M}\}$, where $\mathscr{M}$ is the model in Figure 4, after contracting *token* $\to$ *[buy]hot* from $\mathcal{M}$ we get $\mathcal{M}_{token \to [buy]hot}^{-} = \{\mathcal{M} \cup \{\mathscr{M}_1'\}, \mathcal{M} \cup \{\mathscr{M}_2'\}, \mathcal{M} \cup \{\mathscr{M}_3'\}\}$, where all $\mathscr{M}_i'$'s are as depicted in Figure 8.

In both cases where $\varphi$ is not satisfiable in $\mathscr{M}$ or $\psi$ is valid in $\mathscr{M}$, of course our operator does not succeed in falsifying $\varphi \to [a]\psi$ (cf. end of Section 3.1). Again, this works as expected and it has to do with the Success Postulate (see also Section 7.2).

### 3.3 Model Contraction of Static Laws

When contracting a static law from a model, we want to admit the existence of at least one (new) possible state falsifying it. This means that intuitively we should *add new worlds* to the original model. (In a Situation Calculus setting that would correspond to allowing for situations not satisfying some of the domain constraints.) This is quite easy. A very delicate issue however is what to do with the accessibility relation: should new transitions leave/arrive at the new world? If no transition leaves the new added world, we may lose some executability law. If some transition leaves it, then we may lose some effect law, the same holding if we add a transition pointing to the new world. On the other hand, if no transition arrives at the new world, what about the intuition? Is it intuitive to have an unreachable state? (Similar issues would also arise in Situation Calculus interpretations, which means that they are *independent* of the underlying formalism.)

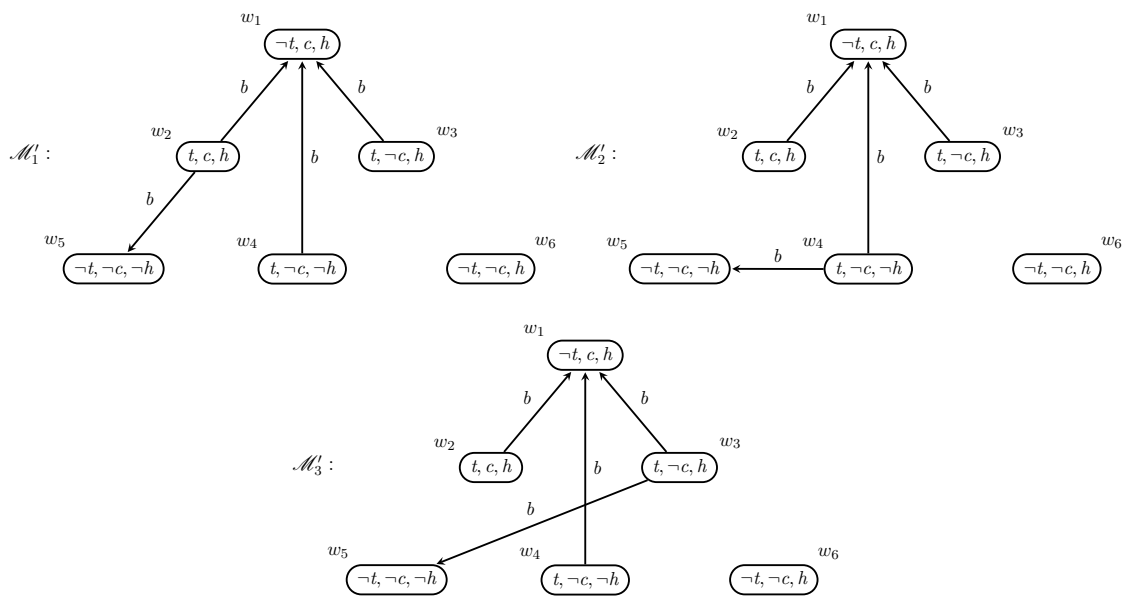

Figure 8: Models resulting from contracting $token \to [buy]hot$ in the model $\mathscr{M}$ of Figure 4.

All this discussion shows how drastic a change in the static laws might be: it is a change in the underlying structure (possible states) of the world! Changing them may have as an indirect, unexpected (and in all probability unwanted) consequence the loss of some effect law(s) or some executability law(s). What we can do is choose which type(s) of laws we may accept to lose in this process and then postpone their change (by the other operators).

Following the tradition in the RAC community, which states that executability laws are in general more difficult to formalize than effect laws, and therefore they are more likely to be incorrect (Shanahan, 1997), here we prefer not to change the accessibility relation, which means that we preserve effect laws and postpone the correction of executability laws, if required. (Remember that this is an approach towards a solution to the qualification problem — cf. Section 2.2 above.)

One may argue that doing things this way makes our three operators incoherent in the sense that for effect and executability laws we adopt a minimal change approach, giving stronger theories, whereas for static laws we adopt a more cautious approach, giving weaker theories (see the next section). It is worth noting however that as largely recognized by the RAC community, the different laws of a domain description do *not* have the same status: a minimal change approach for static law contraction that preserves as many executability laws as possible, even if coherent, would definitely fail to cope with the qualification problem. Moreover, by propagating wrong executability laws, such a coherent method would definitely be less elaboration tolerant (McCarthy, 1998) than the one we are defining with regards to further modifications of the theory.

For those reasons, our contention here is that static law contraction should be cautious. (For a detailed discussion on this, see Section 4.2 below and the end of Section 5.3.)

**Definition 3.8** *Let* $\mathcal{M} = \langle W, R \rangle$. $\mathcal{M}' = \langle W', R' \rangle \in \mathcal{M}_\varphi^-$ *if and only if*

- $W \subseteq W'$;

- $R = R'$; *and*

- *there is* $w \in W'$ *such that* $\not\models_w^{\mathcal{M}'} \varphi$.

Note that $\mathcal{M}_\varphi^- = \emptyset$ if and only if $\models \varphi$. Moreover, $\mathcal{M} \in \mathcal{M}_\varphi^-$ if and only if $\not\models^{\mathcal{M}} \varphi$.

The minimal modifications of one model are defined as usual:

**Definition 3.9** $contract(\mathcal{M}, \varphi) = \bigcup \min\{\mathcal{M}_\varphi^-, \preceq_{\mathcal{M}}\}$

And now we define the sets of models resulting from contracting a static law from a given set of models:

**Definition 3.10** *Let* $\mathcal{M}$ *be a set of models, and* $\varphi$ *a static law. Then*

$$\mathcal{M}_\varphi^- = \{\mathcal{M}' : \mathcal{M}' = \mathcal{M} \cup \{\mathcal{M}'\}, \mathcal{M}' \in contract(\mathcal{M}, \varphi), \mathcal{M} \in \mathcal{M}\}$$

In our scenario example, if the initial set of models is $\mathcal{M} = \{\mathcal{M}\}$, where $\mathcal{M}$ is the model in Figure 4, then contracting the static law *coffee* $\rightarrow$ *hot* from $\mathcal{M}$ would give us the resulting new set of models $\mathcal{M}_{coffee \rightarrow hot}^- = \{\mathcal{M} \cup \{\mathcal{M}_1'\}, \mathcal{M} \cup \{\mathcal{M}_2'\}\}$, where each $\mathcal{M}_i'$ is as depicted in Figure 9 below.

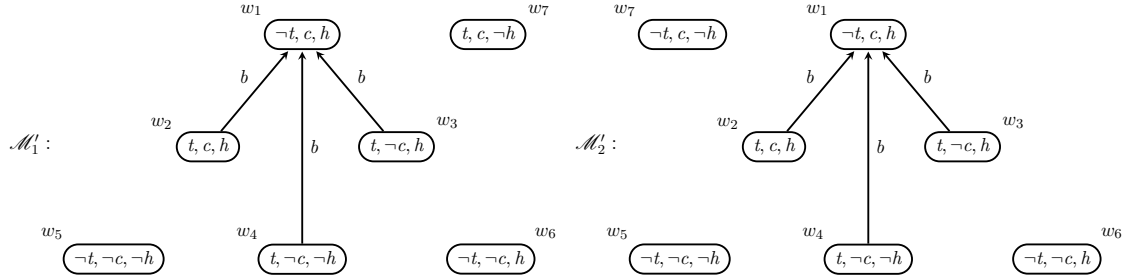

Figure 9: Models resulting from contracting *coffee* $\rightarrow$ *hot* in the model $\mathcal{M}$ of Figure 4.

Notice that by not modifying the accessibility relation all the effect laws which are true in the original model $\mathcal{M}$ are preserved in the resulting models. This is ensured by $[buy]\bot$ being true in the new world $w_7$.

It is only some executability laws that are potentially lost, due to the cautiousness of our approach. For instance, in $\mathcal{M}_1'$ above, it is no longer the case that *token* $\rightarrow \langle buy \rangle \top$ is true, since now there is a world, namely $w_7$, which does not satisfy it anymore. (In $\mathcal{M}_2'$ this executability law is still true in every possible world.)

It is worth point out, however, how our approach is indeed in line with intuition: when learning that a new state is now possible, we do not necessarily know all the behavior of the actions in the new added state. We may expect some action laws to hold in the new world (see end of Section 5.3), but, with the information we dispose, not touching the accessibility relation is the safest way of contracting static laws (cf. Section 4.2 below).

# 4. Interlude

Before presenting the algorithmic counterpart of our action theory change operators, in this section we discuss alternatives to some of our technical constructions. We point out the issues that such alternatives would raise. We also provide more justifications for some of the design choices that have been made in the previous sections.

## 4.1 Other Distance Notions

Here we have defined and used a model distance which is based on the symmetric difference between sets (Definition 2.11). This distance is an extension to Kripke structures of Winslett's (1988) notion of closeness between propositional interpretations in the Possible Models Approach (PMA). Instead of it, however, we could have considered other distance notions as well, like Dalal's (1988) distance, Hamming distance (1950), or weighted distance. Due to space limitations, we do not develop a through comparison among all these distances here. (For more details, the reader may want to refer to Schlechta's 2004 book.) We nevertheless do show that with a cardinality-based distance, for example, we may not always get the intended result.

Let $card(X)$ denote the number of elements in set $X$. Then suppose that our closeness between $\mathsf{K}_n$-models was defined as follows:

**Definition 4.1 (Cardinality-based Closeness between $\mathsf{K}_n$-Models)** *Let $\mathscr{M} = \langle W, R \rangle$ be a model. Then $\mathscr{M}' = \langle W', R' \rangle$ is at least as close to $\mathscr{M}$ as $\mathscr{M}'' = \langle W'', R'' \rangle$, noted $\mathscr{M}' \leq_{\mathscr{M}} \mathscr{M}''$, if and only if*

- *either $card(W \dot{-} W') \leq card(W \dot{-} W'')$;*

- *or $card(W \dot{-} W') = card(W \dot{-} W'')$ and $card(R \dot{-} R') \leq card(R \dot{-} R'')$.*

Such a notion of distance is closely related to Dalal's (1988) closeness.

Because when contracting a static law $\varphi$ from a model $\mathscr{M}$ we usually add one new possible world, it is easy to see that with this cardinality-based distance we get the same result in $contract(\mathscr{M}, \varphi)$ as with the distance from Definition 2.11.

When it comes to the contraction of action laws, and then changing the accessibility relations, however, this cardinality-based distance does not seem to fit with the intuitions. To witness, consider the model $\mathscr{M}$ in Figure 10, in which the law $p_1 \to \langle a \rangle \top$ is true.

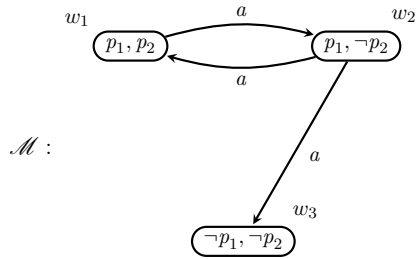

Figure 10: A model of the executability law $p_1 \to \langle a \rangle \top$.

Then, the models resulting from contraction of $p_1 \to \langle a \rangle \top$ in the model $\mathscr{M}$ will be $\mathscr{M}^-_{p_1 \to \langle a \rangle \top} = \{\mathscr{M}', \mathscr{M}''\}$, where $\mathscr{M}'$ and $\mathscr{M}''$ are as depicted in Figure 11.

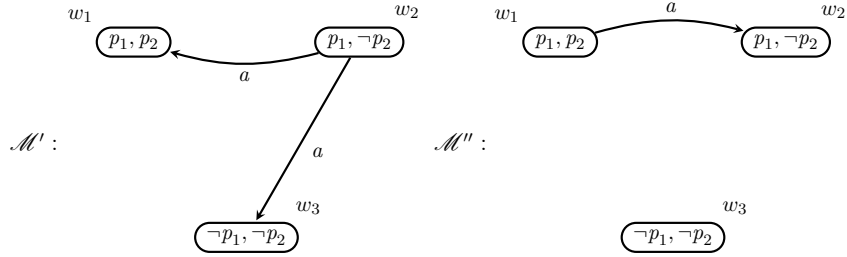

Figure 11: Models resulting from contracting $p_1 \to \langle a \rangle \top$ in the model $\mathscr{M}$ of Figure 10.

Note that $\mathscr{M}''$ is an intended contracted model: $\not\models^{\mathscr{M}''} p_1 \to \langle a \rangle \top$. However, with the cardinality-based distance above we will get $\{\mathscr{M}\}^-_{p_1 \to \langle a \rangle \top} = \{\{\mathscr{M}, \mathscr{M}'\}\}$. We do not have $\{\mathscr{M}, \mathscr{M}''\}$ in the result since $\mathscr{M}' \leq_{\mathscr{M}} \mathscr{M}''$: in $\mathscr{M}'$ only *one* transition has been removed, while in $\mathscr{M}''$ *two*.

## 4.2 Minimal Change v. Cautiousness

As usually done in the literature on classical belief revision, when defining a (traditional) theory change operator one must always make the fundamental decision which of two opposing principles should be the guiding one: that of minimizing change, which leads to strong modified theories, versus that of cautious change, which leads to weak theories. In this regard, one might argue that our action theory change operators are incoherent. That is because we adopt the first principle for the contraction of effect and executability laws, but then the latter principle for contraction of static laws.[3]

It turns out, however, that this view is debatable. From a different perspective one can think of our three operators as being coherent in the following sense: all of them perform a version of *maxichoice*, namely the addition of precisely a *single* model to the original models of the theory.[4]

In any case, in the sequel we give a justification for the behavior of our operators and show that there can be no such an operator for contraction of static laws that is not cautious while being coherent with the operators for contraction of effect and executability laws. (We say that an operator for static law contraction is *coherent* with respect to our operators for contraction of effect and executability laws if it also performs minimal change with respect to the other types of laws, i.e., if it preserves effect and executability laws.)

Where does the claimed 'incoherence' come from? Here our contention is that it is inherent to the problem of action theory change itself, and not a flaw of our definitions. The justification is as follows. Remembering the intuitions for our semantic constructions, it is easy to see that for the contraction of executability laws knowledge about some action's feasibility (the transitions) is removed and only that. For the contraction of effect laws, a piece of knowledge is also added (the new transition), but notice that this one is 'guided' by some given *concrete* extra information, namely the $\neg\psi$ effect that we want to allow.

---

3. We thank an anonymous referee for having pointed this out.
4. We thank another anonymous referee for having pointed this out.

Now, for the contraction of static laws, notice that no extra information whatsoever is given about the new possible state which could guide the addition of some knowledge about the feasibility of an action. The only thing that we know is that the new world should exist. Nothing more is said about whether there should be any transition leaving it or arriving at it at all. This is a property of the problem per se: the problem of removing a static law does not mention executabilities, and it is just reflected by our operator.

Therefore, such an 'incoherence' is already in the problem, and as such it is not surprising to find it again in the proposed operators. These are designed to do what they are allowed to do given the constraints of the problem. Should we have more information in our hands regarding the new added state, a coherent version of the corresponding operator would have been defined. (See the discussion in Section 9 for a comparison with Eiter et al.'s 2005 constraint-based method for update of action theories.)

**Proposition 4.1** *There is no minimal change operator for static law contraction that is coherent with our operators for contraction of effect and executability laws.*

**Proof:** Suppose that we have a minimal change based (non-cautious) contraction operator for static laws that is coherent with the other operators. This operator must be such that when contracting $\varphi \in \mathfrak{Fml}$ only formulas of the type of $\varphi$ are removed (otherwise it is not coherent with the other operators). This means that both effect and executability laws should be preserved. In particular, this operator is coherent in this respect when contracting the formula $p_1 \to \neg p_2$ from model $\mathscr{M}$ in Figure 12 below.

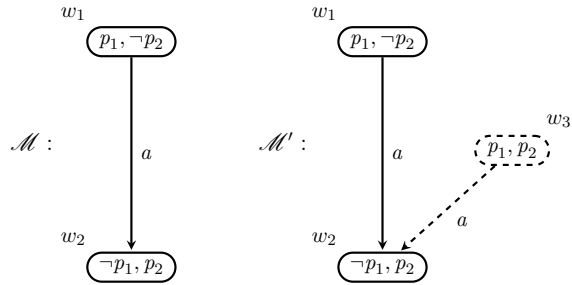

Figure 12: Adding a transition from a new added world in the alternative semantics to static law contraction. $\mathscr{M}$ denotes the original model, while $\mathscr{M}'$ shows the new added world and a candidate transition to add to $R_a$.

Following the intuition about contraction of Boolean formulas, a new world, viz. the valuation $\{p_1, p_2\}$, is added to $W$ in $\mathscr{M}$. Because the operator in question is non-cautious, a transition should also be added from the new added world $\{p_1, p_2\}$ in $\mathscr{M}$, in order to preserve the executability law $p_1 \to \langle a \rangle \top$. Also because the operator is non-cautious, the effect law $p_1 \to [a]\neg p_1$ should be preserved. Hence, such a new transition should point neither to world $\{p_1, \neg p_2\}$ nor to $\{p_1, p_2\}$ itself. Now, if we direct the new transition to $\{\neg p_1, p_2\}$ (the only world that is left), we get the model $\mathscr{M}'$ in Figure 12.

Observe that $\models^{\mathcal{M}} (\neg p_1 \vee p_2) \rightarrow [a]p_1$. However, $\not\models^{\mathcal{M}'} (\neg p_1 \vee p_2) \rightarrow [a]p_1$: the operator makes us lose an effect law! This means that it is not coherent. In order for us to keep this effect law, the only option is not to direct the new transition to $\{\neg p_1, p_2\}$. But then, no transition is added at all: the operator is cautious! Hence there is no such an operator for static law contraction that is based on minimal change and is coherent with the operators for the other laws. □

The result above supports our contention that we cannot have a coherent set of minimal change operators for action theory contraction. This is a general result and it holds not only for modal-based approaches like ours, but it applies to any framework for reasoning about actions which is based on transition systems and which also allows for the three types of laws that we consider here.

Furthermore, the result also illustrates well the difference between action theory change and classical belief change. To witness, even though contraction of static laws amounts to propositional contraction of Boolean formulas, it remains a special case of the latter. The reason is that when contracting static laws one always asks "what happens to the laws of other types?", a question that is not asked in classical propositional contraction for the obvious reason that there simply there are no other *types* of formulas.

## 5. Syntactic Operators for Contraction of Laws

Once having given a semantic construction for action theory change, we now turn our attention to the definition of syntactic operators for changing sets of formulas describing a dynamic domain.

As Nebel (1989) says, "[...] finite bases usually represent [...] laws, and when we are forced to change the theory we would like to stay as close as possible to the original [...] base." Hence, besides the definition of syntactical operators, we should also guarantee that they perform minimal change at the theory level. By that we mean that the resulting theory should of course not entail the law we want to contract the theory with, and it should also preserve as much of the previous knowledge as possible when performing syntactical manipulations on the laws in the original theory. Ideally, from the knowledge engineer's perspective, the modified theory should also keep a certain degree of resemblance with the original one: the resulting laws should be slight modifications of the relevant ones in the original action theory.

By $\mathcal{T}_\Phi^-$ we denote in the sequel the result of contracting a law $\Phi$ from the set of laws $\mathcal{T}$.

### 5.1 Contracting Executability Laws

For the case of contracting an executability law $\varphi \rightarrow \langle a \rangle \top$ from an action theory, first we have to ensure that action $a$ keeps its executability state in all those contexts where the antecedent $\neg\varphi$ holds, if that is the case. We achieve that by *strengthening* the antecedents of the relevant executability laws. Second, in order to get minimality, we must make $a$ executable in *some* contexts where $\varphi$ is true, viz. all $\varphi$-worlds but one. Since there are possibly many different alternatives for that, this means that we can have several action theories as outcome. Algorithm 1 gives a syntactical operator to achieve this.

It can be easily checked that Algorithm 1 always terminates: the input action theory $\mathcal{T}$ is always finite; from finiteness of $\mathfrak{Prop}$ follows that of $\overline{atm(\pi)}$, for any $\pi \in IP(\mathcal{S} \wedge \varphi)$. Moreover, the entailment problem of multimodal $\mathsf{K}$ is decidable (Harel et al., 2000), as is that of classical propositional logic. Therefore contracting executability laws is decidable.

---

**Algorithm 1**: Contraction of an Executability Law

---

    **Input**: $\mathcal{T}$, $\varphi \to \langle a \rangle \top$

    **Output**: $\mathcal{T}^-_{\varphi \to \langle a \rangle \top}$ `/* set of theories output to the knowledge engineer */`

**1 begin**

**2**      $\mathcal{T}^-_{\varphi \to \langle a \rangle \top} := \emptyset$

**3**      **if** $\mathcal{T} \models_{\overline{\mathsf{K}}_n} \varphi \to \langle a \rangle \top$ *and* $\mathcal{S} \not\models_{\mathsf{CPL}} \neg \varphi$ **then**

**4**          **foreach** $\pi \in IP(\mathcal{S} \wedge \varphi)$ **do**

**5**              **forall** $A \subseteq \overline{atm(\pi)}$ **do**

**6**                  $\varphi_A := \bigwedge_{\substack{p_i \in \overline{atm(\pi)} \\ p_i \in A}} p_i \wedge \bigwedge_{\substack{p_i \in \overline{atm(\pi)} \\ p_i \notin A}} \neg p_i$ `/* extend` $\pi$ `to a valuation */`

**7**                  **if** $\mathcal{S} \not\models_{\mathsf{CPL}} (\pi \wedge \varphi_A) \to \bot$ **then**      `/* it is an allowed state */`

                     `/* construct a theory that is weaker for that state */`

**8**                      $\mathcal{T}' := (\mathcal{T} \setminus \mathcal{X}_a) \cup \{(\varphi_i \wedge \neg(\pi \wedge \varphi_A)) \to \langle a \rangle \top : \varphi_i \to \langle a \rangle \top \in \mathcal{X}_a\}$

**9**                      $\mathcal{T}^-_{\varphi \to \langle a \rangle \top} := \mathcal{T}^-_{\varphi \to \langle a \rangle \top} \cup \{\mathcal{T}'\}$

**10**      **else**

**11**          $\mathcal{T}^-_{\varphi \to \langle a \rangle \top} := \{\mathcal{T}\}$

**12**      **return** $\mathcal{T}^-_{\varphi \to \langle a \rangle \top}$

**13 end**

---

It is straightforward to see that Algorithm 1 above can be adapted to Situation Calculus action theories as well. The crucial point however would be its termination, since entailment in the Situation Calculus is in general undecidable.

In our running example, contracting the executability law $token \to \langle buy \rangle \top$ from the action theory $\mathcal{T}$ would give us $\mathcal{T}^-_{token \to \langle buy \rangle \top} = \{\mathcal{T}'_1, \mathcal{T}'_2, \mathcal{T}'_3\}$, where:

$$\mathcal{T}'_1 = \left\{ \begin{array}{c} coffee \to hot, \neg coffee \to [buy] coffee, \\ token \to [buy] \neg token, \neg token \to [buy] \bot, \\ coffee \to [buy] coffee, hot \to [buy] hot, \\ (token \wedge (\neg coffee \vee \neg hot)) \to \langle buy \rangle \top \end{array} \right\}$$

$$\mathcal{T}'_2 = \left\{ \begin{array}{c} coffee \to hot, \neg coffee \to [buy] coffee, \\ token \to [buy] \neg token, \neg token \to [buy] \bot, \\ coffee \to [buy] coffee, hot \to [buy] hot, \\ (token \wedge (coffee \vee \neg hot)) \to \langle buy \rangle \top \end{array} \right\}$$

$$\mathcal{T}_3' = \left\{ \begin{array}{c} coffee \rightarrow hot, \neg coffee \rightarrow [buy]coffee, \\ token \rightarrow [buy]\neg token, \neg token \rightarrow [buy]\bot, \\ coffee \rightarrow [buy]coffee, hot \rightarrow [buy]hot, \\ (token \wedge (coffee \vee hot)) \rightarrow \langle buy \rangle \top \end{array} \right\}$$

Now all the knowledge engineer has to do is choose which theory is more in line with her intuitions and implement the required changes (cf. Figure 5).

## 5.2 Contracting Effect Laws

When contracting an effect law $\varphi \rightarrow [a]\psi$ from an action theory $\mathcal{T}$, intuitively we should contract those effect laws that preclude $\neg\psi$ in target worlds. In order to cope with minimality, we must change only those laws that are relevant to the unwanted $\varphi \rightarrow [a]\psi$.

Let $(\mathcal{E}_a^{\varphi,\psi})_1, \ldots, (\mathcal{E}_a^{\varphi,\psi})_n$ denote minimal subsets (with respect to set inclusion) of $\mathcal{E}_a$ such that $\mathcal{S}, (\mathcal{E}_a^{\varphi,\psi})_i \models_{\overline{\mathsf{R}}_n} \varphi \rightarrow [a]\psi$, for $1 \le i \le n$. In other words, each $(\mathcal{E}_a^{\varphi,\psi})_i$ is a *support set* for the effect law $\varphi \rightarrow [a]\psi$ in $\mathcal{T}$. To make a parallel with the terminology usually adopted in the belief change community, we shall see each $(\mathcal{E}_a^{\varphi,\psi})_i$ as a special type of *kernel* (Hansson, 1994) for the formula $\varphi \rightarrow [a]\psi$.

According to Herzig and Varzinczak (2007), given any action theory one can always ensure that at least one support set for $\varphi \rightarrow [a]\psi$ exists. Now let

$$\mathcal{E}_a^- = \bigcup_{1 \le i \le n} (\mathcal{E}_a^{\varphi,\psi})_i$$

The laws in $\mathcal{E}_a^-$ will serve as guidelines to get rid of $[a]\psi$ in each $\varphi$-world allowed by the theory $\mathcal{T}$: they are the effect laws to be weakened to allow for $\langle a \rangle \neg\psi$ in some $\varphi$-contexts. This resembles classical kernel contraction (Hansson, 1994): finding minimal sets implying a formula and changing them. A crucial difference, however, is that instead of completely removing a formula from each kernel, what we do here is *weaken* the laws.

When modifying the support sets, the first thing we must do is to ensure that action $a$ still has effect $\psi$ in all those contexts in which $\varphi$ does not hold, if that is the case. This means we shall weaken the laws in $\mathcal{E}_a^{\varphi,\psi}$ specializing them to $\neg\varphi$. Now, we need to preserve all old effects in all $\varphi$-worlds but one. To achieve that we specialize the above laws to each possible valuation (maximal consistent conjunction of literals) satisfying $\varphi$ but one. Then, in the left $\varphi$-valuation, we must ensure that action $a$ has either its old effects or $\neg\psi$ as outcome. We achieve that by weakening the *consequent* of the laws in $\mathcal{E}_a^-$. Finally, in order to get minimal change, we must ensure that all literals in this $\varphi$-valuation that are not forced to change in $\neg\psi$-worlds should be preserved. We do this by stating an effect law of the form $(\varphi_k \wedge \ell) \rightarrow [a](\psi \vee \ell)$, where $\varphi_k$ is the above $\varphi$-valuation. The reason this is needed is clear: there can be several $\neg\psi$-valuations, and as far as we want at most one to be reachable from the $\varphi_k$-world, we should force it to be the one whose difference to this $\varphi_k$-valuation is minimal.

In Situation Calculus terms, all these syntactical operations would correspond to strengthening the right-hand side of the relevant successor state axioms and/or weakening their left-hand side. Alternatively, the same can be done with the original effect axioms, then recompiling them again into new successor state axioms afterwards.

The output of the operations described above will be a set of action theories which will be output to the knowledge engineer. Algorithm 2 below gives the operator.

---

**Algorithm 2**: Contraction of an Effect Law

**Input**: $\mathcal{T}$, $\varphi \to [a]\psi$

**Output**: $\mathcal{T}^{-}_{\varphi \to [a]\psi}$ /* set of theories output to the knowledge engineer */

**1 begin**

**2**    $\mathcal{T}^{-}_{\varphi \to [a]\psi} := \emptyset$

**3**    **if** $\mathcal{T} \models_{\overline{\mathsf{K}}_n} \varphi \to [a]\psi$ *and* $\mathcal{S} \not\models_{\mathsf{CPL}} \neg\varphi$ **then**

**4**      **foreach** $\pi \in IP(\mathcal{S} \wedge \varphi)$ **do**

**5**        **forall** $A \subseteq \overline{atm(\pi)}$ **do**

**6**          $\varphi_A := \bigwedge_{\substack{p_i \in \overline{atm(\pi)} \\ p_i \in A}} p_i \wedge \bigwedge_{\substack{p_i \in \overline{atm(\pi)} \\ p_i \notin A}} \neg p_i$ /* extend $\pi$ to a valuation */

**7**          **if** $\mathcal{S} \not\models_{\mathsf{CPL}} (\pi \wedge \varphi_A) \to \bot$ **then**       /* it is an allowed state */

**8**            **foreach** $\pi' \in IP(\mathcal{S} \wedge \neg\psi)$ **do**

**9**              $\mathcal{T}' := \mathcal{T} \setminus \mathcal{E}^{-}_a$ /* the support sets will be weakened */

**10**              $\mathcal{T}' := \mathcal{T}' \cup \{(\varphi_i \wedge \neg(\pi \wedge \varphi_A)) \to [a]\psi_i : \varphi_i \to [a]\psi_i \in \mathcal{E}^{-}_a\}$

             /* allow for $\neg\psi$ after $a$ in this state */

**11**              $\mathcal{T}' := \mathcal{T}' \cup \{(\varphi_i \wedge \pi \wedge \varphi_A) \to [a](\psi_i \vee \pi') : \varphi_i \to [a]\psi_i \in \mathcal{E}^{-}_a\}$

**12**              **forall** $L \subseteq \mathfrak{Lit}$ **do**

**13**                **if** $\mathcal{S} \models_{\mathsf{CPL}} (\pi \wedge \varphi_A) \to \bigwedge_{\ell \in L} \ell$ **and** $\mathcal{S} \not\models_{\mathsf{CPL}} (\pi' \wedge \bigwedge_{\ell \in L} \ell) \to \bot$ **then**

**14**                  **foreach** $\ell \in L$ **do**

**15**                    **if** $\mathcal{T} \not\models_{\overline{\mathsf{K}}_n} (\pi \wedge \varphi_A \wedge \ell) \to [a]\neg\ell$ **or** $\ell \in \pi'$ **then**

**16**                      $\mathcal{T}' := \mathcal{T}' \cup \{(\pi \wedge \varphi_A \wedge \ell) \to [a](\psi \vee \ell)\}$

**17**              $\mathcal{T}^{-}_{\varphi \to [a]\psi} := \mathcal{T}^{-}_{\varphi \to [a]\psi} \cup \{\mathcal{T}'\}$

**18**    **else**

**19**      $\mathcal{T}^{-}_{\varphi \to [a]\psi} := \{\mathcal{T}\}$

**20**    **return** $\mathcal{T}^{-}_{\varphi \to [a]\psi}$

**21 end**

---

Again, from the finiteness of the action theory $\mathcal{T}$ and that of $\overline{atm(\pi)}$, for any $\pi \in IP(\mathcal{S} \wedge \varphi)$, and from the decidability of multimodal $\mathsf{K}$ (Harel et al., 2000) as well as that of classical propositional logic, it can be easily verified that Algorithm 2 always terminates.

Therefore, contracting effect laws is decidable. Of course, the complexity of computing all the support sets as well as the prime implicants is quite high (see Section 5.4 later on for a discussion on this matter).

For an example of execution of Algorithm 2, let us suppose that we want to contract the effect law $token \rightarrow [buy]hot$ from the action theory $\mathcal{T}$ of our running example. First we have to compute the support sets for $token \rightarrow [buy]hot$ in $\mathcal{T}$ (i.e., the minimal subsets of $\mathcal{E}_{buy}$ which together with $\mathcal{S}$ entail $token \rightarrow [buy]hot$). These are the following:

$$(\mathcal{E}_{buy}^{token,hot})_1 = \left\{ \begin{array}{c} coffee \rightarrow [buy]coffee, \\ \neg coffee \rightarrow [buy]coffee \end{array} \right\}$$

$$(\mathcal{E}_{buy}^{token,hot})_2 = \left\{ \begin{array}{c} hot \rightarrow [buy]hot, \\ \neg coffee \rightarrow [buy]coffee \end{array} \right\}$$

Now for each possible context in which the antecedent $token$ is the case, we have to weaken the effect laws in $\mathcal{E}_{buy}^- = (\mathcal{E}_{buy}^{token,hot})_1 \cup (\mathcal{E}_{buy}^{token,hot})_2$. Since $\mathcal{S} = \{coffee \rightarrow hot\}$, such contexts are $token \wedge coffee \wedge hot$, $token \wedge \neg coffee \wedge \neg hot$ and $token \wedge \neg coffee \wedge hot$.

For $token \wedge coffee \wedge hot$: Algorithm 2 replaces in $\mathcal{T}$ the laws from $\mathcal{E}_{buy}^-$ with

$$\left\{ \begin{array}{c} (coffee \wedge \neg(token \wedge coffee \wedge hot)) \rightarrow [buy]coffee, \\ (hot \wedge \neg(token \wedge coffee \wedge hot)) \rightarrow [buy]hot, \\ (\neg coffee \wedge \neg(token \wedge coffee \wedge hot)) \rightarrow [buy]coffee \end{array} \right\}$$

so that we preserve their effects in all possible contexts but $token \wedge coffee \wedge hot$. Now, in order to preserve some effects in $token \wedge coffee \wedge hot$-contexts while allowing for reachable $\neg hot$-worlds, the algorithm adds the laws:

$$\left\{ \begin{array}{c} (token \wedge coffee \wedge hot) \rightarrow [buy](coffee \vee \neg hot), \\ (token \wedge coffee \wedge hot) \rightarrow [buy](hot \vee \neg coffee) \end{array} \right\}$$

Now, we search all possible combinations of laws from $\mathcal{E}_{buy}$ that apply on $token \wedge coffee \wedge hot$ contexts and find $token \rightarrow [buy]\neg token$. Because $\neg token$ must be true after every execution of action $buy$, we do not state the law $(token \wedge coffee \wedge hot) \rightarrow [buy](hot \vee token)$, and end up with the following theory:

$$\mathcal{T}_1' = \left\{ \begin{array}{c} coffee \rightarrow hot, token \rightarrow \langle buy \rangle \top, \\ token \rightarrow [buy]\neg token, \neg token \rightarrow [buy]\bot, \\ (coffee \wedge \neg(token \wedge coffee \wedge hot)) \rightarrow [buy]coffee, \\ (hot \wedge \neg(token \wedge coffee \wedge hot)) \rightarrow [buy]hot, \\ (\neg coffee \wedge \neg(token \wedge coffee \wedge hot)) \rightarrow [buy]coffee, \\ (token \wedge coffee \wedge hot) \rightarrow [buy](coffee \vee \neg hot), \\ (token \wedge coffee \wedge hot) \rightarrow [buy](hot \vee \neg coffee) \end{array} \right\}$$

On the other hand, if in our language we also had an atom $p$ with the same theory $\mathcal{T}$, then we should have added a law $(token \wedge coffee \wedge hot \wedge p) \rightarrow [buy](hot \vee p)$ to meet minimal change by preserving effects that are not relevant to $\neg \psi$ (cf. Definition 3.4).

The execution for contexts $token \wedge \neg coffee \wedge \neg hot$ and $token \wedge \neg coffee \wedge hot$ are analogous and the algorithm ends with $\mathcal{T}^{-}_{token \to [buy]hot} = \{\mathcal{T}'_1, \mathcal{T}'_2, \mathcal{T}'_3\}$, where:

$$\mathcal{T}'_2 = \left\{ \begin{array}{c} coffee \to hot, token \to \langle buy \rangle \top, \\ token \to [buy]\neg token, \neg token \to [buy]\bot, \\ (coffee \wedge \neg(token \wedge \neg coffee \wedge \neg hot)) \to [buy]coffee, \\ (hot \wedge \neg(token \wedge \neg coffee \wedge \neg hot)) \to [buy]hot, \\ (\neg coffee \wedge \neg(token \wedge \neg coffee \wedge \neg hot)) \to [buy]coffee, \\ (token \wedge \neg coffee \wedge \neg hot) \to [buy](coffee \vee \neg hot) \end{array} \right\}$$

$$\mathcal{T}'_3 = \left\{ \begin{array}{c} coffee \to hot, token \to \langle buy \rangle \top, \\ token \to [buy]\neg token, \neg token \to [buy]\bot, \\ (coffee \wedge \neg(token \wedge \neg coffee \wedge hot)) \to [buy]coffee, \\ (hot \wedge \neg(token \wedge \neg coffee \wedge hot)) \to [buy]hot, \\ (\neg coffee \wedge \neg(token \wedge \neg coffee \wedge hot)) \to [buy]coffee, \\ (token \wedge \neg coffee \wedge hot) \to [buy](hot \vee \neg coffee), \\ (token \wedge \neg coffee \wedge hot) \to [buy](coffee \vee \neg hot) \end{array} \right\}$$

Looking at Figure 8, we can see the correspondence between these theories and their respective models. It is now up to the knowledge engineer to look at these action theories and pick up the one corresponding to her expectations.

## 5.3 Contracting Static Laws

Finally, in order to contract a static law from a theory, we can use any contraction/erasure operator $\ominus$ for classical logic that is available in the literature. Because contracting static laws means *admitting* new possible states (cf. the semantics), just modifying the set $\mathcal{S}$ of static laws may not be enough for the multimodal logic case. However, since in general we do not necessarily know the behavior of the actions in a new discovered state of the world, a careful approach is to change the theory so that all action laws remain the same in the contexts where the contracted law is the case. (The reader is invited to see that in the Situation Calculus by allowing a new situation to exist one may need to change the precondition axioms as well, which means that the problem here described is independent of the logical formalism chosen.)

In our scenario example, if in contracting the static law $coffee \to hot$ the knowledge engineer is not really sure whether action $buy$ is still executable or not, then she should weaken the set of executability laws specializing them to the context $coffee \to hot$, and make $buy$ a priori inexecutable in all $\neg(coffee \to hot)$-contexts. It is worth noting that this is in line with the assumption commonly made in the RAC community according to which executability laws are by and large much more likely to be incorrect right from the beginning (Shanahan, 1997). Therefore extrapolating them to previously unknown states might (and in all probability will) result in the propagation of errors and, even worse, the loss of effect laws (remember the discussion in Sections 3.3 and 4.2). The operator given in Algorithm 3 formalizes this.

---

**Algorithm 3:** Contraction of a Static Law

---
**Input:** $\mathcal{T}, \varphi$

**Output:** $\mathcal{T}_\varphi^-$ /* set of theories output to the knowledge engineer */

**1 begin**

**2**    $\mathcal{T}_\varphi^- := \emptyset$

**3**    **if** $\mathcal{S} \not\models_{\overline{\mathsf{CPL}}} \varphi$ **then**

       /* call classical contraction $\mathcal{S} \ominus \varphi$ of $\mathcal{S}$ with $\varphi$ */

**4**      **foreach** $\mathcal{S}^- \in \mathcal{S} \ominus \varphi$ **do**

         /* build a theory preserving executability in old states */

**5**        $\mathcal{T}' := ((\mathcal{T} \setminus \mathcal{S}) \cup \mathcal{S}^-) \setminus \mathcal{X}_a$

**6**        $\mathcal{T}' := \mathcal{T}' \cup \{(\varphi_i \wedge \varphi) \to \langle a \rangle \top : \varphi_i \to \langle a \rangle \top \in \mathcal{X}_a\} \cup \{\neg \varphi \to [a] \bot\}$

**7**        $\mathcal{T}_\varphi^- := \mathcal{T}_\varphi^- \cup \{\mathcal{T}'\}$

**8**    **else**

**9**      $\mathcal{T}_\varphi^- := \{\mathcal{T}\}$

**10**    **return** $\mathcal{T}_\varphi^-$

**11 end**

---

In our running coffee example, contracting the static law *coffee* $\to$ *hot* from the action theory $\mathcal{T}$ produces $\mathcal{T}_{coffee \to hot}^- = \{\mathcal{T}_1', \mathcal{T}_2'\}$, where

$$\mathcal{T}_1' = \left\{ \begin{array}{c} \neg(\neg token \wedge coffee \wedge \neg hot), \\ (token \wedge coffee \to hot) \to \langle buy \rangle \top, \\ \neg coffee \to [buy] coffee, token \to [buy] \neg token, \\ \neg token \to [buy] \bot, coffee \to [buy] coffee, \\ hot \to [buy] hot, (coffee \wedge \neg hot) \to [buy] \bot \end{array} \right\}$$

$$\mathcal{T}_2' = \left\{ \begin{array}{c} \neg(token \wedge coffee \wedge \neg hot), \\ (token \wedge coffee \to hot) \to \langle buy \rangle \top, \\ \neg coffee \to [buy] coffee, token \to [buy] \neg token, \\ \neg token \to [buy] \bot, coffee \to [buy] coffee, \\ hot \to [buy] hot, (coffee \wedge \neg hot) \to [buy] \bot \end{array} \right\}$$

Observe that the effect laws are not affected at all by the change: as far as we do not pronounce ourselves about the executability of some action in the new added world, all the effect laws remain true in it.

If the knowledge engineer is not happy with $(coffee \wedge \neg hot) \to [buy] \bot$, she can contract this formula from the theory using Algorithm 2. Ideally, besides stating that *buy* is executable in the context *coffee* $\wedge$ $\neg hot$, we should want to specify its outcome in this context as well. For example, we could want $(coffee \wedge \neg hot) \to \langle buy \rangle hot$ to be true in the result. This requires theory *revision*. See Section 8 for the semantics of such an operation.

### 5.4 Complexity Issues

While terminating, our algorithms come with a considerable computational cost: the $\mathsf{K}_n$-entailment tests with global axioms in the beginning of the algorithms and inside the loops are known to be EXPTIME-complete (Harel et al., 2000). The computation of all possible contexts allowed by the theory, namely $\bigwedge_{\substack{p_i \in \overline{atm(\pi)} \\ p_i \in A}} p_i \wedge \bigwedge_{\substack{p_i \in \overline{atm(\pi)} \\ p_i \notin A}} \neg p_i$, for all $A \subseteq \overline{atm(\pi)}$ and all $\pi \in IP(\mathcal{S} \wedge \varphi)$, is clearly exponential. Moreover, the computation of prime implicants $IP(.)$ might result in exponential growth (Marquis, 2000).

Given that theory change can be carried out offline, from the perspective of the knowledge engineer what is more important is the complexity of the size of the computed contracted theories: the number of formulas as well as the length of the modified ones. This plays an important role when deciding among several output theories which one corresponds to the knowledge engineer's expectations. In that matter, whereas the length of new added formulas may increase exponentially, with respect to the number of laws our results are positive: the size of the computed contracted theories is *linear* in the size of the original action theory. (Remember that $card(X)$ denotes the number of elements in set $X$.)

**Proposition 5.1** *Let $\mathcal{T}$ be an action theory, $\varphi \rightarrow \langle a \rangle \top$ an executability law, and $\mathcal{T}' \in \mathcal{T}^{-}_{\varphi \rightarrow \langle a \rangle \top}$. Then $card(\mathcal{T}') = card(\mathcal{T})$.*

**Proof:** If $\mathcal{T} \not\models_{\overline{\mathsf{K}}_n} \varphi \rightarrow \langle a \rangle \top$, then $\mathcal{T}^{-}_{\varphi \rightarrow \langle a \rangle \top} = \{\mathcal{T}\}$, and then $\mathcal{T}' = \mathcal{T}$, from which the result follows. Suppose $\mathcal{T} \models_{\overline{\mathsf{K}}_n} \varphi \rightarrow \langle a \rangle \top$ is the case. Then $\mathcal{T}'$ is such that $\mathcal{T}' = (\mathcal{T} \setminus \mathcal{X}_a) \cup \mathcal{X}_a{}'$, where $\mathcal{X}_a{}'$ is obtained from $\mathcal{X}_a$ in such a way that $(\varphi_i \wedge \varphi') \rightarrow \langle a \rangle \top \in \mathcal{X}_a{}'$ if and only if $\varphi_i \rightarrow \langle a \rangle \top \in \mathcal{X}_a$, for a fixed $\varphi'$. From this it follows $card(\mathcal{X}_a{}') = card(\mathcal{X}_a)$. Now, $card((\mathcal{T} \setminus \mathcal{X}_a) \cup \mathcal{X}_a{}') = card(\mathcal{T} \setminus \mathcal{X}_a) + card(\mathcal{X}_a{}') - card((\mathcal{T} \setminus \mathcal{X}_a) \cap \mathcal{X}_a{}') = card(\mathcal{T}) - card(\mathcal{X}_a) + card(\mathcal{X}_a{}') - card(\emptyset) = card(\mathcal{T}) - card(\mathcal{X}_a) + card(\mathcal{X}_a) - 0 = card(\mathcal{T})$. $\quad\square$

**Proposition 5.2** *Let $\mathcal{T}$ be an action theory, $\varphi \rightarrow [a]\psi$ an effect law, and $\mathcal{T}' \in \mathcal{T}^{-}_{\varphi \rightarrow [a]\psi}$. Then $card(\mathcal{T}') \leq card(\mathcal{T}) + card(\mathcal{E}_a^{-}) + card(\mathfrak{Lit})$.*

**Proof:** If $\mathcal{T} \not\models_{\overline{\mathsf{K}}_n} \varphi \rightarrow [a]\psi$, then $\mathcal{T}^{-}_{\varphi \rightarrow [a]\psi} = \{\mathcal{T}\}$, and then $\mathcal{T}' = \mathcal{T}$, from what we get $card(\mathcal{T}') = card(\mathcal{T})$. Since $card(\mathcal{T}) \leq card(\mathcal{T}) + card(\mathcal{E}_a^{-}) + card(\mathfrak{Lit})$, the result follows. Suppose that $\mathcal{T} \models_{\overline{\mathsf{K}}_n} \varphi \rightarrow [a]\psi$ is the case. Then $\mathcal{T}' = (\mathcal{T} \setminus \mathcal{E}_a^{-}) \cup \mathcal{E}_a{}' \cup \mathcal{E}_a{}'' \cup \mathcal{F}_a$, where:

- $\mathcal{E}_a{}'$ and $\mathcal{E}_a{}''$ are both obtained from $\mathcal{E}_a^{-}$ in such a way that $(\varphi_i \wedge \neg\varphi') \rightarrow [a]\psi_i \in \mathcal{E}_a{}'$ and $(\varphi_i \wedge \varphi') \rightarrow [a](\psi_i \wedge \psi') \in \mathcal{E}_a{}''$ if and only if $\varphi_i \rightarrow [a]\psi_i \in \mathcal{E}_a^{-}$, for fixed $\varphi', \psi'$;

- $\mathcal{F}_a \subseteq \{(\varphi' \wedge \ell) \rightarrow [a](\psi \vee \ell) \; : \; \ell \in \mathfrak{Lit}\}$, for a fixed $\varphi'$; and

- $\mathcal{T}, \mathcal{E}_a{}', \mathcal{E}_a{}'', \mathcal{F}_a$ are pairwise disjoint.

Hence $card(\mathcal{E}_a{}') = card(\mathcal{E}_a{}'') = card(\mathcal{E}_a^{-})$, and $card(\mathcal{F}_a) \leq card(\mathfrak{Lit})$. Then $card(\mathcal{T}') = card(\mathcal{T} \setminus \mathcal{E}_a^{-}) + card(\mathcal{E}_a{}') + card(\mathcal{E}_a{}'') + card(\mathcal{F}_a) = card(\mathcal{T} \setminus \mathcal{E}_a^{-}) + card(\mathcal{E}_a^{-}) + card(\mathcal{E}_a^{-}) + card(\mathcal{F}_a) = card(\mathcal{T}) - card(\mathcal{E}_a^{-}) + card(\mathcal{E}_a^{-}) + card(\mathcal{E}_a^{-}) + card(\mathcal{F}_a) = card(\mathcal{T}) + card(\mathcal{E}_a^{-}) + card(\mathcal{F}_a) \leq card(\mathcal{T}) + card(\mathcal{E}_a^{-}) + card(\mathfrak{Lit})$. $\quad\square$

Given the arbitrary choice of the contraction operator for static laws, without loss of generality we can resort to a slightly modified version of it, viz. one that always gives us as result a set of static laws with the same cardinality as the original $\mathcal{S}$. (This is possible since, contrary to $\mathcal{E}$ and $\mathcal{X}$, a conjunction of static laws is still a static law, with no further rewriting.) By agreeing on that, the following proposition is straightforward:

**Proposition 5.3** *Let $\mathcal{T}$ be an action theory, $\varphi$ a static law, and $\mathcal{T}' \in \mathcal{T}_{\varphi}^{-}$. Then $card(\mathcal{T}') = card(\mathcal{T}) + 1$.*

Propositions 5.1–5.3 are positive results: if the knowledge engineer can deal with the original action theory, then she will be able to deal with the output of the algorithms. (Observe that for a given $\mathcal{T}'$ all the conditional frame axioms added to $\mathcal{F}_a$ in the contraction of an effect law can be 'factored' into a single law, so that the resulting theory has a cardinality of at most $card(\mathcal{T}) + card(\mathcal{E}_a^{-}) + 1$.)

We finish this section by observing that the size of $\mathcal{T}_{\Phi}^{-}$, the set of resulting contracted theories, depends solely on the set of static laws plus the law we contract $\mathcal{T}$ with:

**Proposition 5.4** *Let $\mathcal{T}$ be an action theory, and let $\Phi$ be a law such that $\mathcal{T} \models_{\overline{\mathsf{K}}_n} \Phi$. Then*

- $card(\mathcal{T}_{\Phi}^{-}) = card(\mathcal{S} \ominus \varphi)$, *if $\Phi$ is $\varphi$*

- $card(\mathcal{T}_{\Phi}^{-}) = card(val(\mathcal{S} \cup \{\varphi\}))$, *if $\Phi$ is either $\varphi \rightarrow \langle a \rangle \top$ or $\varphi \rightarrow [a]\psi$.*

**Proof:** The proof follows straightforwardly from the outermost loops in Algorithms 1–3. $\square$

## 6. Correctness of the Operators

We now address the correctness of our algorithms with respect to our semantics for contraction. Correctness here is understood as completeness and adequacy. Adequacy means that the algorithms output only theories whose models result from our semantic modifications of models of the original theory. Conversely, completeness says that every model resulting from the semantic modifications of models of the original theory is indeed a model of some theory output by the algorithm.

### 6.1 Challenges to Completeness and Adequacy

Let the theory $\mathcal{T} = \{p_1 \rightarrow \langle a \rangle \top, (\neg p_1 \vee p_2) \rightarrow [a]\bot, [a]\neg p_2\}$ and consider its model $\mathscr{M}$ depicted in Figure 13. (Notice that $\mathcal{T} \models_{\overline{\mathsf{K}}_n} \neg(p_1 \wedge p_2)$.) When contracting $p_1 \rightarrow [a]\neg p_2$ in $\mathscr{M}$, we get $\mathscr{M}'$ in Figure 13.

Now contracting $p_1 \rightarrow [a]\neg p_2$ from $\mathcal{T}$ using Algorithm 2 gives $\mathcal{T}_{p_1 \rightarrow [a]\neg p_2}^{-} = \{\mathcal{T}'\}$, where

$$\mathcal{T}' = \left\{ \begin{array}{l} p_1 \rightarrow \langle a \rangle \top, (\neg p_1 \vee p_2) \rightarrow [a]\bot, \\ (p_1 \wedge \neg p_2) \rightarrow [a](\neg p_2 \vee p_2), \\ (p_1 \wedge \neg p_2) \rightarrow [a](\neg p_2 \vee p_1) \end{array} \right\}$$

Notice that the formula $(p_1 \wedge \neg p_2) \rightarrow [a](\neg p_2 \vee p_1)$ is put in $\mathcal{T}'$ by Algorithm 2 because there is $\{p_1\} \subseteq \mathfrak{Lit}$ such that $\mathcal{S} \not\models_{\mathsf{CPL}} (p_1 \wedge p_2) \rightarrow \bot$ and $\mathcal{T} \not\models_{\overline{\mathsf{K}}_n} (p_1 \wedge \neg p_2) \rightarrow [a]\neg p_1$. It is clearly the case that $\not\models^{\mathscr{M}'} \mathcal{T}'$ and no theory in $\mathcal{T}^-_{p_1 \to [a]\neg p_2}$ has $\mathscr{M}'$ as model. This means that there are theories for which the contraction operators are not complete.

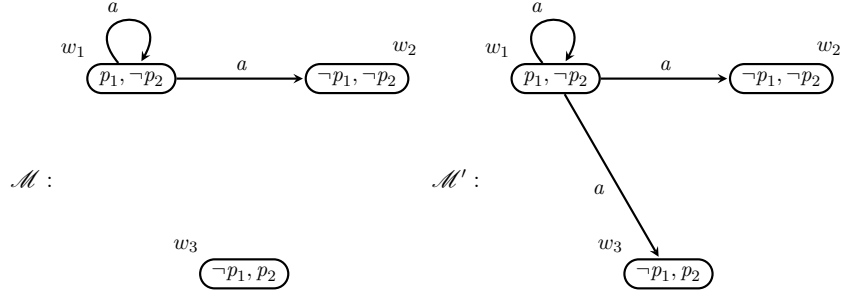

Figure 13: A model $\mathscr{M}$ of $\mathcal{T}$ and the result $\mathscr{M}'$ of contracting $p_1 \to [a]\neg p_2$ in it.

This issue arises because Algorithm 2 tries to allow a transition from the $p_1 \wedge \neg p_2$-world to a $p_2$-world that is closest to it, viz. $\{p_1, p_2\}$, but has no way of knowing that such a world does not exist. A remedy for that is replacing the test $\mathcal{T} \not\vdash_{\mathsf{K}_n} (\pi' \wedge \bigwedge_{\ell \in L} \ell) \to \bot$ for $\mathcal{S} \not\vdash_{\mathsf{CPL}} (\pi' \wedge \bigwedge_{\ell \in L} \ell) \to \bot$, but that would increase even more the complexity of the algorithm. A better option would be to have $\mathcal{S}$ 'complete enough' to allow the algorithm to determine the worlds to which a new transition could exist.

The other way round, it does not hold in general that the models of each $\mathcal{T}' \in \mathcal{T}^-_\Phi$ result from the semantic contraction of models of $\mathcal{T}$ by $\Phi$. To see why suppose that there is only one atom $p$ and one action $a$, and consider the action theory $\mathcal{T} = \{p \to [a]\bot, \langle a \rangle \top\}$. The only model of $\mathcal{T}$ is $\mathscr{M} = \langle \{\{\neg p\}\}, \{(\{\neg p\}, \{\neg p\})\} \rangle$ in Figure 14.

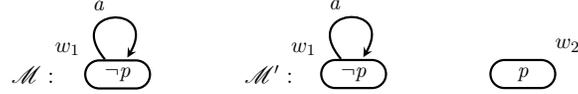

Figure 14: Inadequacy of contraction: a model $\mathscr{M}$ of $\mathcal{T}$ and a model $\mathscr{M}'$ of the theory resulting from contracting $p \to \langle a \rangle \top$ from $\mathcal{T}$.

From our definitions, $contract(\mathscr{M}, p \to \langle a \rangle \top) = \{\mathscr{M}\}$. (There is no $p$-world in $\mathscr{M}$ from which to remove an arrow.) On the other hand, $\mathcal{T}^-_{p \to \langle a \rangle \top}$ is the singleton $\{\mathcal{T}'\}$ such that $\mathcal{T}' = \{p \to [a]\bot, \neg p \to \langle a \rangle \top\}$. Then $\mathscr{M}' = \langle \{\{\neg p\}, \{p\}\}, (\{\neg p\}, \{\neg p\}) \rangle$ in Figure 14 is a model of the resulting contracted theory. Clearly, $\mathscr{M}'$ does not result from the semantic contraction of $p \to \langle a \rangle \top$ from $\mathscr{M}$: while $\neg p$ is valid in the contraction of the models of $\mathcal{T}$, it is not valid in the models of $\mathcal{T}'$. This means that there are theories for which the operators are not adequate.

This problem occurs because, in our example, the worlds that are forbidden by $\mathcal{T}$, e.g. $\{p\}$, are not preserved as such in $\mathcal{T}'$. When contracting an executability or an effect law, we are not supposed to change the possible worlds of a theory (cf. Section 3).

Fortunately correctness of the algorithms with respect to our semantics can be established for those action theories whose $\mathcal{S}$ is maximal, i.e., the set of static laws in $\mathcal{S}$ alone characterize what worlds are possible in the models of the theory. This is the principle of *modularity* (Herzig & Varzinczak, 2005b) and we briefly review it in the next section.

## 6.2 Modular Action Theories

A quite useful, albeit simple, property of domain descriptions in reasoning about actions is that of action theory modularity (Herzig & Varzinczak, 2005b).

**Definition 6.1 (Modularity)** *An action theory $\mathcal{T}$ is* modular *if and only if for every Boolean formula $\varphi \in \mathfrak{Fml}$, if $\mathcal{T} \models_{\overline{\mathsf{K}}_n} \varphi$, then $\mathcal{S} \models_{\overline{\mathsf{CPL}}} \varphi$.*

For an example of a non-modular theory, let us suppose that the action theory $\mathcal{T}$ of our coffee machine scenario were stated as

$$\mathcal{T} = \left\{ \begin{array}{c} coffee \to hot, \langle buy \rangle \top, \\ \neg coffee \to [buy] coffee, \\ token \to [buy] \neg token, \underline{\neg token \to [buy] \bot}, \\ coffee \to [buy] coffee, hot \to [buy] hot \end{array} \right\}$$

The modified law is underlined: we have (in this case wrongly) stated that the agent can always buy at the machine. Then $\mathcal{T} \models_{\overline{\mathsf{K}}_n} token$, but $\mathcal{S} \not\models_{\overline{\mathsf{CPL}}} token$.

Since the underlying multimodal logic is independently axiomatized (see Section 2.1), we can use the algorithms given by Herzig and Varzinczak (2005b) to check whether an action theory satisfies the principle of modularity. Whenever this is not the case, the algorithms return the Boolean formulas entailed by the theory which are not consequences of $\mathcal{S}$ alone. For the theory $\mathcal{T}$ above, they would return $\{token\}$: as we have stated $\langle buy \rangle \top$, from this and the inexecutability law $\neg token \to [buy] \bot$ we have that $\mathcal{T} \models_{\overline{\mathsf{K}}_n} token$. Because $\mathcal{S} \not\models_{\overline{\mathsf{CPL}}} token$, *token* is what is called an *implicit static law* (Herzig & Varzinczak, 2004) of the action theory $\mathcal{T}$.[5]

Modular action theories have several interesting computational properties. For example, consistency can be checked by just checking consistency of the static laws in $\mathcal{S}$: if $\mathcal{T}$ is modular, then $\mathcal{T} \models_{\overline{\mathsf{K}}_n} \bot$ if and only if $\mathcal{S} \models_{\overline{\mathsf{CPL}}} \bot$. Deduction of effect laws does not need the executability ones and vice versa. Deduction of an effect of a sequence of actions $a_1; \ldots; a_n$ (prediction) does not need to take into account the effect laws for actions other than $a_1, \ldots, a_n$. This applies in particular to plan validation when deciding whether $\langle a_1; \ldots; a_n \rangle \varphi$ is the case.

Modularity is not an exclusive property of action theories formalized in $\mathsf{K}_n$: similar notions have also been investigated for different contexts in other formalisms, like regulation consistency in deontic logic (Cholvy, 1999), Situation Calculus (Herzig & Varzinczak, 2005a), DL ontologies (Herzig & Varzinczak, 2006), dynamic logic (Zhang, Chopra, & Foo, 2002) and also in the Fluent Calculus (Thielscher, 2010). For more details on modularity in $\mathsf{K}_n$ action theories, as well as its role in the presence of a solution to the frame and ramification problems, see the work by Varzinczak (2006).

---

5. Implicit static laws are closely related to veridical paradoxes (Quine, 1962). It turns out that sometimes they are intuitive, but sometimes they are not. For a deep discussion on implicit static laws, see the work by Varzinczak (2006).

Another interesting property of modular action theories is the following:

**Theorem 6.1** $\mathcal{T}$ *is modular if and only if* $\mathcal{T}$ *has a canonical model.*

**Proof:** Let $\mathscr{M}_{can} = \langle W_{can}, R_{can} \rangle$ be the canonical frame of $\mathcal{T}$.

($\Rightarrow$): By definition, $\mathscr{M}_{can}$ is such that $\models^{\mathscr{M}_{can}} \mathcal{S} \wedge \mathcal{E}$. It remains to show that $\models^{\mathscr{M}_{can}} \mathcal{X}$. Let $\varphi_i \rightarrow \langle a \rangle \top \in \mathcal{X}_a$, and let $w \in W_{can}$ be such that $\models^{\mathscr{M}_{can}}_{w} \varphi_i$. Therefore for all $\varphi_j \in \mathfrak{Fml}$ such that $\mathcal{T} \models_{\overline{\mathsf{K}}_n} \varphi_j \rightarrow [a]\bot$, we must have $\not\models^{\mathscr{M}_{can}}_{w} \varphi_j$, because $\mathcal{T} \models_{\overline{\mathsf{K}}_n} \neg(\varphi_i \wedge \varphi_j)$, and as $\mathcal{T}$ is modular, $\mathcal{S} \models_{\overline{\mathsf{CPL}}} \neg(\varphi_i \wedge \varphi_j)$, and hence $\models^{\mathscr{M}_{can}} \neg(\varphi_i \wedge \varphi_j)$. Then by the construction of $\mathscr{M}_{can}$, there is some $w' \in W_{can}$ such that $\models^{\mathscr{M}_{can}}_{w'} \psi$ for all $\varphi \rightarrow [a]\psi \in \mathcal{E}_a$ such that $\models^{\mathscr{M}_{can}}_{w} \varphi$. Thus $R_a(w) \neq \emptyset$ and $\models^{\mathscr{M}_{can}}_{w} \varphi_i \rightarrow \langle a \rangle \top$.

($\Leftarrow$): Suppose $\mathcal{T}$ is not modular. Then there must be some $\varphi \in \mathfrak{Fml}$ such that $\mathcal{T} \models_{\overline{\mathsf{K}}_n} \varphi$ and $\mathcal{S} \not\models_{\overline{\mathsf{CPL}}} \varphi$. This means that there is $v \in val(\mathcal{S})$ such that $v \not\models \varphi$. As $v \in W_{can}$ (because $W_{can}$ contains all possible valuations of $\mathcal{S}$), $\mathscr{M}_{can}$ is not a model of $\mathcal{T}$. $\square$

### 6.3 Correctness Under Modularity

As shown by Herzig and Varzinczak (2007), given an action theory formalized with any framework available in the literature allowing for the expression of our three basic types of laws, it is always possible to ensure modularity. Moreover, as we are going to see in the sequel (cf. Section 7.2), it has to be computed at most once during the evolution of the action theory. Hence, relying on modular theories is not a limitation at all to our approach.

The following theorem establishes that under the assumption that the action theory $\mathcal{T}$ is modular, the semantic contraction of a formula $\Phi$ from the set of models of $\mathcal{T}$ produces models of some contracted theory in $\mathcal{T}_{\Phi}^{-}$.

**Theorem 6.2** *Let* $\mathcal{T}$ *be modular, and* $\Phi$ *be a law. For all* $\mathcal{M}' \in \mathcal{M}_{\Phi}^{-}$ *such that* $\models^{\mathscr{M}} \mathcal{T}$ *for every* $\mathscr{M} \in \mathcal{M}$, *there is* $\mathcal{T}' \in \mathcal{T}_{\Phi}^{-}$ *such that* $\models^{\mathscr{M}'} \mathcal{T}'$ *for every* $\mathscr{M}' \in \mathcal{M}'$.

**Proof:** See Appendix A. $\square$

The next theorem establishes the other way round: under modularity models of theories in $\mathcal{T}_{\Phi}^{-}$ are all models of the semantic contraction of $\Phi$ from models of $\mathcal{T}$.

**Theorem 6.3** *Let* $\mathcal{T}$ *be modular,* $\Phi$ *a law, and* $\mathcal{T}' \in \mathcal{T}_{\Phi}^{-}$. *For all* $\mathscr{M}'$ *such that* $\models^{\mathscr{M}'} \mathcal{T}'$, *there is* $\mathcal{M}' \in \mathcal{M}_{\Phi}^{-}$ *such that* $\mathscr{M}' \in \mathcal{M}'$ *and* $\models^{\mathscr{M}} \mathcal{T}$ *for every* $\mathscr{M} \in \mathcal{M}$.

**Proof:** See Appendix B. $\square$

With these two theorems we get the correctness of our operators:

**Corollary 6.1** *Let* $\mathcal{T}$ *be modular,* $\Phi$ *a law, and* $\mathcal{T}' \in \mathcal{T}_{\Phi}^{-}$. *Then* $\mathcal{T}' \models_{\overline{\mathsf{K}}_n} \Psi$ *if and only if* $\models^{\mathscr{M}'} \Psi$ *for every* $\mathscr{M}' \in \mathcal{M}'$ *such that* $\mathcal{M}' \in \mathcal{M}_{\Phi}^{-}$ *for some* $\mathcal{M}$ *such that* $\models^{\mathscr{M}} \mathcal{T}$ *for all* $\mathscr{M} \in \mathcal{M}$.

**Proof:**

($\Rightarrow$): Let $\mathscr{M}'$ be such that $\models^{\mathscr{M}'} \mathcal{T}'$. By Theorem 6.3, there is $\mathcal{M}' \in \mathcal{M}_{\Phi}^{-}$ such that $\mathscr{M}' \in \mathcal{M}'$ for some $\mathcal{M}$ such that $\models^{\mathscr{M}} \mathcal{T}$ for all $\mathscr{M} \in \mathcal{M}$. From this and $\mathcal{T}' \Vdash_{\overline{\mathsf{K}}_n} \Psi$, we have $\models^{\mathscr{M}'} \Psi$.

($\Leftarrow$): Suppose $\mathcal{T}' \not\Vdash_{\overline{\mathsf{K}}_n} \Psi$. (We show that there is some model $\mathscr{M}' \in \mathcal{M}'$ such that $\mathcal{M}' \in \mathcal{M}_{\Phi}^{-}$ for some $\mathcal{M}$ with $\models^{\mathscr{M}} \mathcal{T}$ for all $\mathscr{M} \in \mathcal{M}$, and $\not\models^{\mathscr{M}'} \Psi$.)

Given that $\mathcal{T}$ is modular, by Lemma B.1 $\mathcal{T}'$ is modular, too. Then, by Lemma B.3, there is some $\mathscr{M}' = \langle val(\mathcal{S}'), R' \rangle$ such that $\not\models^{\mathscr{M}'} \Psi$. Clearly $\models^{\mathscr{M}'} \mathcal{T}'$, and from Lemma B.4 the result follows. $\qquad\square$

# 7. Assessment of Postulates for Change

Do our action theory change operators satisfy the classical postulates for change? Before answering this question, one should ask: do our operators behave like revision or update operators? We here address this issue and then show which postulates for theory change are satisfied by our definitions.

## 7.1 Contraction or Erasure?

The distinction between revision/contraction and update/erasure for classical theories is historically controversial in the literature. The same is true for the case of modal theories describing actions and their effects. We here rephrase Katsuno and Mendelzon's definitions (1992) in our terms so that we can see to which one our method is closer.

In Katsuno and Mendelzon's view, contracting a law $\Phi$ from an action theory $\mathcal{T}$ intuitively means that the description of the possible behavior of the dynamic world $\mathcal{T}$ must be adjusted to the possibility of $\Phi$ being false. This amounts to selecting from the models of $\neg\Phi$ those that are closest to models of $\mathcal{T}$ and allow them as models of the result.

In contrast, update methods select, for each model $\mathscr{M}$ of $\mathcal{T}$, the set of models of $\Phi$ that are closest to $\mathscr{M}$. Erasing $\Phi$ from $\mathcal{T}$ means adding models to $\mathcal{T}$; for each model $\mathscr{M}$, we add all those models closest to $\mathscr{M}$ in which $\Phi$ is false. Hence, from our constructions so far it seems that our operators are closer to update than to revision.

Moreover, according to Katsuno and Mendelzon's view (1992), our change operators would also be classified as update because we make modifications in each model independently, i.e., without changing other models.[6] Besides that, in our setting a different ordering on the resulting models is induced by each model of the theory $\mathcal{T}$ (see Definitions 3.3, 3.7 and 3.10), which according to Katsuno and Mendelzon is a typical property of an update/erasure method.

Nevertheless, things get quite different when it comes to the postulates for theory change.

## 7.2 The Postulates

In this section we analyze the behavior of our action theory change operators with respect to AGM-like postulates. Here we follow Katsuno and Mendelzon's presentation of the postulates to assess both contraction and erasure. Let $\mathcal{T} = \mathcal{S} \cup \mathcal{E} \cup \mathcal{X}$ denote an action theory and $\Phi$ denote a law.

---

6. Even if when contracting an effect law from one particular model we need to check the other models of the theory, those are not modified.

**Monotonicity Postulate:** $\mathcal{T} \models_{\overline{\mathsf{K}}_n} \mathcal{T}'$, for all $\mathcal{T}' \in \mathcal{T}^-_\Phi$.

This postulate is our version of Katsuno and Mendelzon's (C1) and (E1) postulates for contraction and erasure, respectively, and it is satisfied by our change operators. The proof is in Lemma A.1. Such a postulate is not satisfied by the operators proposed by Herzig et al. (2006): there when removing e.g. an executability law $\varphi \rightarrow \langle a \rangle \top$ one may make $\varphi \rightarrow [a] \bot$ valid in all models of the resulting theory.

**Preservation Postulate:** If $\mathcal{T} \not\models_{\overline{\mathsf{K}}_n} \Phi$, then $\models_{\overline{\mathsf{K}}_n} \mathcal{T} \leftrightarrow \mathcal{T}'$, for all $\mathcal{T}' \in \mathcal{T}^-_\Phi$.

This is Katsuno and Mendelzon's (C2) postulate. Our operators satisfy it as far as whenever $\mathcal{T} \not\models_{\overline{\mathsf{K}}_n} \Phi$, then the models of the resulting theory are exactly the models of $\mathcal{T}$, because these are the minimal models falsifying $\Phi$.

The corresponding version of Katsuno and Mendelzon's (E2) postulate about erasure, i.e., if $\mathcal{T} \models_{\overline{\mathsf{K}}_n} \neg \Phi$, then $\models_{\overline{\mathsf{K}}_n} \mathcal{T} \leftrightarrow \mathcal{T}'$, for all $\mathcal{T}' \in \mathcal{T}^-_\Phi$, is clearly also satisfied by our operators as a special case of the postulate above. Satisfaction of (C2) indicates that our operators are closer to contraction than to erasure.

**Success Postulate:** If $\mathcal{T} \not\models_{\overline{\mathsf{K}}_n} \bot$ and $\not\models_{\overline{\mathsf{K}}_n} \Phi$, then $\mathcal{T}' \not\models_{\overline{\mathsf{K}}_n} \Phi$, for all $\mathcal{T}' \in \mathcal{T}^-_\Phi$.

This postulate is our version of Katsuno and Mendelzon's (C3) and (E3) postulates. If $\Phi$ is a propositional $\varphi \in \mathfrak{Fml}$, our operators satisfy it, as long as the classical propositional change operator satisfies it as well. For the general case, however, as stated the postulate is not always satisfied. This is shown by the following example: let $\mathcal{T} = \{\neg p, \langle a \rangle \top, p \rightarrow [a] \bot\}$. Note that $\mathcal{T}$ is modular and consistent. Now, contracting the (contingent) formula $p \rightarrow \langle a \rangle \top$ from $\mathcal{T}$ gives us $\mathcal{T}' = \mathcal{T}$. Clearly $\mathcal{T}' \models_{\overline{\mathsf{K}}_n} p \rightarrow \langle a \rangle \top$. This happens because, despite not being a tautology, $p \rightarrow \langle a \rangle \top$ is a 'trivial' formula with respect to $\mathcal{T}$: since $\neg p$ is valid in all $\mathcal{T}$-models, $p \rightarrow \langle a \rangle \top$ is trivially true in these models (cf. end of Section 3.1).

Fortunately, for all those formulas that are non-trivial consequences of the theory, our operators guarantee success of contraction:

**Theorem 7.1** *Let $\mathcal{T}$ be consistent, and $\Phi$ be an executability or an effect law such that $\mathcal{S} \not\models_{\overline{\mathsf{K}}_n} \Phi$. If $\mathcal{T}$ is modular, then $\mathcal{T}' \not\models_{\overline{\mathsf{K}}_n} \Phi$ for every $\mathcal{T}' \in \mathcal{T}^-_\Phi$.*

**Proof:** Let us suppose that there is some $\mathcal{T}' \in \mathcal{T}^-_\Phi$ such that $\mathcal{T}' \models_{\overline{\mathsf{K}}_n} \Phi$. Since $\mathcal{T}$ is modular, Corollary 6.1 tells us that $\models^{\mathscr{M}'} \Phi$ for every $\mathscr{M}' \in \mathcal{M}'$ such that $\mathcal{M}' \in \mathcal{M}^-_\Phi$, where $\mathcal{M} = \{\mathscr{M} : \models^{\mathscr{M}} \mathcal{T} \text{ and } \mathscr{M} = \langle val(\mathcal{S}), R \rangle\}$.

If $\models^{\mathscr{M}'} \Phi$ for every $\mathscr{M}' \in \mathcal{M}'$, then even for $\mathscr{M}'' \in \mathcal{M}' \setminus \mathcal{M}$ we have $\models^{\mathscr{M}''} \Phi$. But $\mathscr{M}'' \in \mathcal{M}^-_\Phi$ for some $\mathscr{M} \in \mathcal{M}$, and by definition $\not\models^{\mathscr{M}''} \Phi$. Hence $\mathcal{M}^-_\Phi = \emptyset$, and then the truth of $\Phi$ in $\mathscr{M}$ does not depend on the accessibility relation $R_a$. Hence, whether $\Phi$ has the form $\varphi \rightarrow \langle a \rangle \top$ or $\varphi \rightarrow [a]\psi$, for $\varphi, \psi \in \mathfrak{Fml}$, this holds only if $\mathcal{S} \models_{\mathsf{CPL}} \neg \varphi$ (see Definitions 3.1 and 3.5), and therefore we get $\mathcal{S} \models_{\overline{\mathsf{K}}_n} \Phi$. $\square$

**Equivalences Postulate:** If $\models_{\overline{\mathsf{K}}_n} \mathcal{T}_1 \leftrightarrow \mathcal{T}_2$ and $\models_{\overline{\mathsf{K}}_n} \Phi_1 \leftrightarrow \Phi_2$, then $\models_{\overline{\mathsf{K}}_n} \mathcal{T}_1' \leftrightarrow \mathcal{T}_2'$, for some $\mathcal{T}_1' \in (\mathcal{T}_1)^-_{\Phi_2}$ and $\mathcal{T}_2' \in (\mathcal{T}_2)^-_{\Phi_1}$.

This postulate corresponds to Katsuno and Mendelzon's (C4) and (E4) postulates. It is worth noting that equivalence here is considered always modulo action laws, i.e., the formulas are assumed to be either static laws, effect laws or executability laws, as well as their equivalents. Moreover we remember that the theories here must be action theories, i.e., sets of action laws of our three basic types. Under modularity and the assumption that the propositional change operator satisfies (C4)/(E4), our operations satisfy this postulate:

**Theorem 7.2** *Let $\mathcal{T}_1$ and $\mathcal{T}_2$ be modular. If $\models_{\overline{\mathsf{K}}_n} \mathcal{T}_1 \leftrightarrow \mathcal{T}_2$ and $\models_{\overline{\mathsf{K}}_n} \Phi_1 \leftrightarrow \Phi_2$, then for each $\mathcal{T}_1' \in (\mathcal{T}_1)^-_{\Phi_2}$ there is $\mathcal{T}_2' \in (\mathcal{T}_2)^-_{\Phi_1}$ such that $\models_{\overline{\mathsf{K}}_n} \mathcal{T}_1' \leftrightarrow \mathcal{T}_2'$, and vice-versa.*

**Proof:** The proof follows straight from our results: since $\models_{\overline{\mathsf{K}}_n} \mathcal{T}_1 \leftrightarrow \mathcal{T}_2$ and $\models_{\overline{\mathsf{K}}_n} \Phi_1 \leftrightarrow \Phi_2$, they have pairwise the same models. Hence, given $\mathscr{M}$ such that $\models^{\mathscr{M}} \mathcal{T}_1$ and $\models^{\mathscr{M}} \mathcal{T}_2$, the semantic contraction of $\Phi_1$ and that of $\Phi_2$ from $\mathscr{M}$ have the same operations on $\mathscr{M}$. As $\mathcal{T}_1$ and $\mathcal{T}_2$ are modular, Corollary 6.1 guarantees we get the same syntactical results. Moreover, as the classical operator $\ominus$ satisfies (C4)/(E4), if follows that $\models_{\overline{\mathsf{K}}_n} \mathcal{T}_1' \leftrightarrow \mathcal{T}_2'$. □

**Recovery Postulate:** $\mathcal{T}' \cup \{\Phi\} \models_{\overline{\mathsf{K}}_n} \mathcal{T}$, for all $\mathcal{T}' \in \mathcal{T}^-_\Phi$.

This is the action theory counterpart of Katsuno and Mendelzon's (C5) and (E5) postulates. Again we rely on modularity in order to satisfy it.

**Theorem 7.3** *Let $\mathcal{T}$ be modular. $\mathcal{T}' \cup \{\Phi\} \models_{\overline{\mathsf{K}}_n} \mathcal{T}$, for all $\mathcal{T}' \in \mathcal{T}^-_\Phi$.*

**Proof:** If $\mathcal{T} \not\models_{\overline{\mathsf{K}}_n} \Phi$, because our operators satisfy the preservation postulate, $\mathcal{T}' = \mathcal{T}$, and then the result follows by monotonicity.

Let $\mathcal{T} \models_{\overline{\mathsf{K}}_n} \Phi$, and let $\mathcal{M}'$ denote the set of all models of $\mathcal{T}'$. As $\mathcal{T}$ is modular, by Corollary 6.1 every $\mathscr{M}' \in \mathcal{M}'$ is such that either $\models^{\mathscr{M}'} \mathcal{T}$ (and then $\models^{\mathscr{M}'} \Phi$) or $\mathscr{M}' \in contract(\mathscr{M}, \Phi)$ (and then $\mathscr{M}' \in \mathcal{M}^-_\Phi$) for some $\mathscr{M}$ such that $\models^{\mathscr{M}} \mathcal{T}$.

Let $\mathcal{M}''$ denote the set of all models of $\mathcal{T}' \cup \{\Phi\}$. Clearly $\mathcal{M}'' \subseteq \mathcal{M}'$, by monotonicity. Moreover, every $\mathscr{M}'' \in \mathcal{M}''$ is such that $\models^{\mathscr{M}''} \Phi$, hence $\mathscr{M}'' \notin \mathcal{M}^-_\Phi$ for every $\mathscr{M}$ such that $\models^{\mathscr{M}} \mathcal{T}$, and then $\mathscr{M}'' \notin contract(\mathscr{M}, \Phi)$, for any $\mathscr{M}$ model of $\mathcal{T}$. Thus $\mathscr{M}''$ is a model of $\mathcal{T}$ and then $\mathcal{T}' \cup \{\Phi\} \models_{\overline{\mathsf{K}}_n} \mathcal{T}$. □

Let $\bigvee \mathcal{T}^-_\Phi$ denote the disjunction of all $\mathcal{T}'$ in $\mathcal{T}^-_\Phi$.

**Disjunctive Rule:** $(\mathcal{T}_1 \vee \mathcal{T}_2)^-_\Phi$ is equivalent to $\bigvee (\mathcal{T}_1)^-_\Phi \vee \bigvee (\mathcal{T}_2)^-_\Phi$.

This is our version of (E8) erasure postulate by Katsuno and Mendelzon. Clearly our syntactical operators do not manage to contract a law from a disjunction of theories: $\mathcal{T}_1 \vee \mathcal{T}_2$ is not an action theory and cannot in general be rewritten as one. Nevertheless, by proving that it holds in the semantics, from the correctness of our operators, we get an equivalent operation. Again the fact that the theories under concern are modular gives us the result.

**Theorem 7.4** *Let $\mathcal{T}_1$ and $\mathcal{T}_2$ be modular, and $\Phi$ be a law. Then*

$$\models_{\overline{K}_n} \bigvee (\mathcal{T}_1 \vee \mathcal{T}_2)_\Phi^- \leftrightarrow (\bigvee (\mathcal{T}_1)_\Phi^- \vee \bigvee (\mathcal{T}_2)_\Phi^-)$$

**Proof:**

($\Leftarrow$): Let $\mathscr{M}'$ be such that $\models^{\mathscr{M}'} \bigvee (\mathcal{T}_1)_\Phi^- \vee \bigvee (\mathcal{T}_2)_\Phi^-$. Then $\models^{\mathscr{M}'} \bigvee (\mathcal{T}_1)_\Phi^-$ or $\models^{\mathscr{M}'} \bigvee (\mathcal{T}_2)_\Phi^-$. Suppose $\models^{\mathscr{M}'} \bigvee (\mathcal{T}_1)_\Phi^-$ (the other case is analogous). Then there is $(\mathcal{T}_1)' \in (\mathcal{T}_1)_\Phi^-$ such that $\models^{\mathscr{M}'} (\mathcal{T}_1)'$. Then by Corollary 6.1, there is $\mathcal{M}' \in \mathcal{M}_\Phi^-$ such that $\mathscr{M}' \in \mathcal{M}'$, for $\mathcal{M}$ a set of models of $\mathcal{T}_1$. Then $\mathscr{M}'$ is a model resulting from contracting $\Phi$ from models of $\mathcal{T}_1$ , and then $\mathscr{M}'$ also results from contracting $\Phi$ in models of $\mathcal{T}_1 \vee \mathcal{T}_2$, viz. those models of $\mathcal{T}_1$. Then by Corollary 6.1, there is $(\mathcal{T}_1 \vee \mathcal{T}_2)' \in (\mathcal{T}_1 \vee \mathcal{T}_2)_\Phi^-$ such that $\models^{\mathscr{M}'} (\mathcal{T}_1 \vee \mathcal{T}_2)'$, and then $\models^{\mathscr{M}'} \bigvee (\mathcal{T}_1 \vee \mathcal{T}_2)_\Phi^-$.

($\Rightarrow$): Let $\mathscr{M}'$ be such that $\models^{\mathscr{M}'} \bigvee (\mathcal{T}_1 \vee \mathcal{T}_2)_\Phi^-$. Then there is $(\mathcal{T}_1 \vee \mathcal{T}_2)' \in (\mathcal{T}_1 \vee \mathcal{T}_2)_\Phi^-$ such that $\models^{\mathscr{M}'} (\mathcal{T}_1 \vee \mathcal{T}_2)'$. By Corollary 6.1, there is $\mathcal{M}' \in \mathcal{M}_\Phi^-$ such that $\mathscr{M}' \in \mathcal{M}'$, for $\mathcal{M}$ a set of models of $\mathcal{T}_1 \vee \mathcal{T}_2$. Then $\mathscr{M}'$ is a model resulting from contracting $\Phi$. Hence $\mathscr{M}'$ results from contracting $\Phi$ from models of $\mathcal{T}_1$ or from models of $\mathcal{T}_2$. Suppose the former is the case (the second is analogous). Then by Corollary 6.1 there is $(\mathcal{T}_1)' \in (\mathcal{T}_1)_\Phi^-$ such that $\models^{\mathscr{M}'} (\mathcal{T}_1)'$, and then $\models^{\mathscr{M}'} \bigvee (\mathcal{T}_1)_\Phi^-$. $\hfill\square$

We have thus shown that our constructions satisfy the (E8) postulate. Nevertheless, as far as we see, it is not immediate whether it is really expected here. This supports our position that our operators' behavior is closer to contraction than to erasure.

As we have seen from the results above, modularity is a sufficient condition for the satisfaction of the AGM-like postulates for action theory contraction. To finish up we state a new postulate:

**Preservation of Modularity:**   If $\mathcal{T}$ is modular, then every $\mathcal{T}' \in \mathcal{T}_\Phi^-$ is modular.

Changing a modular theory should not make it non-modular. This is not a standard postulate, but we think that since it is a good property modularity should be preserved across changing an action theory. If so, this means that whether a theory is modular or not can be checked once for all and one does not need to care about it during the future evolution of the action theory, i.e., when other changes will be made on it. Our operators satisfy this postulate and the proof is given in Appendix B.

Now one may naturally asks whether we can get a characterization result in the traditional AGM sense, i.e., whether any contraction operator satisfying all our versions of the postulates is one of our three contraction operations. Unfortunately, good sense points towards a negative answer: there might well be an operator satisfying all the above postulates that, by not complying with all the assumptions in the RAC community (Shanahan, 1997), is not necessarily one of the operators defined in Section 3 (cf. the discussion on general formula contraction in Section 10). To witness, consider for example an operator that also modifies worlds when contracting effect laws. This supports one of the contentions of the present work, viz. that classical belief change cannot be fully transposed to action theories and expected to give exactly the same kind of outcome. Similar negative results have also been found for revision in DL ontologies (Flouris, Plexousakis, & Antoniou, 2004) and contraction of Horn theories (Booth, Meyer, & Varzinczak, 2009).

## 8. A Semantics for Action Theory Revision

So far we have analyzed the case of contraction: the knowledge engineer realizes that the theory is too strong and therefore it has to be weakened. Let us now take a look at the other way round, i.e., the theory is (possibly) too liberal and the agent discovers new *laws* about the world that should be added to her beliefs, which amounts to strengthening them.

Suppose that the action theory of our scenario example were initially stated as follows:

$$\mathcal{T} = \left\{ \begin{array}{c} coffee \rightarrow hot, token \rightarrow \langle buy \rangle \top, \\ \neg coffee \rightarrow [buy] coffee, \neg token \rightarrow [buy] \bot, \\ coffee \rightarrow [buy] coffee, hot \rightarrow [buy] hot \end{array} \right\}$$

Then the canonical model of theory $\mathcal{T}$ is as shown in Figure 15.

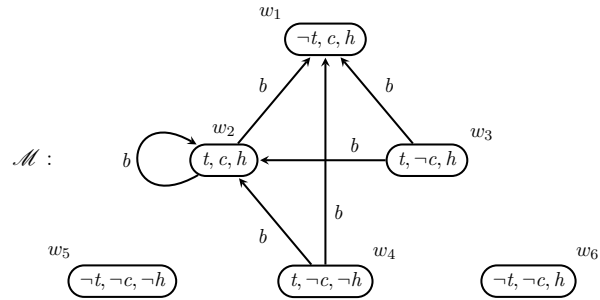

Figure 15: Canonical model of the new initial action domain description.

Looking at model $\mathscr{M}$ in Figure 15 we can see that, for example, the agent does not know that she loses her token every time she buys coffee at the machine. This is a new law that she should incorporate to her knowledge base at some stage of her action theory evolution.

Contrary to contraction, where we want the negation of some law to become *satisfiable*, in revision we want to make a new law *valid*. This means that one has to eliminate all cases satisfying its negation. This depicts the duality between revision and contraction: whereas in the latter one invalidates a formula by making its negation satisfiable, in the former one makes a formula valid by forcing its negation to be unsatisfiable prior to adding the new law to the theory.

The idea behind our semantics for revision is as follows: we initially have a set of models $\mathcal{M}$ in which a given formula $\Phi$ is (potentially) not valid, i.e., $\Phi$ is (possibly) not true in every model in $\mathcal{M}$. In the result we want to have only models of $\Phi$. Adding $\Phi$-models to $\mathcal{M}$ is of no help. Moreover, adding models makes us lose laws: the corresponding resulting theory would be more liberal.

One solution amounts to deleting from $\mathcal{M}$ those models that are not $\Phi$-models. Of course removing only some of them does not solve the problem, we must delete every such a model. By doing that, all resulting models will be models of $\Phi$. (This corresponds to *theory expansion*, when the resulting theory is satisfiable.) However, if $\mathcal{M}$ contains no model of $\Phi$, we will end up with $\emptyset$. Consequence: the resulting theory is inconsistent. (This is the main revision problem.) In this case the solution is to *substitute* each model $\mathscr{M}$ in $\mathcal{M}$ by its *nearest modification* $\mathcal{M}_\Phi^\star$ that makes $\Phi$ true. This lets us to keep as close as possible to the original models we had. But, what if for one model in $\mathcal{M}$ there are several minimal (incomparable) modifications of it validating $\Phi$? In that case we shall consider all of them. The result will also be a *list of models* $\mathcal{M}_\Phi^\star$, all being models of $\Phi$.

Before defining the revision of sets of models, we present what modifications of (individual) models are.

## 8.1 Revising a Model by a Static Law

Suppose that our coffee deliverer agent discovers that the only hot drink that is served on the machine is coffee. In this case, she might want to revise her beliefs with the new static law $\neg coffee \rightarrow \neg hot$: she cannot hold a hot drink that is not a coffee.

Considering the model depicted in Figure 15, one can see that the Boolean formula $\neg coffee \wedge hot$ is satisfiable (there is a world of the model in which it holds). Since we do not want this to be the case, the first step is to *remove* all worlds in which $\neg coffee \wedge hot$ is true. The second step is to guarantee that all the remaining worlds (if any) satisfy the new static law. Such an issue has been largely addressed in the literature on propositional belief base revision and update (Gärdenfors, 1988; Winslett, 1988; Katsuno & Mendelzon, 1992; Herzig & Rifi, 1999). Here we can achieve that with a semantics similar to that of classical revision operators: basically one can change the set of possible valuations, by removing or adding worlds.

In our example, removing the possible worlds $\{t, \neg c, h\}$ and $\{\neg t, \neg c, h\}$ would do the job (there is no need to add new valuations since the new incoming law is already satisfied in at least one world of the original model, and therefore the resulting set of worlds is non-empty).

The delicate point in removing worlds is that this may have as consequence the loss of some executability laws: in the example, if there were some transition from some world $w$ to say $\{\neg t, \neg c, h\}$, then removing the latter from the model would make the action under concern no longer executable in $w$, if it was the only transition labeled by that action leaving it. From a semantic point of view, this is intuitive: if the state of the world to which we could move is no longer possible, then we do not have a transition to that state anymore. Therefore, if that transition was the only one we had, it is natural to lose it.

Similarly, one could ask what to do with the accessibility relation if new worlds are added, i.e., when expansion is not possible. Following the discussion in Section 3.3, we here prefer not to add new transitions systematically to the accessibility relation. Hence we shall postpone correction of executability laws, if needed. This approach may be debatable, but with the information we have at hand, this is the safest way of changing static laws. (See also the discussion in Sections 3.3 and 4.2.)

The semantics for revision of one model by a static law is as follows:

**Definition 8.1** *Let $\mathcal{M} = \langle W, R \rangle$. $\mathcal{M}' = \langle W', R' \rangle \in \mathcal{M}_\varphi^\star$ if and only if:*

- $W' = (W \setminus val(\neg\varphi)) \cup W_\varphi$, *where* $W_\varphi \subseteq val(\varphi)$; *and*

- $R' \subseteq R$.

Clearly unless $\varphi \models_{\mathsf{CPL}} \bot$, we have that $\models^{\mathscr{M}'} \varphi$ for each $\mathscr{M}' \in \mathscr{M}^{\star}_{\varphi}$. The minimal models resulting from revising a model $\mathscr{M}$ by $\varphi$ are those closest to $\mathscr{M}$ with respect to $\preceq_{\mathscr{M}}$:

**Definition 8.2** *Let $\mathscr{M}$ be a model and $\varphi$ a static law.* $revise(\mathscr{M}, \varphi) = \bigcup \min\{\mathscr{M}^{\star}_{\varphi}, \preceq_{\mathscr{M}}\}$.

In the example of model $\mathscr{M}$ in Figure 15, $revise(\mathscr{M}, \neg coffee \rightarrow \neg hot)$ is the singleton $\{\mathscr{M}'\}$, where $\mathscr{M}'$ is as shown in Figure 16.

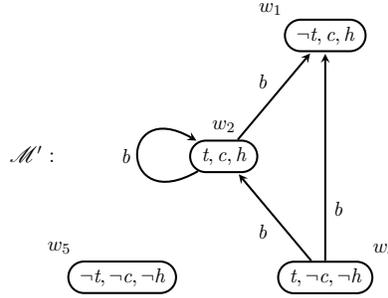

Figure 16: Model resulting from revising the model $\mathscr{M}$ in Figure 15 with $\neg coffee \rightarrow \neg hot$.

## 8.2 Revising a Model by an Effect Law

Let us suppose now that our agent eventually discovers that after buying coffee she does not keep her token anymore. (That was a design mistake that the agent still possesses a token even after ordering a coffee at the machine). This means that her theory should now be revised in such a way that the new effect law $token \rightarrow [buy]\neg token$ holds. Looking at model $\mathscr{M}$ in Figure 15, this amounts to guaranteeing that the formula $token \wedge \langle buy \rangle token$ is satisfiable in none of its worlds. To do that, we have to look at all the worlds satisfying this formula (if any) and either (*i*) make *token* false in each of these worlds; or (*ii*) make $\langle buy \rangle token$ false in all of them. If we chose the first option, we will essentially flip the truth value of literal *token* in the respective worlds, which changes the set of valuations of the model. If we chose the latter, we will basically remove *buy*-arrows leading to *token*-worlds. In that case, a change in the accessibility relation will be made.

In our example, we have that the possible worlds $\{token, coffee, hot\}$, $\{token, \neg coffee, hot\}$ and $\{token, \neg coffee, \neg hot\}$ satisfy $token \wedge \langle buy \rangle token$ and all they have to change.

Flipping *token* in all these worlds to $\neg token$ would do the job, but would also have as consequence the introduction of a new static law: $\neg token$ would now be valid, i.e., the agent never has a token! Another issue with this approach is that by making $\neg token$ true everywhere, the new incoming law $token \rightarrow [buy]\neg token$ will be *trivially* true in the resulting model, which does not mean that there is an execution of action *buy* from a *token*-world to a $\neg token$ one. This defeats the purpose of changing the action theory on the basis that it has been observed that every execution of the action under consideration should lead to $\neg token$-contexts.

One of our contentions in the present work is that changing action laws should never have as a side effect a change in the static laws (cf. Sections 3 and 4). Given their special status (Shanahan, 1997), these should change only if explicitly required. In this case, each world satisfying $token \land \langle buy \rangle token$ has to be changed so that $\langle buy \rangle token$ is no longer true in it. In our example, we should remove the transitions $(\{token, coffee, hot\}, \{token, coffee, hot\})$, $(\{token, \neg coffee, hot\}, \{token, coffee, hot\})$ and $(\{token, \neg coffee, \neg hot\}, \{token, coffee, hot\})$.

The semantics of one model revision for the case of a new effect law is:

**Definition 8.3** *Let $\mathscr{M} = \langle W, R \rangle$. $\mathscr{M}' = \langle W', R' \rangle \in \mathscr{M}^{\star}_{\varphi \to [a]\psi}$ if and only if:*

- $W' = W$;

- $R' \subseteq R$;

- *If $(w, w') \in R \setminus R'$, then $\models^{\mathscr{M}}_{w} \varphi$; and*

- $\models^{\mathscr{M}'} \varphi \to [a]\psi$.

The minimal models resulting from the revision of a model $\mathscr{M}$ by a new effect law are those that are closest to $\mathscr{M}$ with respect to our order on the models $\preceq_{\mathscr{M}}$:

**Definition 8.4** *Let $\mathscr{M}$ be a model and $\varphi \to [a]\psi$ an effect law. $revise(\mathscr{M}, \varphi \to [a]\psi) = \bigcup \min\{\mathscr{M}^{\star}_{\varphi \to [a]\psi}, \preceq_{\mathscr{M}}\}$.*

Taking once again $\mathscr{M}$ as shown in Figure 15, $revise(\mathscr{M}, token \to [buy]\neg token)$ will be the singleton $\{\mathscr{M}'\}$ (Figure 17).

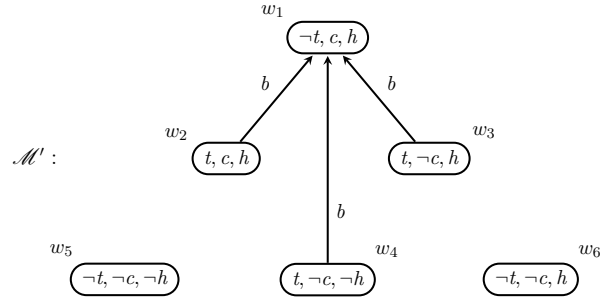

Figure 17: Model resulting from revising the model $\mathscr{M}$ in Figure 15 with the new effect law $token \to [buy]\neg token$.

## 8.3 Revising a Model by an Executability Law

Let us now suppose that in some stage it has been decided to grant free coffee to everybody. Faced with this information, the agent will now revise her laws to reflect the fact that *buy* can also be executed in $\neg token$-contexts: $\neg token \to \langle buy \rangle \top$ is a new executability law (and therefore we will have $\langle buy \rangle \top$ in all new models of the agent's beliefs).

Considering again the model in Figure 15, we observe that $\neg(\neg token \to \langle buy \rangle \top)$ is satisfiable in $\mathscr{M}$. This means that we must throw $\neg token \land [buy]\bot$ away to ensure that the new formula becomes true in the new model, i.e., satisfied by all of its worlds.

To remove $\neg token \land [buy]\bot$ we have to look at all worlds satisfying it and modify $\mathscr{M}$ so that they no longer satisfy that formula. Given worlds $\{\neg token, \neg coffee, \neg hot\}$ and $\{\neg token, \neg coffee, hot\}$, we have two options: change the interpretation of $token$ or add new transitions leaving these worlds. A question that arises is 'what choice is more drastic: change a world or a transition'? Again, here we think that changing the world's content (the valuation) is more drastic, as the existence of such a world was foreseen by some static law and is hence assumed to be as it is, unless we have enough information supporting the contrary, in which case we explicitly change the static laws (see above). Moreover, changing the truth value of $token$ in these worlds would *trivialize* the new incoming law $\neg token \to \langle buy \rangle \top$ in the new model, defeating the purpose of guaranteeing the existence of a $buy$-transition from a $\neg token$-context. Therefore we shall add a new $buy$-arrow from each of $\{\neg token, \neg coffee, \neg hot\}$ and $\{\neg token, \neg coffee, hot\}$.

Having agreed on that, the issue now is: which worlds should the new transitions be directed to? Recalling the reasoning developed in Section 3.2, in order to comply with minimal change, the new transitions shall be directed to worlds that are relevant targets of each of the $\neg token$-worlds in question. In our example, $\{\neg token, coffee, hot\}$ is the only relevant target world here: the two other $\neg token$-worlds violate the effect $coffee$ of $buy$, while the three $token$-worlds would make us violate the frame axiom $\neg token \to [buy]\neg token$.

The semantics for one model revision by a new executability law is as follows:

**Definition 8.5** *Let* $\mathscr{M} = \langle W, R \rangle$. $\mathscr{M}' = \langle W', R' \rangle \in \mathscr{M}^{\star}_{\varphi \to \langle a \rangle \top}$ *if and only if:*

- $W' = W$;

- $R \subseteq R'$;

- *If* $(w, w') \in R' \setminus R$, *then* $w' \in RelTarget(w, \varphi \to [a]\bot, \mathscr{M}, \mathscr{M})$; *and*

- $\models^{\mathscr{M}'} \varphi \to \langle a \rangle \top$.

The minimal models resulting from revising a model $\mathscr{M}$ by a new executability law are those closest to $\mathscr{M}$ with respect to $\preceq_{\mathscr{M}}$:

**Definition 8.6** *Let* $\mathscr{M}$ *be a model and* $\varphi \to \langle a \rangle \top$ *be an executability law.* $revise(\mathscr{M}, \varphi \to \langle a \rangle \top) = \bigcup \min\{\mathscr{M}^{\star}_{\varphi \to \langle a \rangle \top}, \preceq_{\mathscr{M}}\}$.

In our running example, $revise(\mathscr{M}, \neg token \to \langle buy \rangle \top)$ is the singleton $\{\mathscr{M}'\}$, where $\mathscr{M}'$ is as depicted in Figure 18.

In this example, observe that because we have a single relevant target world we get a single model in the result of revision.

## 8.4 Revising Sets of Models

Up until now we have seen what the revision of single models means. This is needed when expansion by the new law is not possible due to inconsistency. We here give a unified definition of revision of a set of models $\mathcal{M}$ by a new law $\Phi$:

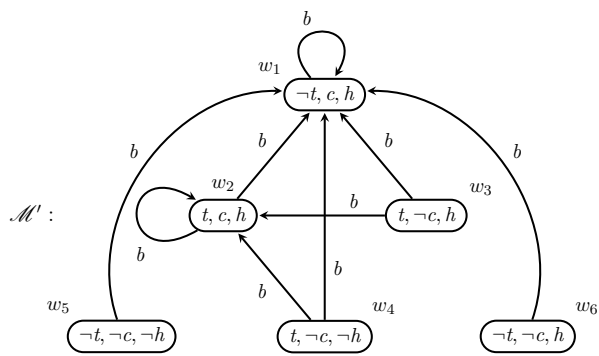

Figure 18: Result of revising $\mathscr{M}$ in Figure 15 by the new executability law $\neg token \rightarrow \langle buy \rangle \top$.

**Definition 8.7** *Let $\mathcal{M}$ be a set of models and $\Phi$ be a law. Then*

$$\mathcal{M}^{\star}_{\Phi} = \begin{cases} \mathcal{M} \setminus \{\mathscr{M} : \not\models^{\mathscr{M}} \Phi\}, \ \text{if there is } \mathscr{M} \in \mathcal{M} \ \text{such that } \models^{\mathscr{M}} \Phi; \\ \bigcup_{\mathscr{M} \in \mathcal{M}} revise(\mathscr{M}, \Phi), \ \text{otherwise.} \end{cases}$$

Observe that Definition 8.7 comprises both *expansion* and *revision*: in the first one, simple addition of the new law gives a satisfiable theory; in the latter a deeper change is needed to get rid of inconsistency.

## 9. Related Work

To the best of our knowledge, the first work on updating an action domain description is that by Li and Pereira (1996) in a narrative-based action description language (Gelfond & Lifschitz, 1993). Contrary to us, however, they mainly investigate the problem of updating the narrative with new observed *facts* and (possibly) with occurrences of actions that explain those facts. This amounts to updating a given state/configuration of the world (in our terms, what is true in a possible world) and focusing on the models of the narrative in which some actions took place (in our terms, the models of the action theory with a particular sequence of action executions). Clearly the models of the action laws remain the same.

Baral and Lobo (1997) introduce extensions of action languages that allow for some causal laws to be stated as defeasible. Their work is similar to ours in that they also allow for weakening of laws: in their setting, effect propositions can be replaced by what they call defeasible (weakened versions of) effect propositions. Our approach is different from theirs in the way executability laws are dealt with. Here executability laws are explicit and we are also able to contract them. This feature is important when the qualification problem is considered: we may always discover contexts that preclude the execution of a given action (cf. the Introduction).

Liberatore (2000) proposes a framework for reasoning about actions in which it is possible to express a given semantics of belief update, like Winslett's (1988) and Katsuno and Mendelzon's (1992). This means it is the formalism, essentially an action description lan-guage, that is used to describe updates (the change of propositions from one state of the world to another) by expressing them as laws in the action theory.

The main difference between Liberatore's work (2000) and Li and Pereira's (1996) is that, despite not being concerned, at least a priori, with changing action laws, Liberatore's framework allows for abductively introducing in the action theory new effect propositions (effect laws, in our terms) that consistently explain the occurrence of an event.

The work by Eiter et al. (2005) is similar to ours in that they also propose a framework which is oriented to updating action laws. They mainly investigate the case where e.g. a new effect law is added to the description (and then has to be true in all models of the modified theory). This problem is the dual of contraction and is then closer to our definition of revision (cf. Section 8).

In Eiter et al.'s framework (2005), action theories are described in a variant of a narrative-based action description language. Like in the present work, the semantics is also in terms of transition systems, with transitions (action occurrences) linking states (configurations of the world). Contrary to us, however, the minimality condition on the outcome of the update is in terms of inclusion of sets of laws, which means that the approach is more syntax oriented to some extent.

In their setting, during an update an action theory $\mathcal{T}$ is seen as composed of two pieces, $\mathcal{T}_u$ and $\mathcal{T}_m$, where $\mathcal{T}_u$ stands for the part of $\mathcal{T}$ that is not supposed to change and $\mathcal{T}_m$ contains the laws which may be modified. In our terms, when contracting a static law we would have $\mathcal{T}_m = \mathcal{S} \cup \mathcal{X}_a$; when contracting an executability law $\mathcal{T}_m = \mathcal{X}_a$; and when contracting effects laws $\mathcal{T}_m = \mathcal{E}_a^-$. The difference here is that in our approach it is always clear what laws should not change in a given type of contraction, and therefore $\mathcal{T}_u$ and $\mathcal{T}_m$ do not need to be explicitly specified prior to the update.

Their approach and ours can both be described as *constraint-based* update, in that the theory change is carried out relative to some constraints (a set of laws that we want to hold in the result). In our framework, for example, all changes in the action laws are relative to the set of static laws $\mathcal{S}$ (and that is why we concentrate on models of $\mathcal{T}$ having $val(\mathcal{S})$ as worlds). When changing a law, we want to keep the same set of states. The difference with respect to Eiter et al.'s (2005) approach is that there it is also possible to update a theory relatively to e.g. executability laws: when expanding $\mathcal{T}$ with a new effect law, one may want to constrain the change so that the action under concern is guaranteed to be executable in the result.[7] As shown in the referred work, this may require the withdrawal of some static law. Hence, in Eiter et al.'s framework, static laws do not have the same status as in ours.

Herzig et al. (2006) define a method for action theory contraction that, despite the similarity with the current work and the common underlying motivations, is more limited than the present constructions.

First, with the referred approach we do not get minimal change. For example, in the referred work the operator for contracting executability laws is such that in the resulting theory the modified set of executability laws is given by

$$\mathcal{X}_a^- = \{(\varphi_i \wedge \neg\varphi) \rightarrow \langle a \rangle\top : \varphi_i \rightarrow \langle a \rangle\top \in \mathcal{X}_a\}$$

---

7. We could simulate that in our approach with two successive modifications of $\mathcal{T}$: first adding the effect law and then an executability law (cf. Section 8).

which, according to its semantics, gives theories among whose models are those resulting from removing transitions from *all* $\varphi$-worlds. A similar comment can be made with respect to contraction of effect laws.

Second, Herzig et al.'s (2006) contraction method does not satisfy most of the postulates for action theory change that we have addressed in Section 7. Besides not satisfying the monotonicity postulate, it does not satisfy the preservation one. To witness, suppose that we have a language with only one atom $p$, and the model $\mathscr{M}$ depicted in Figure 19.

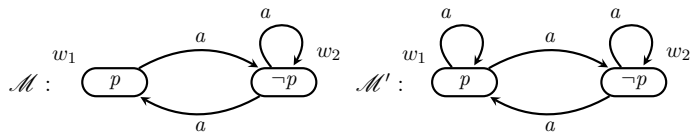

Figure 19: Counter-example to preservation in the method by Herzig et al. (2006).

Then $\models^{\mathscr{M}} p \to [a]\neg p$ and $\not\models^{\mathscr{M}} [a]\neg p$. Now the contraction operator defined there is such that when removing $[a]\neg p$ from $\mathscr{M}$ yields the model $\mathscr{M}'$ in Figure 19 such that $R'_a = W \times W$. Then $\not\models^{\mathscr{M}'} p \to [a]\neg p$, i.e., the effect law $p \to [a]\neg p$ is not preserved.

Finally, another work that is related to ours is that by Zhang and Ding (2008). Like ours, their approach is also about giving a semantic characterization of the basic operations for changing Kripke models. Contrary to us however, their focus is on *model checking*, not on entailment. Despite the definition and use of operations that in essence are similar to ours (modifications of the set of possible worlds or of the accessibility relation), their work is concerned mainly with modifications of a *single* model, not with that of sets of models as we do, and hence it does not provide operations for changing action laws. Because of that, their approach is not directly comparable to ours, since here we are interested in entailment-based revision.

## 10. Concluding Remarks

In this work we have addressed the problem of changing an action domain description for reasoning about actions, a problem not sufficiently investigated in the literature so far. We have seen the intuitions behind such a kind of theory modification and have given a semantics for action theory change in terms of distances between models that captures the notion of minimal change. We have given algorithms to contract a formula from a theory that terminate and that are correct with respect to our semantics (Corollary 6.1). We have shown the importance that our modularity notion has in this result and in others.

We have also extended Varzinczak's investigations (2008) by defining a semantics for action theory revision based on minimal modifications of models. For the corresponding revision algorithms, the reader is referred to the work by Varzinczak (2009). One of our ongoing research topics is on assessing our revision operators' behavior with respect to appropriate versions of the AGM postulates for revision (Alchourrón et al., 1985) and its links with the contraction counterpart.

With our algorithms we provide a set of tools to be used by the knowledge engineer in an interactive and possibly iterative way to modify an action theory. These tools are guaranteed to perform minimal change when assisting the knowledge engineer in implementing her desired modifications. They give her a set of options and it is up to the knowledge engineer to decide which one is more in line with her intuitions.

Given that action theory change is not a single step operation, the knowledge engineer is expected to make use of the contraction/revision operators to make a series of modifications that eventually will give a fine-grained theory not entailing the contracted laws and entailing the new learned laws about the domain.

For the sake of presentation, here we have abstracted from the frame and ramification problems. However our definitions could have been stated in a formalism with a suitable solution to them, like Castilho et al.'s approaches (1999, 2002). With regards to the qualification problem, this is not ignored here: contracting wrong executability laws is an approach towards its solution. Indeed, given the difficulty of stating all sufficient conditions for executability of an action, the knowledge engineer writes down some of them and lets the theory 'evolve' via subsequent revisions.

A possible criticism to the approach here developed concerns the cautiousness of our operator for contracting static laws: we prefer to lose some executability laws rather than induce them and lose effect laws. This behavior could make our operators to be interpreted as incoherent. We have pointed out nevertheless that this is in line with largely accepted assumptions in the RAC community, and moreover we have shown the impossibility of a non-cautious static law contraction operator that complies with all that and is coherent with the other operators.

Indeed one of the purposes of the present work is to shed some light on the fundamental differences between belief change in action domain descriptions and in logical theories in general. Classical belief change cannot be fully transplanted to action theories, and here we have shown why (cf. Sections 3.2, 4.2, 5.3 and 8.3).

In particular, looking at the postulates of classical belief change (or our versions thereof) one sees that they are not enough to fully characterize operators for action theory change. For that to be achieved the fundamental assumptions in reasoning about actions that we have extensively used throughout this work should somehow be 'compiled' into postulates supplementing the classical ones. It is not immediately clear what these new postulates would look like, but this is an interesting thread of investigation worth pursuing.

It might also be argued that our semantic operations do not respect the principle of categorical matching, given that the input and output are different sorts of objects, viz. a set of models and a set of sets of models (cf. Definitions 3.3, 3.7 and 3.10). It is easy to see, however, that our semantic constructions could have been defined in such a way that each $\mathcal{M}' \in \mathcal{M}_\Phi^-$ corresponds to the result of *one* contraction operator. The choice for defining the result of an operation as a set of possible outputs was driven by the definition of the algorithms, where a theory (corresponding to a set of models) is given as input and the output is a set of theories (hence corresponding to a set of set of models).

Although the semantic operators can be redefined to satisfy the principle of categorical matching, the same is not immediate about the algorithms (they would be non-deterministic). Therefore we preferred to keep a balance between the semantic and the syntactic definitions so that we see more clearly their direct correspondence.

One of our contentions here is that sticking to modular theories (and hence to canonical models) is not a big deal: we can use existent algorithms in the literature (Herzig & Varzinczak, 2007) to ensure that an action theory $\mathcal{T}$ is characterized by its canonical models.

We have seen that under modularity, our operators satisfy all the postulates for contraction: Modularity is one of the sufficient conditions for Success in Theorem 7.1. It is also a sufficient condition in Theorem 7.2, and, as shown in Theorem 7.3, it is a sufficient condition for Recovery. Finally it is also a sufficient condition for the Disjunctive Rule to hold, and is shown to be preserved by the contraction operators (cf. last paragraph of Section 7.2, proof in Appendix B). Preservation of modularity is an important result since it means that it has to be checked/ensured at most once during the lifetime of the action theory. All these results support the thesis that our modularity notion is fruitful.

By forcing formulas to be explicitly stated in their respective modules (and thus possibly making them inferable in independently different ways), modularity intuitively could be seen to diminish *elaboration tolerance* (McCarthy, 1998). For instance, when contracting a Boolean formula $\varphi$ in a non-modular theory, it seems reasonable to expect not to change the set of static laws $\mathcal{S}$, while the theory being modular surely forces changing such a module. It is not difficult, however, to conceive non-modular theories in which contraction of a formula $\varphi$ may demand a change in $\mathcal{S}$ as well. As an example, suppose $\mathcal{S} = \{\varphi_1 \rightarrow \varphi_2\}$ in an action theory from whose dynamic part we (implicitly) infer $\neg\varphi_2$. In this case, contracting $\neg\varphi_1$ while keeping $\neg\varphi_2$ would necessarily ask for a change in $\mathcal{S}$.

We point out nevertheless that in both cases (modular and non-modular) the extra work in changing other modules stays in the mechanical level, i.e., in the algorithms that carry out the modification, and does not augment in a significant way the amount of work the knowledge engineer is expected to do.

Contrary to the trend in the belief change community, where the focus is either on belief bases or belief sets (Hansson, 1999), the method here proposed is a hybrid one (Delgrande, 2009). On one hand, semantics plays a crucial role in the notion of minimal change here studied. On the other hand, we deal only with domain descriptions in reasoning about actions, which are sets of laws of specific types. On top of that, the modularity property (a syntactical one) is fundamental to our main results.

Following those lines, another issue that drives our future research on the subject is how to contract not only laws but *any* $\mathsf{K}_n$-formula. As defined, the order of application of our operators matter in the final result: if we contract $\varphi$ and then $\varphi \rightarrow [a]\psi$ from a theory $\mathcal{T}$, the result may not be the same as contracting $\varphi \rightarrow [a]\psi$ first and then removing $\varphi$. This problem would not appear in a more general framework in which any formula could be contracted: removing $\varphi \wedge (\varphi \rightarrow [a]\psi)$ should give the same result as $(\varphi \rightarrow [a]\psi) \wedge \varphi$. This is the principle of syntax independence (Dalal, 1988).

Related to that is the question on how our revision definitions relate to our contraction operators. What is known is that the Levi identity (1977), $\mathcal{T}^*_\Phi = \mathcal{T}_{\neg\Phi} \cup \{\Phi\}$, in general does not hold for action laws (effect and executability ones). The reason is that up to now there is no contraction operator for $\neg\Phi$ where $\Phi$ is an effect or an executability law. Indeed this is the general contraction problem for non-classical logics: contraction of a general formula (like $\neg\Phi$ above) is still an open problem in the belief change area. Some insights in this direction are given by our revision definitions, with which we make $\neg\Phi$ *false* in every possible world of a Kripke model.

Definitions 3.1, 3.5 and 3.8 appear to be important for better understanding the problem of contracting general formulas: basically the set of modifications to perform in a given model in order to force it to falsify a general formula will comprise removal/addition of transitions/worlds. The definition of a general revision/contraction method will then benefit from our constructions.

Furthermore, given the well-known connection between multimodal logics and Description Logics (Baader, Calvanese, McGuinness, Nardi, & Patel-Schneider, 2003), we believe that our definitions may also contribute to ontology evolution and debugging in some specific families of DLs.

## Acknowledgments

Parts of this work have been done during the author's stay at the Institut de Recherche en Informatique de Toulouse (IRIT), France, and during his visit to the National ICT Australia (NICTA), Sydney.

The author is grateful to the anonymous referees for their constructive and useful remarks, which helped improving the quality of the work. The paper has also benefited from discussions with Andreas Herzig and Laurent Perrussel.

Special thanks to my colleagues at the Meraka Institute Arina Britz, Ken Halland, Johannes Heidema and Tommie Meyer for their invaluable comments and suggestions on earlier versions of this article.

## Appendix A. Proof of Theorem 6.2

*Let $\mathcal{T}$ be modular, and $\Phi$ be a law. For all $\mathcal{M}' \in \mathcal{M}_{\Phi}^{-}$ such that $\models^{\mathcal{M}} \mathcal{T}$ for every $\mathcal{M} \in \mathcal{M}$, there is $\mathcal{T}' \in \mathcal{T}_{\Phi}^{-}$ such that $\models^{\mathcal{M}'} \mathcal{T}'$ for every $\mathcal{M}' \in \mathcal{M}'$.*

Before we give the proof of this theorem, we will need the following lemma (cf. the Monotonicity Postulate in Section 7.2):

**Lemma A.1** $\mathcal{T} \models_{\overline{\mathsf{K}}_n} \mathcal{T}'$.

**Proof:** Let $\mathcal{T}$ be an action theory, and let $\mathcal{T}' \in \mathcal{T}_{\Phi}^{-}$, for a given law $\Phi$. We are going to analyze each case.

Let $\Phi$ be of the form $\varphi \to \langle a \rangle \top$, for some $\varphi \in \mathfrak{Fml}$. Then $\mathcal{T}'$ is such that

$$\mathcal{T}' = (\mathcal{T} \setminus \mathcal{X}_a) \cup \{(\varphi_i \wedge \neg(\pi \wedge \varphi_A)) \to \langle a \rangle \top : \varphi_i \to \langle a \rangle \top \in \mathcal{X}_a\}$$

where $\pi \in IP(\mathcal{S} \wedge \varphi)$ and $\varphi_A = \bigwedge_{\substack{p_i \in \overline{atm(\pi)} \\ p_i \in A}} p_i \wedge \bigwedge_{\substack{p_i \in \overline{atm(\pi)} \\ p_i \notin A}} \neg p_i$, for some $A \subseteq \overline{atm(\pi)}$.

Let $\mathcal{M} = \langle W, R \rangle$ be such that $\models^{\mathcal{M}} \mathcal{T}$. It is enough to show that $\mathcal{M}$ is a model of the new laws. For every $(\varphi_i \wedge \neg(\pi \wedge \varphi_A)) \to \langle a \rangle \top$, for every $w \in W$, if $\models_{w}^{\mathcal{M}} \varphi_i \wedge \neg(\pi \wedge \varphi_A)$, then $\models_{w}^{\mathcal{M}} \varphi_i$. Because $\mathcal{T} \models_{\overline{\mathsf{K}}_n} \varphi_i \to \langle a \rangle \top$, $\models^{\mathcal{M}} \varphi_i \to \langle a \rangle \top$, and then $R_a(w) \neq \emptyset$.

Therefore we have that $\models^{\mathcal{M}} \mathcal{T}'$.

Let now $\Phi$ have the form $\varphi \to [a]\psi$, for $\varphi, \psi \in \mathfrak{Fml}$. Then $\mathcal{T}'$ is such that

$$\mathcal{T}' = \begin{aligned}&(\mathcal{T} \setminus \mathcal{E}_a^-) \cup \\ &\{(\varphi_i \wedge \neg(\pi \wedge \varphi_A)) \to [a]\psi_i : \varphi_i \to [a]\psi_i \in \mathcal{E}_a^-\} \cup \\ &\{(\varphi_i \wedge \pi \wedge \varphi_A) \to [a](\psi_i \vee \pi') : \varphi_i \to [a]\psi_i \in \mathcal{E}_a^- \cup \\ &\left\{(\pi \wedge \varphi_A \wedge \ell) \to [a](\psi \vee \ell) : \begin{array}{l} \ell \in L, \text{ for some } L \subseteq \mathfrak{Lit} \text{ s.t.} \\ \mathcal{S} \not\vdash (\pi' \wedge \bigwedge_{\ell \in L} \ell) \to \bot, \text{ and } \ell \in \pi' \\ \text{or } \mathcal{T} \not\models_{\overline{\mathsf{K}}_n} (\pi \wedge \varphi_A \wedge \ell) \to [a]\neg\ell \end{array}\right\}\end{aligned}$$

where $\mathcal{E}_a^- = \bigcup_{1 \le i \le n}(\mathcal{E}_a^{\varphi,\psi})_i$, $\pi \in IP(\mathcal{S} \wedge \varphi)$, $\varphi_A = \bigwedge_{\substack{p_i \in \overline{atm(\pi)} \\ p_i \in A}} p_i \wedge \bigwedge_{\substack{p_i \in \overline{atm(\pi)} \\ p_i \notin A}} \neg p_i$, for some $A \subseteq \overline{atm(\pi)}$, and $\pi' \in IP(\mathcal{S} \wedge \neg\psi)$.

Let $\mathscr{M} = \langle W, R \rangle$ be such that $\models^{\mathscr{M}} \mathcal{T}$. It is enough to show that $\mathscr{M}$ is a model of the added laws. Given $(\varphi_i \wedge \neg(\pi \wedge \varphi_A)) \to [a]\psi_i$, for every $w \in W$, if $\models_w^{\mathscr{M}} \varphi_i \wedge \neg(\pi \wedge \varphi_A)$, then $\models_w^{\mathscr{M}} \varphi_i$. Because $\mathcal{T} \models_{\overline{\mathsf{K}}_n} \varphi_i \to [a]\psi_i$, $\models^{\mathscr{M}} \varphi_i \to [a]\psi_i$, and then $\models_{w'}^{\mathscr{M}} \psi_i$ for every $w' \in W$ such that $(w, w') \in R_a$.

For $(\varphi_i \wedge \pi \wedge \varphi_A) \to [a](\psi_i \vee \pi')$, for every $w \in W$, if $\models_w^{\mathscr{M}} \varphi_i \wedge \pi \wedge \varphi_A$, then again $\models_{w'}^{\mathscr{M}} \psi_i$ for every $w' \in W$ such that $(w, w') \in R_a$.

Now, given $(\pi \wedge \varphi_A \wedge \ell) \to [a](\psi \vee \ell)$, for every $w \in W$, if $\models_w^{\mathscr{M}} \pi \wedge \varphi_A \wedge \ell$, then $\models_w^{\mathscr{M}} \pi$, and then $\models_w^{\mathscr{M}} \varphi$. Since $\mathcal{T} \models_{\overline{\mathsf{K}}_n} \varphi \to [a]\psi$, we have $\models^{\mathscr{M}} \varphi \to [a]\psi$, and then $\models_{w'}^{\mathscr{M}} \psi$ for every $w' \in W$ such that $(w, w') \in R_a$.

Therefore $\models^{\mathscr{M}} \mathcal{T}'$.

Let $\Phi$ be a propositional $\varphi$. Then $\mathcal{T}'$ is such that

$$\mathcal{T}' = \begin{aligned}&((\mathcal{T} \setminus \mathcal{S}) \cup \mathcal{S}^-) \setminus \mathcal{X}_a \cup \\ &\{(\varphi_i \wedge \varphi) \to \langle a\rangle\top : \varphi_i \to \langle a\rangle\top \in \mathcal{X}_a\} \cup \\ &\{\neg\varphi \to [a]\bot\}\end{aligned}$$

for some $\mathcal{S}^- \in \mathcal{S} \ominus \varphi$.

Let $\mathscr{M} = \langle W, R \rangle$ be such that $\models^{\mathscr{M}} \mathcal{T}$. It suffices to show that $\mathscr{M}$ satisfies the added laws.

Since we assume $\ominus$ behaves like a classical contraction operator, like e.g. Katsuno and Mendelzon's (1992), we have $\models_{\mathsf{CPL}} \mathcal{S} \to \mathcal{S}^-$, and then, because $\models^{\mathscr{M}} \mathcal{S}$, we have $\models^{\mathscr{M}} \mathcal{S}^-$.

Now given $(\varphi_i \wedge \varphi) \to \langle a\rangle\top$, for every $w \in W$, if $\models_w^{\mathscr{M}} \varphi_i \wedge \varphi$, then $\models_w^{\mathscr{M}} \varphi_i$, and because $\models^{\mathscr{M}} \varphi_i \to \langle a\rangle\top$, we have $R_a(w) \neq \emptyset$.

Finally, for $\neg\varphi \to [a]\bot$, because $\models^{\mathscr{M}} \varphi$, $\mathscr{M}$ trivially satisfies $\neg\varphi \to [a]\bot$.

Therefore $\models^{\mathscr{M}} \mathcal{T}'$. $\qquad\square$

**Proof of Theorem 6.2**

Let $\mathcal{M} = \{\mathscr{M} : \models^{\mathscr{M}} \mathcal{T}\}$, and $\mathscr{M}' \in \mathcal{M}_\Phi^-$. We show that there is $\mathcal{T}' \in \mathcal{T}_\Phi^-$ such that $\models^{\mathscr{M}'} \mathcal{T}'$ for every $\mathscr{M}' \in \mathcal{M}'$.

By definition, each $\mathscr{M}' \in \mathcal{M}'$ is such that either $\models^{\mathscr{M}'} \mathcal{T}$ or $\not\models^{\mathscr{M}'} \Phi$. Because $\mathcal{T}_\Phi^- \neq \emptyset$, there must be $\mathcal{T}' \in \mathcal{T}_\Phi$. If $\models^{\mathscr{M}'} \mathcal{T}$, by Lemma A.1 $\models^{\mathscr{M}'} \mathcal{T}'$ and we are done. Let us then suppose that $\not\models^{\mathscr{M}'} \Phi$. We analyze each case.

Let $\Phi$ have the form $\varphi \to \langle a \rangle \top$ for some $\varphi \in \mathfrak{Fml}$. Then $\mathscr{M}' = \langle W', R' \rangle$, where $W' = W$, $R' = R \setminus R_a^\varphi$, with $R_a^\varphi = \{(w, w') : \models_w^{\mathscr{M}'} \varphi \text{ and } (w, w') \in R_a\}$, for some $\mathscr{M} \in \mathcal{M}$.

Let $u \in W'$ be such that $\not\models_u^{\mathscr{M}'} \varphi \to \langle a \rangle \top$, i.e., $\models_u^{\mathscr{M}'} \varphi$ and $R_a'(u) = \emptyset$.

Because $u \Vdash \varphi$, there must be $v \in base(\varphi, W')$ such that $v \subseteq u$. Let $\pi = \bigwedge_{\ell \in v} \ell$. Clearly $\pi$ is a prime implicant of $\mathcal{S} \wedge \varphi$. Let also $\varphi_A = \bigwedge_{\ell \in u \setminus v} \ell$, and consider

$$\mathcal{T}' = (\mathcal{T} \setminus \mathcal{X}_a) \cup \{(\varphi_i \wedge \neg(\pi \wedge \varphi_A)) \to \langle a \rangle \top : \varphi_i \to \langle a \rangle \top \in \mathcal{X}_a\}$$

(Clearly, $\mathcal{T}'$ is a theory produced by Algorithm 1.)

It is enough to show that $\mathscr{M}'$ is a model of the new added laws. Given $(\varphi_i \wedge \neg(\pi \wedge \varphi_A)) \to \langle a \rangle \top \in \mathcal{T}'$, for every $w \in W'$, if $\models_w^{\mathscr{M}'} \varphi_i \wedge \neg(\pi \wedge \varphi_A)$, then $\models_w^{\mathscr{M}'} \varphi_i$, from what it follows $\models_w^{\mathscr{M}} \varphi_i$. Because $\models^{\mathscr{M}} \varphi_i \to \langle a \rangle \top$, there is $w' \in W$ such that $w' \in R_a(w)$. We need to show that $(w, w') \in R_a'$. If $\not\models_w^{\mathscr{M}'} \varphi$, then $R_a^\varphi = \emptyset$, and $(w, w') \in R_a'$. If $\models_w^{\mathscr{M}'} \varphi$, either $w = u$, and then from $\models_u^{\mathscr{M}'} \pi \wedge \varphi_A$ we conclude $\models_u^{\mathscr{M}'} (\varphi_i \wedge \neg(\pi \wedge \varphi_A)) \to \langle a \rangle \top$, or $w \neq u$ and then we must have $(w, w') \in R_a'$, otherwise there is $S_a^\varphi \subseteq R_a^\varphi$ such that $R \dot{-} (R \setminus S_a^\varphi) \subset R \dot{-} (R \setminus R_a^\varphi)$, and then $\mathscr{M}'' = \langle W', R \setminus S_a^\varphi \rangle$ is such that $\not\models^{\mathscr{M}''} \varphi \to \langle a \rangle \top$ and $\mathscr{M}'' \preceq_{\mathscr{M}} \mathscr{M}'$, a contradiction because $\mathscr{M}'$ is minimal with respect to $\preceq_{\mathscr{M}}$. Thus $(w, w') \in R_a'$, and then $\models_w^{\mathscr{M}'} \langle a \rangle \top$. Hence $\models^{\mathscr{M}'} \mathcal{T}'$.

Now let $\Phi$ be of the form $\varphi \to [a]\psi$, for $\varphi, \psi$ both Boolean. Then $\mathscr{M}' = \langle W', R' \rangle$, where $W' = W$, $R' = R \cup R_a^{\varphi, \neg \psi}$, with

$$R_a^{\varphi, \neg \psi} = \{(w, w') : w' \in RelTarget(w, \varphi \to [a]\psi, \mathscr{M}, \mathcal{M})\}$$

for some $\mathscr{M} = \langle W, R \rangle \in \mathcal{M}$.

Let $u \in W'$ be such that $\not\models_u^{\mathscr{M}'} \varphi \to [a]\psi$. Then there is $u' \in W'$ such that $(u, u') \in R_a'$ and $\not\models_{u'}^{\mathscr{M}'} \psi$. Because $u \Vdash \varphi$, there is $v \in base(\varphi, W')$ such that $v \subseteq u$, and as $u' \Vdash \neg \psi$, there must be $v' \in base(\neg \psi, W')$ such that $v' \subseteq u'$. Let $\pi = \bigwedge_{\ell \in v} \ell$, $\varphi_A = \bigwedge_{\ell \in u \setminus v} \ell$, and $\pi' = \bigwedge_{\ell \in v'} \ell$. Clearly $\pi$ (resp. $\pi'$) is a prime implicant of $\mathcal{S} \wedge \varphi$ (resp. $\mathcal{S} \wedge \neg \psi$).

Now let $\mathcal{E}_a^- = \bigcup_{1 \leq i \leq n} (\mathcal{E}_a^{\varphi, \psi})_i$ and let the theory

$$\mathcal{T}' = \begin{aligned} &(\mathcal{T} \setminus \mathcal{E}_a^-) \cup \\ &\{(\varphi_i \wedge \neg(\pi \wedge \varphi_A)) \to [a]\psi_i : \varphi_i \to [a]\psi_i \in \mathcal{E}_a^-\} \cup \\ &\{(\varphi_i \wedge \pi \wedge \varphi_A) \to [a](\psi_i \vee \pi') : \varphi_i \to [a]\psi_i \in \mathcal{E}_a^-\} \cup \\ &\left\{ (\pi \wedge \varphi_A \wedge \ell) \to [a](\psi \vee \ell) : \begin{matrix} \ell \in L, \text{ for some } L \subseteq \mathfrak{Lit} \text{ s.t.} \\ \mathcal{S} \not\vdash (\pi' \wedge \bigwedge_{\ell \in L} \ell) \to \bot, \text{ and } \ell \in \pi' \\ \text{or } \mathcal{T} \not\vdash_{\mathsf{K}_n} (\pi \wedge \varphi_A \wedge \ell) \to [a]\neg \ell \end{matrix} \right\} \end{aligned}$$

(Clearly, $\mathcal{T}'$ is a theory produced by Algorithm 2.)

In order to show that $\mathscr{M}'$ is a model of $\mathcal{T}'$, it is enough to show that it is a model of the added laws. Given $(\varphi_i \wedge \neg(\pi \wedge \varphi_A)) \to [a]\psi_i \in \mathcal{T}'$, for every $w \in W'$, if $\models^{\mathscr{M}'}_w \varphi_i \wedge \neg(\pi \wedge \varphi_A)$, then $\models^{\mathscr{M}'}_w \varphi_i$, and then $\models^{\mathscr{M}}_w \varphi_i$. Because $\models^{\mathscr{M}} \varphi_i \to [a]\psi_i$, $\models^{\mathscr{M}}_{w'} \psi_i$ for all $w' \in W$ such that $(w, w') \in R_a$. We need to show that $R'_a(w) = R_a(w)$. If $\not\models^{\mathscr{M}'}_w \varphi$, then $R_a^{\varphi, \neg\psi} = \emptyset$, and then $R'_a(w) = R_a(w)$. If $\models^{\mathscr{M}'}_w \varphi$, then either $w = u$, and from $\models^{\mathscr{M}'}_u \pi \wedge \varphi_A$ we conclude $\models^{\mathscr{M}'}_u (\varphi_i \wedge \neg(\pi \wedge \varphi_A)) \to [a]\psi_i$, or $w \neq u$, and then we must have $R_a^{\varphi, \neg\psi} = \emptyset$, otherwise there would be $S_a^{\varphi, \neg\psi} \subset R_a^{\varphi, \neg\psi}$ such that $R \dot{-} (R \cup S_a^{\varphi, \neg\psi}) \subset R \dot{-} (R \cup R_a^{\varphi, \neg\psi})$, and then $\mathscr{M}'' = \langle W', R \cup S_a^{\varphi, \neg\psi}\rangle$ would be such that $\not\models^{\mathscr{M}''} \varphi \to [a]\psi$ and $\mathscr{M}'' \preceq_{\mathscr{M}} \mathscr{M}'$, a contradiction since $\mathscr{M}'$ is minimal with respect to $\preceq_{\mathscr{M}}$. Hence $R'_a(w) = R_a(w)$, and $\models^{\mathscr{M}'}_{w'} \psi_i$ for all $w'$ such that $(w, w') \in R'_a$.

Now, given $(\varphi_i \wedge \pi \wedge \varphi_A) \to [a](\psi_i \vee \pi')$, for every $w \in W'$, if $\models^{\mathscr{M}'}_w \varphi_i \wedge \pi \wedge \varphi_A$, then $\models^{\mathscr{M}'}_w \varphi_i$, and then $\models^{\mathscr{M}}_w \varphi_i$. Because, $\models^{\mathscr{M}} \varphi_i \to [a]\psi_i$, we have $\models^{\mathscr{M}}_{w'} \psi_i$ for all $w' \in W$ such that $(w, w') \in R_a$, and then $\models^{\mathscr{M}'}_{w'} \psi_i$ for every $w' \in W'$ such that $(w, w') \in R'_a \setminus R_a^{\varphi, \neg\psi}$. Now, given $(w, w') \in R_a^{\varphi, \neg\psi}$, $\models^{\mathscr{M}'}_{w'} \pi'$, and the result follows.

Now, for each $(\pi \wedge \varphi_A \wedge \ell) \to [a](\psi \vee \ell)$, for every $w \in W'$, if $\models^{\mathscr{M}'}_w \pi \wedge \varphi_A \wedge \ell$, then $\models^{\mathscr{M}'}_w \varphi$, and then $\models^{\mathscr{M}}_w \varphi$. Because $\models^{\mathscr{M}} \varphi \to [a]\psi$, we have $\models^{\mathscr{M}}_{w'} \psi$ for every $w' \in W$ such that $(w, w') \in R_a$, and then $\models^{\mathscr{M}'}_{w'} \psi$ for all $w' \in W'$ such that $(w, w') \in R'_a \setminus R_a^{\varphi, \neg\psi}$. It remains to show that $\models^{\mathscr{M}'}_{w'} \ell$ for every $w' \in W'$ such that $(w, w') \in R_a^{\varphi, \neg\psi}$. Since $\mathscr{M}'$ is minimal, it is enough to show that $\models^{\mathscr{M}'}_{u'} \ell$ for every $\ell \in \mathfrak{Lit}$ such that $\models^{\mathscr{M}'}_u \pi \wedge \varphi_A \wedge \ell$. If $\ell \in \pi'$, the result follows. Otherwise, suppose $\not\models^{\mathscr{M}'}_{u'} \ell$. Then

- either $\neg\ell \in \pi'$, then $\pi'$ and $\ell$ are unsatisfiable, and in this case Algorithm 2 has not put the law $(\pi \wedge \varphi_A \wedge \ell) \to [a](\psi \vee \ell)$ in $\mathcal{T}'$, a contradiction;

- or $\neg\ell \in u' \setminus v'$. In this case, there is a valuation $u'' = (u' \setminus \{\neg\ell\}) \cup \{\ell\}$ such that $u'' \not\models \psi$. We must have $u'' \in W$, otherwise there will be $L' = \{\ell_i : \ell_i \in u''\}$ such that $\mathcal{T} \models_{\overline{\mathsf{K}}_n} (\pi' \wedge \bigwedge_{\ell_i \in L'} \ell_i) \to \bot$, and, because $\mathcal{T}$ is modular, $\mathcal{S} \models_{\mathsf{CPL}} (\pi' \wedge \bigwedge_{\ell_i \in L'} \ell_i) \to \bot$, and then Algorithm 2 has not put the law $(\pi \wedge \varphi_A \wedge \ell) \to [a](\psi \vee \ell)$ in $\mathcal{T}'$, a contradiction. Then $u'' \in W$, and moreover $u'' \notin R_a^{\varphi, \neg\psi}(u)$, otherwise $\mathscr{M}'$ is not minimal. As $u'' \setminus u \subset u' \setminus u$, the only reason why $u'' \notin R_a^{\varphi, \neg\psi}(u)$ is that there is $\ell' \in u \cap u''$ such that $\models^{\mathscr{M}_i} \bigwedge_{\ell_j \in u} \ell_j \to [a]\neg\ell'$ for every $\mathscr{M}_i \in \mathcal{M}$ if and only if $\ell' \notin v'$ for any $v' \in base(\neg\psi, W')$ such that $v' \subseteq u''$. Clearly $\ell' = \ell$, and because $\ell \notin \pi'$, we have $\models^{\mathscr{M}_i} \bigwedge_{\ell_j \in u} \ell_j \to [a]\neg\ell$ for every $\mathscr{M}_i \in \mathcal{M}$. Then $\mathcal{T} \models_{\overline{\mathsf{K}}_n} (\pi \wedge \varphi_A \wedge \ell) \to [a]\neg\ell$, and then Algorithm 2 has not put the law $(\pi \wedge \varphi_A \wedge \ell) \to [a](\psi \vee \ell)$ in $\mathcal{T}'$, a contradiction.

Hence we have $\models^{\mathscr{M}'}_{w'} \psi \vee \ell$ for every $w' \in W'$ such that $(w, w') \in R'_a$.

Putting the above results together, we get $\models^{\mathscr{M}'} \mathcal{T}'$.

Let now $\Phi$ be some propositional $\varphi$. Then $\mathscr{M}' = \langle W', R'\rangle$, where $W \subseteq W'$, $R' = R$, is minimal with respect to $\preceq_{\mathscr{M}}$, i.e., $W'$ is a minimal superset of $W$ such that there is $u \in W'$ with $u \nVdash \varphi$. Because we have assumed the syntactical classical contraction operator is sound and complete with respect to its semantics and is moreover minimal, then there must be $\mathcal{S}^- \in \mathcal{S} \ominus \varphi$ such that $W' = val(\mathcal{S}^-)$. Therefore $\models^{\mathscr{M}'} \mathcal{S}^-$.

Because $R' = R$, every effect law of $\mathcal{T}$ remains true in $\mathscr{M}'$.

Now, let

$$
\begin{aligned}
& ((\mathcal{T} \setminus \mathcal{S}) \cup \mathcal{S}^-) \setminus \mathcal{X}_a \,\cup \\
\mathcal{T}' = \; & \{(\varphi_i \wedge \varphi) \rightarrow \langle a \rangle \top : \varphi_i \rightarrow \langle a \rangle \top \in \mathcal{X}_a\} \,\cup \\
& \{\neg \varphi \rightarrow [a] \bot\}
\end{aligned}
$$

(Clearly, $\mathcal{T}'$ is a theory produced by Algorithm 3.)

For every $(\varphi_i \wedge \varphi) \rightarrow \langle a \rangle \top \in \mathcal{T}'$ and every $w \in W'$, if $\models_w^{\mathscr{M}'} \varphi_i \wedge \varphi$, then $R_a(w) \neq \emptyset$, because $\models_w^{\mathscr{M}} \varphi_i \rightarrow \langle a \rangle \top$. Given $\neg \varphi \rightarrow [a] \bot$, for every $w \in W'$, if $\models_w^{\mathscr{M}'} \neg \varphi$, then $w = u$, and $R_a(w) = \emptyset$.

Putting all these results together, we have $\models^{\mathscr{M}'} \mathcal{T}'$. $\qquad\square$

## Appendix B. Proof of Theorem 6.3

*Let $\mathcal{T}$ be modular, $\Phi$ a law, and $\mathcal{T}' \in \mathcal{T}_\Phi^-$. For all $\mathscr{M}'$ such that $\models^{\mathscr{M}'} \mathcal{T}'$, there is $\mathcal{M}' \in \mathcal{M}_\Phi^-$ such that $\mathscr{M}' \in \mathcal{M}'$ and $\models^{\mathscr{M}} \mathcal{T}$ for every $\mathscr{M} \in \mathcal{M}$.*

In order to prove this result, first we need to show four important lemmas.

**Lemma B.1** *Let $\Phi$ be a law. If $\mathcal{T}$ is modular, then every $\mathcal{T}' \in \mathcal{T}_\Phi^-$ is modular.*

**Proof:** Let $\Phi$ be nonclassical, and suppose there is $\mathcal{T}' \in \mathcal{T}_\Phi^-$ such that $\mathcal{T}'$ is not modular. Then there is some $\varphi' \in \mathfrak{Fml}$ such that $\mathcal{T}' \models_{\overline{\mathsf{R}}_n} \varphi'$ and $\mathcal{S}' \nvDash_{\mathsf{CPL}} \varphi'$, where $\mathcal{S}'$ is the set of static laws in $\mathcal{T}'$. By Lemma A.1, $\mathcal{T} \models_{\overline{\mathsf{R}}_n} \mathcal{T}'$, and then we have $\mathcal{T} \models_{\overline{\mathsf{R}}_n} \varphi'$. Because $\Phi$ is nonclassical, $\mathcal{S}' = \mathcal{S}$. Thus $\mathcal{S} \nvDash_{\mathsf{CPL}} \varphi'$, and therefore $\mathcal{T}$ is not modular.

Let now $\Phi$ be some $\varphi \in \mathfrak{Fml}$. Then

$$
\begin{aligned}
& ((\mathcal{T} \setminus \mathcal{S}) \cup \mathcal{S}^-) \setminus \mathcal{X}_a \,\cup \\
\mathcal{T}' = \; & \{(\varphi_i \wedge \varphi) \rightarrow \langle a \rangle \top : \varphi_i \rightarrow \langle a \rangle \top \in \mathcal{X}_a\} \,\cup \\
& \{\neg \varphi \rightarrow [a] \bot\}
\end{aligned}
$$

for some $\mathcal{S}^- \in \mathcal{S} \ominus \varphi$.

Assume that $\mathcal{T}$ is modular, and let $\varphi' \in \mathfrak{Fml}$ be such that $\mathcal{T}' \models_{\overline{\mathsf{R}}_n} \varphi'$ and $\mathcal{S}^- \nvDash_{\mathsf{CPL}} \varphi'$.

As $\mathcal{S}^- \nvDash_{\mathsf{CPL}} \varphi'$, there is $v \in val(\mathcal{S}^-)$ such that $v \nVdash \varphi'$. If $v \in val(\mathcal{S})$, then $\mathcal{S} \nvDash_{\mathsf{CPL}} \varphi'$, and as $\mathcal{T}$ is modular, $\mathcal{T} \nvDash_{\overline{\mathsf{R}}_n} \varphi'$. By Lemma A.1, $\mathcal{T} \models_{\overline{\mathsf{R}}_n} \mathcal{T}'$, and we have $\mathcal{T}' \nvDash_{\overline{\mathsf{R}}_n} \varphi'$, a contradiction. Hence $v \notin val(\mathcal{S})$. Moreover, we must have $v \nVdash \varphi$, otherwise $\ominus$ has not worked as expected.

Let $\mathscr{M} = \langle W, R \rangle$ be such that $\models^{\mathscr{M}} \mathcal{T}'$. (We extend $\mathscr{M}$ to another model of $\mathcal{T}'$.) Let $\mathscr{M}' = \langle W', R' \rangle$ be such that $W' = W \cup \{v\}$ and $R' = R$. To show that $\mathscr{M}'$ is a model of $\mathcal{T}'$, it suffices to show that $v$ satisfies every law in $\mathcal{T}'$. As $v \in val(\mathcal{S}^-)$, $\models_v^{\mathscr{M}'} \mathcal{S}^-$. Given $\neg\varphi \to [a]\bot \in \mathcal{T}'$, as $v \not\Vdash \varphi$ and $R'_a(v) = \emptyset$, $\models_v^{\mathscr{M}'} \neg\varphi \to [a]\bot$. Now, for every $\varphi_i \to [a]\psi_i \in \mathcal{T}'$, if $\models_v^{\mathscr{M}'} \varphi_i$, then we trivially have $\models_{v'}^{\mathscr{M}'} \psi_i$ for every $v'$ such that $(v, v') \in R'_a$. Finally, given $(\varphi_i \wedge \varphi) \to \langle a \rangle \top \in \mathcal{T}'$, as $v \not\Vdash \varphi$, the formula trivially holds in $v$. Hence $\models^{\mathscr{M}'} \mathcal{T}'$, and because there is $v \in W'$ such that $\not\models_v^{\mathscr{M}'} \varphi'$, we have $\mathcal{T}' \not\models_{\overline{\mathsf{K}}_n} \varphi'$, a contradiction. Hence for all $\varphi' \in \mathfrak{Fml}$ such that $\mathcal{T}' \models_{\overline{\mathsf{K}}_n} \varphi'$, $\mathcal{S}^- \models_{\mathsf{CPL}} \varphi'$, and then $\mathcal{T}'$ is modular. $\qquad\square$

**Lemma B.2** *If $\mathscr{M}_{can} = \langle W_{can}, R_{can} \rangle$ is a model of $\mathcal{T}$, then for every $\mathscr{M} = \langle W, R \rangle$ such that $\models^{\mathscr{M}} \mathcal{T}$ there is a minimal (with respect to set inclusion) extension $R' \subseteq R_{can} \setminus R$ such that $\mathscr{M}' = \langle val(\mathcal{S}), R \cup R' \rangle$ is a model of $\mathcal{T}$.*

**Proof:** Let $\mathscr{M}_{can} = \langle W_{can}, R_{can} \rangle$ be a model of $\mathcal{T}$, and let $\mathscr{M} = \langle W, R \rangle$ be such that $\models^{\mathscr{M}} \mathcal{T}$. Consider $\mathscr{M}' = \langle val(\mathcal{S}), R \rangle$. If $\models^{\mathscr{M}'} \mathcal{T}$, we have $R' = \emptyset \subseteq R_{can} \setminus R$ that is minimal. Suppose then $\not\models^{\mathscr{M}'} \mathcal{T}$. We extend $\mathscr{M}'$ to a model of $\mathcal{T}$ that is a minimal extension of $\mathscr{M}$. As $\not\models^{\mathscr{M}'} \mathcal{T}$, there is $v \in val(\mathcal{S}) \setminus W$ such that $\not\models_v^{\mathscr{M}'} \mathcal{T}$. Then there is $\Phi \in \mathcal{T}$ such that $\not\models_v^{\mathscr{M}'} \Phi$. If $\Phi$ is some $\varphi \in \mathfrak{Fml}$, as $v \in W_{can}$, $\mathscr{M}_{can}$ is not a model of $\mathcal{T}$. If $\Phi$ is of the form $\varphi \to [a]\psi$, for $\varphi, \psi \in \mathfrak{Fml}$, there is $v' \in val(\mathcal{S})$ such that $(v, v') \in R_a$ and $v' \not\Vdash \psi$, a contradiction since $R_a(v) = \emptyset$. Let now $\Phi$ have the form $\varphi \to \langle a \rangle \top$ for some $\varphi \in \mathfrak{Fml}$. Then $\models_v^{\mathscr{M}'} \varphi$. As $v \in W_{can}$, if $\not\models_v^{\mathscr{M}_{can}} \varphi \to \langle a \rangle \top$, then $\not\models^{\mathscr{M}_{can}} \mathcal{T}$. Hence, $R_{can_a}(v) \neq \emptyset$. Thus taking any $(v, v') \in R_{can_a}$ gives us a minimal $R' = \{(v, v')\}$ such that $\mathscr{M}'' = \langle val(\mathcal{S}), R \cup R' \rangle$ is a model of $\mathcal{T}$. $\qquad\square$

**Lemma B.3** *Let $\mathcal{T}$ be modular, and $\Phi$ be a law. Then $\mathcal{T} \models_{\overline{\mathsf{K}}_n} \Phi$ if and only if every $\mathscr{M}' = \langle val(\mathcal{S}), R' \rangle$ such that $\models^{\langle W, R \rangle} \mathcal{T}$ and $R \subseteq R'$ is a model of $\Phi$.*

**Proof:**
($\Rightarrow$): Straightforward, since $\mathcal{T} \models_{\overline{\mathsf{K}}_n} \Phi$ implies $\models^{\mathscr{M}} \Phi$ for every $\mathscr{M}$ such that $\models^{\mathscr{M}} \mathcal{T}$, in particular for those which are extensions of some model of $\mathcal{T}$.

($\Leftarrow$): Suppose $\mathcal{T} \not\models_{\overline{\mathsf{K}}_n} \Phi$. Then there is $\mathscr{M} = \langle W, R \rangle$ such that $\models^{\mathscr{M}} \mathcal{T}$ and $\not\models^{\mathscr{M}} \Phi$. As $\mathcal{T}$ is modular, the canonical frame $\mathscr{M}_{can} = \langle W_{can}, R_{can} \rangle$ of $\mathcal{T}$ is a model of $\mathcal{T}$. Then by Lemma B.2 there is a minimal extension $R'$ of $R$ with respect to $R_{can}$ such that $\mathscr{M}' = \langle val(\mathcal{S}), R \cup R' \rangle$ is a model of $\mathcal{T}$. Because $\not\models^{\mathscr{M}} \Phi$, there is $w \in W$ such that $\not\models_w^{\mathscr{M}} \Phi$. If $\Phi$ is some propositional $\varphi \in \mathfrak{Fml}$ or an effect law, any extension $\mathscr{M}'$ of $\mathscr{M}$ is such that $\not\models_w^{\mathscr{M}'} \Phi$. If $\Phi$ is of the form $\varphi \to \langle a \rangle \top$, then $\models_w^{\mathscr{M}} \varphi$ and $R_a(w) = \emptyset$. As any extension of $\mathscr{M}$ is such that $(u, v) \in R'$ if and only if $u \in val(\mathcal{S}) \setminus W$, only worlds other than those in $W$ get a new departing transition. Thus $(R \cup R')_a(w) = \emptyset$, and then $\not\models_w^{\mathscr{M}'} \Phi$. $\qquad\square$

**Lemma B.4** *Let $\mathcal{T}$ be modular, $\Phi$ a law, and $\mathcal{T}' \in \mathcal{T}_{\Phi}^-$. If $\mathscr{M}' = \langle val(\mathcal{S}'), R' \rangle$ is a model of $\mathcal{T}'$, then there is $\mathcal{M} = \{\mathscr{M} : \mathscr{M} = \langle val(\mathcal{S}), R \rangle$ and $\models^{\mathscr{M}} \mathcal{T}\}$ such that $\mathscr{M}' \in \mathcal{M}'$ for some $\mathcal{M}' \in \mathcal{M}_{\Phi}^-$.*

**Proof:** Let $\mathscr{M}' = \langle val(\mathcal{S}'), R' \rangle$ be such that $\models^{\mathscr{M}'} \mathcal{T}'$. If $\models^{\mathscr{M}'} \mathcal{T}$, the result follows. Let us suppose then $\not\models^{\mathscr{M}'} \mathcal{T}$. We analyze each case.

Let $\Phi$ be of the form $\varphi \rightarrow \langle a \rangle \top$, for some $\varphi \in \mathfrak{Fml}$. Let $\mathcal{M} = \{\mathscr{M} : \mathscr{M} = \langle val(\mathcal{S}), R \rangle\}$. Since by hypothesis $\mathcal{T}$ is modular, from Lemmas B.2 and B.3 it follows that $\mathcal{M}$ is non-empty and contains only models of $\mathcal{T}$.

Suppose $\mathscr{M}'$ is not a minimal model of $\mathcal{T}'$, i.e., there is $\mathscr{M}''$ such that $\mathscr{M}'' \preceq_{\mathscr{M}} \mathscr{M}'$ for some $\mathscr{M} \in \mathcal{M}$. Then $\mathscr{M}'$ and $\mathscr{M}''$ differ only in the executability of $a$ in a given $\varphi$-world, viz. a $\pi \wedge \varphi_A$-context, for some $\pi \in IP(\mathcal{S} \wedge \varphi)$ and $\varphi_A = \bigwedge_{\substack{p_i \in \overline{atm(\pi)} \\ p_i \in A}} p_i \wedge \bigwedge_{\substack{p_i \in \overline{atm(\pi)} \\ p_i \notin A}} \neg p_i$ such that $A \subseteq \overline{atm(\pi)}$. Because $\not\models^{\mathscr{M}'} (\pi \wedge \varphi_A) \rightarrow \langle a \rangle \top$, we must have $\models^{\mathscr{M}''} (\pi \wedge \varphi_A) \rightarrow \langle a \rangle \top$ and then $\models^{\mathscr{M}''} \mathcal{T}$. Hence $\mathscr{M}'$ is minimal with respect to $\preceq_{\mathscr{M}}$.

When contracting executability laws, $\mathcal{S}' = \mathcal{S}$. Hence taking the right $R$ and a minimal $R_a^\varphi$ such that $\mathscr{M} = \langle val(\mathcal{S}), R \rangle$ and $R' = R \setminus R_a^\varphi$, for some $R_a^\varphi \subseteq \{(w, w') : \not\models^{\mathscr{M}}_w \varphi$ and $(w, w') \in R_a\}$, we construct $\mathscr{M}' = \mathcal{M} \cup \{\mathscr{M}'\} \in \mathcal{M}^-_{\varphi \rightarrow \langle a \rangle \top}$.

Let $\Phi$ now be of the form $\varphi \rightarrow [a]\psi$, for $\varphi, \psi \in \mathfrak{Fml}$. Let $\mathcal{M} = \{\mathscr{M} : \mathscr{M} = \langle val(\mathcal{S}), R \rangle\}$. Since by hypothesis $\mathcal{T}$ is modular, from Lemmas B.2 and B.3 it follows that $\mathcal{M}$ is non-empty and contains only models of $\mathcal{T}$.

We claim that $\mathscr{M}'$ has only one transition linking a $\varphi$-world, viz. a context $\varphi_i \wedge \pi \wedge \varphi_A$ for some $\pi \in IP(\mathcal{S} \wedge \varphi)$ and $\varphi_A = \bigwedge_{\substack{p_i \in \overline{atm(\pi)} \\ p_i \in A}} p_i \wedge \bigwedge_{\substack{p_i \in \overline{atm(\pi)} \\ p_i \notin A}} \neg p_i$, such that $A \subseteq \overline{atm(\pi)}$, to a $\pi'$-world, where $\pi' \in IP(\mathcal{S} \wedge \neg\psi)$. The proof is as follows: given $\ell \in \mathfrak{Lit}$ such that $\ell$ holds in this $\varphi_i \wedge \pi \wedge \varphi_A$-world

- if $(\pi \wedge \varphi_A \wedge \ell) \rightarrow [a](\psi \vee \ell) \notin \mathcal{T}'$, then $\ell \notin \pi'$ and $\mathcal{T} \models_{\overline{\mathsf{K}_n}} (\pi \wedge \varphi_A \wedge \ell) \rightarrow [a]\neg\ell$. Then this world has only $\neg\ell$-successors.

- if $(\pi \wedge \varphi_A \wedge \ell) \rightarrow [a](\psi \vee \ell) \in \mathcal{T}'$, then every $\pi'$-successor is an $\ell$-world.

By successively applying this reasoning to each $\ell$ that holds in this $\varphi_i \wedge \pi \wedge \varphi_A$-world, we will end up with only one $\pi'$-successor.

Suppose now that $\mathscr{M}'$ is not a minimal model of $\mathcal{T}'$, i.e., there is $\mathscr{M}''$ such that $\models^{\mathscr{M}''} \mathcal{T}'$ and $\mathscr{M}'' \preceq_{\mathscr{M}} \mathscr{M}'$ for some $\mathscr{M} \in \mathcal{M}$. Then $\mathscr{M}'$ and $\mathscr{M}''$ differ only in the effects on that $\varphi_i \wedge \pi \wedge \varphi_A$-world: $\mathscr{M}''$ has no transition linking it to a $\pi'$-world. Then we have $\models^{\mathscr{M}''} (\varphi_i \wedge \pi \wedge \varphi_A) \rightarrow [a]\psi_i$, and then $\models^{\mathscr{M}''} \mathcal{T}$. Therefore $\mathscr{M}'$ is a minimal model of $\mathcal{T}'$ with respect to $\preceq_{\mathscr{M}}$.

When contracting effect laws, $\mathcal{S}' = \mathcal{S}$. Thus taking the right $R$ and a minimal $R_a^{\varphi, \psi}$ such that $\mathscr{M} = \langle val(\mathcal{S}), R \rangle$ and $R' = R \cup R_a^{\varphi, \psi}$, for some $R_a^{\varphi, \psi} \subseteq \{(w, w') : \models^{\mathscr{M}}_w \varphi$ and $w' \in RelTarget(w, \varphi \rightarrow [a]\psi, \mathscr{M}, \mathcal{M})\}$, we construct $\mathscr{M}' = \mathcal{M} \cup \{\mathscr{M}'\} \in \mathcal{M}^-_{\varphi \rightarrow [a]\psi}$.

Let now $\Phi$ be $\varphi$ for some $\varphi \in \mathfrak{Fml}$. Since $\mathcal{T}$ is modular, by Lemmas B.2 and B.3 there is $\mathscr{M} = \langle val(\mathcal{S}), R \rangle$ such that $\models^{\mathscr{M}} \mathcal{T}$. We know $val(\mathcal{S}) \subseteq val(\mathcal{S}^-)$. Because $\neg\varphi \rightarrow [a]\bot \in \mathcal{T}'$, $R'_a(v) = \emptyset$ for every $\neg\varphi$-world $v$ added in $\mathscr{M}'$. Hence, because $\ominus$ is minimal, taking $\mathcal{M} = \{\mathscr{M}\}$ gives us the result. $\qquad \square$

**Proof of Theorem 6.3**

From the hypothesis that $\mathcal{T}$ is modular and Lemma B.1, it follows that $\mathcal{T}'$ is modular, too. Then $\mathcal{M}' = \langle val(\mathcal{S}'), R \rangle$ is a model of $\mathcal{T}'$, by Lemma B.3. From this and Lemma B.4 the result follows. $\qquad\square$